\documentclass[10pt,journal,compsoc]{IEEEtran}

\usepackage{amsmath}
\usepackage{diagbox}
\usepackage{graphics}
\usepackage{epsfig}
\usepackage{threeparttable}
\usepackage{xspace}
\usepackage{siunitx}
\usepackage{graphicx}
\usepackage{amssymb}
\usepackage{threeparttable}
\usepackage{color}
\usepackage[normalem]{ulem}
\usepackage{multirow}
\usepackage{float}
\usepackage{amsfonts}
\usepackage{bm}
\usepackage{array}
\usepackage[table]{xcolor}
\usepackage{colortbl}
\usepackage{diagbox}
\usepackage{rotating}
\usepackage{booktabs}
\usepackage{overpic}
\usepackage{enumitem}
\usepackage{colortbl}
\usepackage{algorithm}
\usepackage{algorithmic}
\usepackage{cite}
\usepackage{dblfloatfix}
\usepackage[export]{adjustbox}
\usepackage{pifont}
\usepackage[pagebackref=false,breaklinks=true,colorlinks,bookmarks=false]{hyperref}

\definecolor{mygray}{gray}{.85}
\definecolor{mygray1}{gray}{.7}
\definecolor{mygray2}{gray}{.93}

\usepackage{algorithm}
\usepackage{listings}
\usepackage{etoolbox}
\makeatletter
\AfterEndEnvironment{algorithm}{\let\@algcomment\relax}
\AtEndEnvironment{algorithm}{\kern2pt\hrule\relax\vskip3pt\@algcomment}
\let\@algcomment\relax
\newcommand\algcomment[1]{\def\@algcomment{\footnotesize#1}}
\renewcommand\fs@ruled{\def\@fs@cfont{\bfseries}\let\@fs@capt\floatc@ruled
  \def\@fs@pre{\hrule height.8pt depth0pt \kern2pt}%
  \def\@fs@post{}%
  \def\@fs@mid{\kern2pt\hrule\kern2pt}%
  \let\@fs@iftopcapt\iftrue}
\makeatother

\usepackage{tabulary}
\newcolumntype{I}{!{\vrule width 1pt}}
\newcolumntype{x}[1]{>{\centering\arraybackslash}p{#1pt}}
\newcolumntype{y}[1]{>{\raggedright\arraybackslash}p{#1pt}}
\newcolumntype{z}[1]{>{\raggedleft\arraybackslash}p{#1pt}}

\definecolor{codegreen}{RGB}{79,126,127}
\definecolor{codedefine}{RGB}{153,54,159}
\definecolor{codefunc}{RGB}{73,122,234}
\definecolor{codecall}{RGB}{73,122,234}
\definecolor{codepro}{RGB}{212,96,80}
\definecolor{codedim}{RGB}{89,152,195}
\newcommand{\pub}[1]{{\color{gray}{\tiny{[{#1}]\!}}}}

\makeatletter
\newcommand{\thickhline}{%
    \noalign {\ifnum 0=`}\fi \hrule height 1pt
    \futurelet \reserved@a \@xhline
}
\makeatother

\makeatletter
\DeclareRobustCommand\onedot{\futurelet\@let@token\@onedot}
\def\@onedot{\ifx\@let@token.\else.\null\fi\xspace}

\def\eg{\emph{e.g}\onedot} 
\def\ie{\emph{i.e}\onedot} 
\def\cf{\emph{c.f}\onedot} 
\def\etc{\emph{etc}\onedot} 
 
\def\etal{\emph{et al}\onedot}

\makeatother

\newcommand{\redminor}[1]{{\textcolor{black}{{#1}}}}

\begin{document}
\title{Towards Data-and Knowledge-Driven AI:\\ A Survey on Neuro-Symbolic Computing}

\author{Wenguan~Wang,~\IEEEmembership{Senior~Member,~IEEE,}\\
Yi Yang,~\IEEEmembership{Senior~Member,~IEEE}, and Fei Wu,~\IEEEmembership{Senior~Member,~IEEE}
\IEEEcompsocitemizethanks{
\IEEEcompsocthanksitem W. Wang, Y. Yang, and F. Wu are with College of Computer Science and Technology, Zhejiang University (Email: wenguanwang.ai@gmail.com, yangyics@zju.edu.cn, wufei@cs.zju.edu.cn)
\IEEEcompsocthanksitem Corresponding Author: Wenguan~Wang
}
}

\markboth{IEEE TRANSACTIONS ON PATTERN ANALYSIS AND MACHINE INTELLIGENCE}%
{Shell \MakeLowercase{\textit{et al.}}: Bare Demo of IEEEtran.cls for Journals}

\IEEEtitleabstractindextext{
\begin{abstract}
Neural-symbolic computing (NeSy), which pursues the integration of the symbolic and statistical paradigms of cognition, has been an active research area of Artificial Intelligence (AI) for many years. As NeSy shows promise of reconciling the advantages of reasoning and interpretability of symbolic representation and robust learning in neural networks, it may serve as a catalyst for the next generation of AI. In the present paper, we provide a systematic overview of the recent developments and  important contributions of NeSy research. Firstly, we introduce study history of this area, covering early work and foundations. We further discuss background concepts and identify key driving factors behind the development of NeSy. Afterward, we categorize recent landmark approaches along several main characteristics that underline this research paradigm, including neural-symbolic integration, knowledge representation, knowledge embedding, and functionality. Next, we briefly discuss the successful application of modern NeSy approaches in several domains. Then, we benchmark several NeSy methods
on three representative application tasks.  Finally, we identify the open problems together with potential future research directions. This survey is expected to help new researchers enter this rapidly evolving field and accelerate the progress towards data-and knowledge-driven AI.
\end{abstract}
}

\maketitle
\IEEEdisplaynontitleabstractindextext
\IEEEpeerreviewmaketitle

\IEEEraisesectionheading{\section{Introduction}\label{sec1}}
\label{sec:intro}
{\IEEEPARstart{C}{urrent} advances in Artificial Intelligence (AI), especially large AI models, have caused significant changes in numerous research fields, and had profound impacts on every \redminor{aspect} of societal and industrial sectors.
At the same time, there is also growing concern in the public and scientific communities regarding the trustworthiness, safety, interpretability, and accountability of the modern AI techniques~\cite{yang2021multiple}. This leads to a natural question: \textit{\textbf{What could be the key enabler for the next generation of AI?}}

AI has historically been dominated by two paradigms: symbolism and connectionism. \textit{Symbolism} conjectures that symbols representing things in the world are the fundamental units of human intelligence, and that the cognitive~pro- cess$_{\!}$ can$_{\!}$ be$_{\!}$ accomplished$_{\!}$ by$_{\!}$ the$_{\!}$ manipulation$_{\!}$ of$_{\!}$ the$_{\!}$~sym-
 bols,$_{\!}$ through$_{\!}$ a$_{\!}$ series$_{\!}$ of$_{\!}$ rules$_{\!}$ and$_{\!}$ logic$_{\!}$ operations$_{\!}$ upon$_{\!}$ the symbolic representations~\cite{haugeland1989artificial,zhang2021neural}. Many$_{\!}$ early$_{\!}$ AI$_{\!}$ systems, from$_{\!}$ the$_{\!}$ mid-1950s$_{\!}$ to$_{\!}$ the$_{\!}$ late$_{\!}$ 1980s,$_{\!}$  were$_{\!}$ built$_{\!}$ upon$_{\!}$ sym- bolistic$_{\!}$ models.$_{\!}$ Symbolic$_{\!}$ methods$_{\!}$ have$_{\!}$ several$_{\!}$ virtues:$_{\!}$ they require only a few input samples, use powerful declarative
 languages for knowledge representation, and have
conceptually clear internal functionality. It soon became apparent, however, that such a \textit{rule-based}, \textit{top-down} strategy demands 
 substantial hand-tuning and lacks true learning. Moreover, as \textit{discrete} symbolic representations and hand-crafted rules are intolerant of ambiguous and noisy data, symbolic methods typically fall short when solving real-world problems.

\begin{figure*}[h]%
    \centering
    \includegraphics[width=0.99\textwidth]{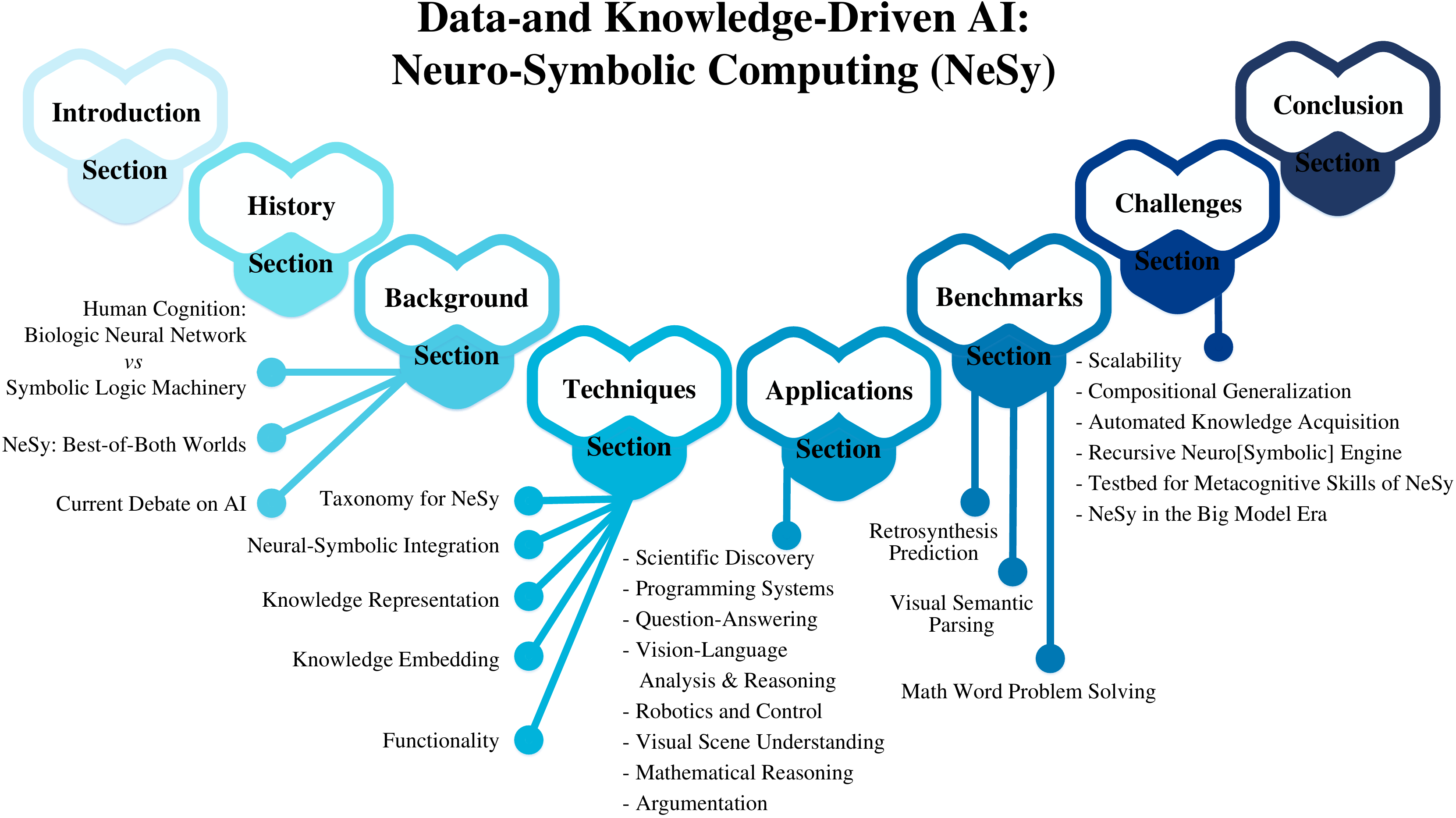}
    \put(-470, 216){\textbf{\ref{sec1}}}
    \put(-412, 183){\textbf{\ref{sec2}}}
    \put(-354, 151){\textbf{\ref{sec3}}}
    \put(-293, 118){\textbf{\ref{sec4}}}
    \put(-219, 118){\textbf{\ref{sec5}}}
    \put(-159, 151){\textbf{\ref{sec5.5}}}
    \put(-100, 184){\textbf{\ref{sec6}}}
    \put(-43,  219){\textbf{\ref{sec7}}}
       \vspace{-16pt}
    \caption{Structure of the overall review.}\label{fig1}
       \vspace{-10pt}
\end{figure*}
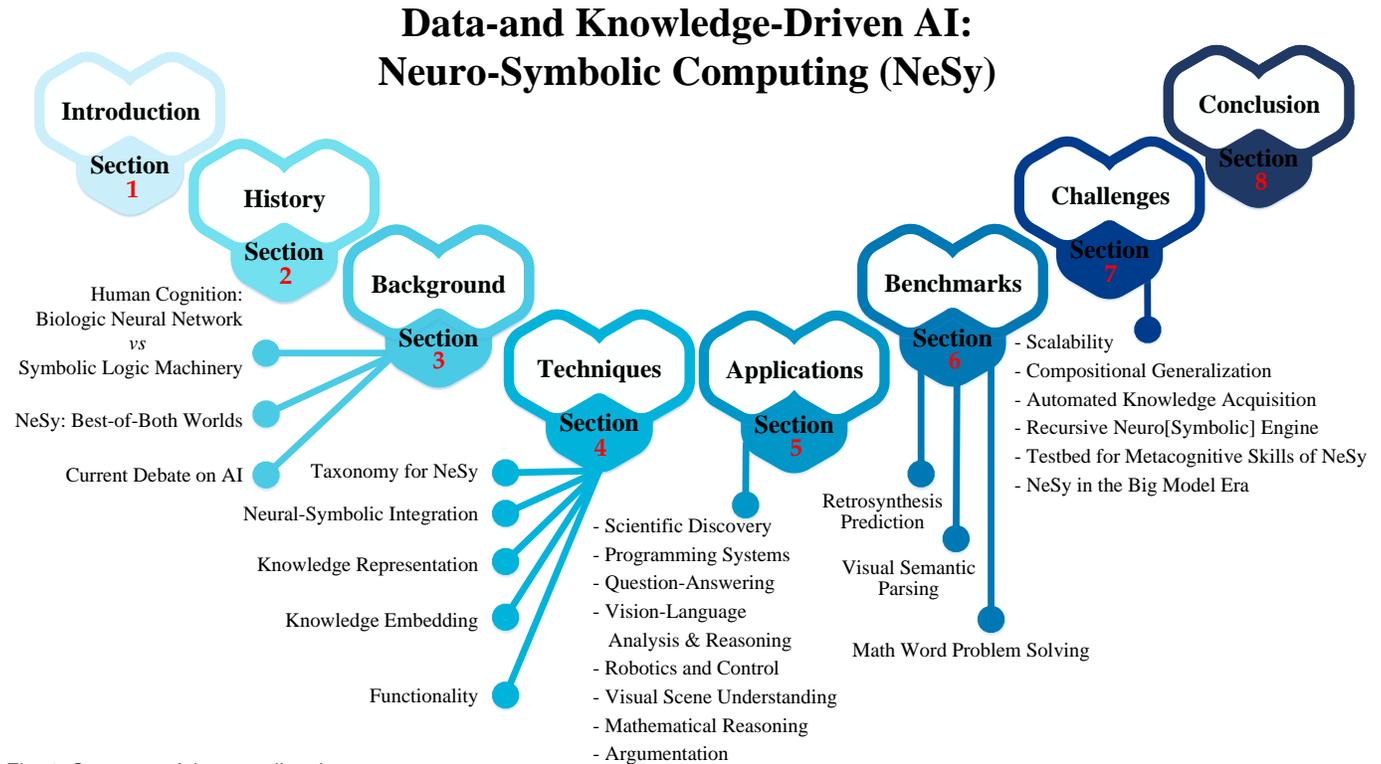

\textit{Connectionism}, represented by its most successful technique, deep neural networks (DNNs)~\cite{lecun2015deep}, serves as the architecture behind the majority of recent successful AI systems. Inspired by the physiology of the nervous system, connec- tionism$_{\!}$ explains cognition by interconnected networks of simple and often uniform units. Learning happens as weight modification, in a data-driven manner; the network weights are adjusted in the direction that minimises the cumulative error from all the training samples, using techniques such as gradient back-propagation~\cite{rumelhart1985learning}. Connectionist models are  fault-tolerant, since they learn sub-symbolics, \ie, \textit{continuous} embedding vectors, and compare these vectorized represen- tations instead of the literal meaning between entities and relations by {discrete} symbolic representations. Moreover, by learning \textit{statistical} patterns from data, connectionist models$_{\!}$ enjoy the advantages of inductive learning and generaliza-  tion capabilities. Yet, like every
coin has two sides,~such~ap- proaches$_{\!}$ also$_{\!}$ suffer$_{\!}$ from$_{\!}$ several$_{\!}$  fundamental$_{\!}$ problems$_{\!}$~\cite{smolensky2022neurocompositionala,smolensky2022neurocompositionalb}.$_{\!}$ First,$_{\!}$ connectionist$_{\!}$ models$_{\!}$ fall$_{\!}$ significantly$_{\!}$ short$_{\!}$ of$_{\!}$ \textit{compositional generalization}, the robust ability of human cognition to correctly solve any problem that is
composed of fami- liar$_{\!}$ parts$_{\!}$~\cite{fodor1988connectionism}.$_{\!}$ Second,$_{\!}$ such$_{\!}$ bottom-up$_{\!}$ approaches$_{\!}$ are$_{\!}$ known to$_{\!}$ be$_{\!}$ \textit{data$_{\!}$ inefficient}.$_{\!}$ Third,$_{\!}$ connectionist models are logically opaque, lacking \textit{comprehensibility}.
It is almost impossible to understand why decisions are made. In the absence of any kind of identifiable or verifiable train of logic, people are left with systems that make potentially catastrophic decisions that are difficult to understand, arduous to correct, and therefore hard to trust. These shortcomings hinder the adoption of connectionist systems in decision-critical applications and reasoning-heavy tasks, such as medical diagnosis, autonomous driving, and mathematical reasoning, and lead to the increasing concern about contemporary AI techniques.

Against this background, neural-symbolic computing, pioneered by combining logic and neural networks$_{\!}$~\cite{mcculloch1943logical}, and then officially introduced by the Neural-Symbolic Learning and Reasoning (NeSy) Association as a hybrid of symbolism and connectionism,~is widely recognized as an enabler of the next generation of~AI
\cite{lake2017building,marcus2018deep}. NeSy essentially looks for the integration of two 
 fundamental cognitive abilities~\cite{valiant2003three,garcez2019neural}: learning (the ability$_{\!}$  to$_{\!}$ learn$_{\!}$ from$_{\!}$ experience),$_{\!}$ and$_{\!}$ reasoning$_{\!}$ (the$_{\!}$ ability$_{\!}$ to$_{\!}$ reason from what has been learned), so as to exploit the major strengths$_{\!}$ and$_{\!}$ circumvent$_{\!}$ the$_{\!}$ inherent$_{\!}$ deficiencies$_{\!}$ of$_{\!}$ the$_{\!}$ two
 paradigms. However, building such an integrated machinery is challenging -- one has to conciliate the methodologies 
 of distinct areas$_{\!}$~\cite{garcez2022neural}, for example, statistical inductive learning$_{\!}$ based$_{\!}$ on$_{\!}$ distributed$_{\!}$ representations$_{\!}$ \textit{vs}$_{\!}$ logical$_{\!}$ deductive reasoning based on localist representations. Though challenging, NeSy has attracted soaring research attention in the recent past, and has demonstrated its superiority in many application scenarios, including visual relationship understanding$_{\!}$~\cite{donadello2017logic,zhou2021cascaded}, visual question answering$_{\!}$~\cite{andreas2016neural,mao2019neuro,amizadeh2020neuro}, visual scene parsing$_{\!}$~\cite{wang2021hierarchical,li2022deep}, and commonsense reasoning$_{\!}$~\cite{arabshahi2021conversational}.

In order to facilitate readers to catch up on the rapidly-developing evolution of this field, this paper offers a systematical and timely collection of recent important literature on NeSy, with a focus on the past five years. The surveyed papers are those works published in the flagship repositories for machine
learning and related areas, such as computer vision, {natural} language processing (NLP), and knowledge graph, or have been widely cited. This survey is expected~to {offer an exhaustive} and up-to-date literature overview to~re-
 searchers of interest, and nourish the exploration of open and$_{\!}$ developmental$_{\!}$ issues.$_{\!}$ We$_{\!}$ also$_{\!}$ remark$_{\!}$ that$_{\!}$ this$_{\!}$ survey$_{\!}$~is
 inevitably a biased view, since there is a broad spectrum~of research in this fast-growing area, but we do attempt to identify and analyze common and critical properties of land-
  mark$_{\!}$ practices$_{\!}$ in$_{\!}$ order$_{\!}$ to$_{\!}$ cover$_{\!}$ major$_{\!}$ research$_{\!}$ threads.$_{\!}$ Rea- ders$_{\!}$ are$_{\!}$ also$_{\!}$ encouraged$_{\!}$ to$_{\!}$ refer to discussions$_{\!}$ in$_{\!}$~\cite{garcez2019neural,belle2020symbolic,lamb2021graph,yu2021survey,ijcai2022p767,giunchiglia2022deep},$_{\!}$ among$_{\!}$ others,$_{\!}$ to$_{\!}$ gain$_{\!}$ a$_{\!}$ sense$_{\!}$ of$_{\!}$ the$_{\!}$ breadth$_{\!}$ of$_{\!}$ this$_{\!}$ area.$_{\!}$

A summary of the structure of this article can be found in Fig.$_{\!}$~\ref{fig1}, which {is presented as follows}: Sec.$_{\!}$~\ref{sec2} gives a brief review of early research results of NeSy, which shape the latest effort in this area. Sec.$_{\!}$~\ref{sec3} introduces the general concepts of mind in psychology and cognitive science, which underpin the theoretical foundations of NeSy, and discusses the recent debate on the necessary and sufficient building blocks of AI, which promotes the advance of this area. Sec.$_{\!}$~\ref{sec4} presents our taxonomy of NeSy,  which classifies recent important NeSy  literature according to four dimensions: neural-symbolic interrelation, knowledge representation,
knowledge embedding, and functionality. Sec.$_{\!}$~\ref{sec5} elaborates on popular and emerging application areas of NeSy. Sec.$_{\!}$~\ref{sec5.5}  conducts performance evaluation and analysis. Finally, Sec.$_{\!}$~\ref{sec6} and \ref{sec7} suggest potential valuable directions for further research and conclude the survey. We hope that this survey will help newcomers and practitioners to navigate in this massive field that gained significant momentum in the past few {years}, as
well as provide AI community with background information for generating future research.

   \vspace{-5pt}
\section{History}\label{sec2}

This section offers a historical perspective of NeSy, prior to its recent acceleration
in activity. NeSy aims {to provide} a unifying view for symbolism and connectionism, {advance} the modelling of cognition and further behaviour, and build preferable computational methodologies for integrated machine learning and logical reasoning~\cite{garcez2022neural}. NeSy has a long-standing tradition that can be traced back to McCulloch and Pitts in 1943~\cite{mcculloch1943logical}, even before AI was recognized as a new scientific field. 
For readers who are eager to obtain a more particular overview of the primitive works, we recommend consulting previous review articles, such as~\cite{bader2005dimensions,garcez2008neural,garcez2022neural}.

Although in the seminal work~\cite{mcculloch1943logical} McCulloch and Pitts established strong connection between finite automata (boolean logic) and artificial neural networks, by interpreting simple logical connectives such as conjunction,
disjunction$_{\!}$ and$_{\!}$ negation$_{\!}$ as$_{\!}$ binary$_{\!}$ threshold$_{\!}$ units$_{\!}$ in neural networks~\cite{bader2005dimensions}, NeSy only began to be a formalized field of study {since} {the} 1990s and gained systematic research in the early 2000s~\cite{shi2020neural}. For instance, Towell \etal~\cite{towell1990refinement} compiled {hand-coded} symbolic rules into a neural network, and the approximately correct knowledge can be further corrected by empirical learning. Based on some landmark efforts~\cite{pollack1990recursive,shastri1993simple,holldobler1991towards}, researchers developed various neural systems for logical inference$_{\!}$~\cite{garcez1999connectionist,garcez2008neural}$_{\!}$ and$_{\!}$ knowledge$_{\!}$ representation$_{\!}$~\cite{towell1994knowledge,plate1995holographic,cloete2000knowledge,browne2001connectionist,garcez2002neural}. As their neural architectures are mainly meticulously designed for hard logic reasoning, {they lacked the ability to learn representations and to reason over large-scale, heterogeneous, and noisy data}~\cite{shi2020neural}. Nevertheless,  these early NeSy systems laid the groundwork for today's research.

During the 2010s, NeSy received relatively less attention, as DNN-based connectionist techniques achieved remarkable success across a variety of AI tasks. However, as the shortcomings of DNNs became evident, NeSy has recently ushered in its renaissance in the research community.

   \vspace{-5pt}
\section{Background and Context}\label{sec3}
This section elucidates the two main driving forces behind the field of NeSy: The first one is the theoretical aspiration to understand and model human cognition (Sec.$_{\!}$~\ref{subsec31}), while the second one is the practical value of combining connectionism and symbolism paradigms in AI application scenario (Sec.$_{\!}$~\ref{subsec32}). Sec.$_{\!}$~\ref{subsec33} further summarizes the recent AI debate among influential thinkers, which motivates a broad range of AI researchers to recognize the significance of NeSy.

\subsection{Human Cognition: Biologic Neural Network \textit{vs} Symbolic Logic Machinery}\label{subsec31}

\noindent$\bullet$~\textbf{Symbols \textit{vs} Neurons.} What is the essence of human cognition?  Many researchers agree that symbolic facility is what distinguishes humans from other animals. The prosperity of human sociology and technology is closely concerned with the co-evolution of human brain with symbolic thinking, making us the ``symbolic species''~\cite{deacon1997co,russell2016artificial,horst2003computational}. Many cognitive scientists hold the view that human thinking relies on symbol manipulation. From this perspective, human mind is undisputedly symbolic. Therefore, symbolism was conceived in the attempts to structurally code knowledge and logic reasoning into machines. However, human cognition has a physical basis in the brain, which is composed of numerous mostly homogenous neurons. The neurons, together with the connections, or synapses, as well as diverse firing patterns among them, support different cognitive processes, such as attention, problem-solving, memory, learning, decision-making, language, perception, imagination, and logic reasoning. So it seems reasonable to assume if we can simulate the anatomy and physiology of the nervous system with artificial neurons, intelligence will be developed in computers.  This belief leads to the emergence of connectionism.

Spontaneously, in order to advance the understanding of the human mind, it appears to be reasonable to seek ways of integration of symbolic and connectionist approaches, instead of focusing on the dichotomy. In this context, artificial neural networks can be regarded as an abstraction of the
physical workings of the brain, while the symbolic logic can be viewed as an abstraction of what we introspect, when we engage in  explicit cognitive reasoning~\cite{hitzler2022neuro}. Therefore, it is of necessity {to ask how these two abstractions can be related or even unified, or how symbol manipulation can emerge from a neural substrate}~\cite{marcus2020next,garcez2022neural}.

\noindent$\bullet$~\textbf{Deduction \textit{vs} Induction.} Deductive reasoning and inductive learning arguably constitute two indispensable building blocks of human thinking, helping human to develop knowledge of the world (even though there are yet other building blocks, such as abductive reasoning)~\cite{maruyama2020symbolic}. However, their tension might {be the most fundamental issue
in areas such as philosophy, cognition, and, of course, AI}$_{\!}$~\cite{belle2020symbolic}.
The$_{\!}$ deduction$_{\!}$
camp$_{\!}$~\cite{dantsin2001complexity}$_{\!}$ is$_{\!}$ aware$_{\!}$~of$_{\!}$ {the$_{\!}$ expressiveness$_{\!}$~of formal languages for representing knowledge about the world, along with proof systems for reasoning from
such knowledge bases. The learning camp}$_{\!}$~\cite{mitchell1997machine} attempts to generalize from examples about partial descriptions of the~world \cite{belle2020symbolic}. Historically, the dichotomy between the two camps roughly divided the development of AI. 
Symbolic techniques clearly stand on the side of deductive reasoning; symbolic logic {emphasizes high-level reasoning, and sticks to structure the world in terms of objects, attributes, and relations}~\cite{belle2020symbolic,wang2024visual}.

{By contrast}, neural networks are in the statistical learning camp; they learn statistical patterns, \ie, distributed representations of entities, from data. Nevertheless, humans make extensive use of both deduction and induction in everyday
life as well as scientific investigation. We cannot precisely determine which part of human cognition is essentially symbolic, and which part is essentially statistical. Consequently, it is imperative
to {rethink the relationships between deductive
reasoning and inductive learning}, necessitating
robust computational models that are able to coordinate the symbolic essence of reasoning with the statistical nature of learning.

\noindent$\bullet$~\textbf{Compositionality \textit{vs} Continuity.} Smolensky \etal~\cite{smolensky2022neurocompositionalb} proposed to simultaneously exploit two scientific principles, which can explain the way the human brain works, for machine intelligence, from the viewpoint of the underlying computation mode of human cognition. Neurophysiological measurements suggest that information is encoded in the brain through the numerical activation levels of massive neurons, and is processed
by spreading this activation through myriad synapses of varying strengths and permanence~\cite{smolensky2022neurocompositionala}. Hence {it seems  evident that human cognition deploys neural computing~\cite{churchland1994computational}, which conforms to the \textit{Continuity Principle}: ``the encoding and processing of information are formalized with real numbers that vary continuously''~\cite{smolensky2022neurocompositionalb}. However, modern scientific studies~\cite{kiparsky1969syntactic} in philosophy and cognition suggested that all aspects of human intelligence, from language and perception to reasoning and planning, rely on a different type of computing: compositional-structure processing~\cite{janssen2012compositionality}. This type of computing follows the \textit{Compositionality Principle}~\cite{szabo2012case}: complex information is encoded in large structures which are systematically composed from smaller structures that encode simpler information. Compositionality is widely acknowledged as a core of human intelligence~\cite{pagin2010compositionality}. Our knowledge representation is naturally compositional. For example, we understand the world as a sum of its parts: objects can be broken down into pieces, events are a sequence of actions, and sentences are a series of words. {Human cognition exhibits strong compositional generalization -- the ability of reorganizing familiar knowledge components in novel ways to solve new problems, so as to handle the potentially infinite number of states of the world}~\cite{smolensky2022neurocompositionala}. Historically, {compositional-structure processing is formalized in the form of discrete symbolic computing}, like using words to make sentences. Thus, to some extent, the nature of computation in our brains is both neural and compositional-structure. How can this be?  Smolensky called this the \textit{Central Paradox of Cognition}~\cite{smolensky1988proper}. Resolving this paradox inevitably calls for a new computing mechanism, that addresses both the Continuity and Compositionality Principles simultaneously\footnote{Note that the solution -- \textit{neurocompositional computing} -- proposed by Smolensky \etal~\cite{smolensky2022neurocompositionalb} is slightly different from NeSy. NeSy broadly refers to any possible hybrid systems that couple, loosely or tightly, neural and symbolic approaches. Neurocompositional computing, instead, is to directly realize compositional-structure processing through continuous neural computing, which can be viewed as a compact, neural network based NeSy system. However, in spite of such difference, both NeSy and neurocompositional computing share the same motivation.}.

\noindent$\bullet$~\textbf{System 1 \textit{vs} System 2.} Kahneman's `\textit{fast and slow thinking}' theory, which explains the machinery of human thought, also motivated recent research interest in NeSy~\cite{booch2021thinking}. In~\cite{kahneman2011thinking}, Kahneman argued that humans' decisions are supported by the cooperation of two different kinds of capabilities, called \textit{system 1} (`fast thinking') and \textit{system 2} (`slow thinking'). Specifically, system 1 thinking is a near-instantaneous and experience-driven process for intuitive, imprecise, quick, and largely unconscious decisions, accounting for 98\% of thinking. System 1 thinking, for example, can be in the form of knowing how to zip your jacket without a second thought. Differently, system 2 thinking is slower, deliberative, and conscious, often associated with the subjective experience of agency, choice, and concentration; it provides a powerful tool for solving more complicated problems, where logical, sequential, algorithmic thinking is needed. For example, system 2 thinking is used when working on math problems. It is also worth mentioning that compositional generalization is exhibited in both system 1 thinking and system 2 thinking~\cite{smolensky2022neurocompositionalb}. Interestingly, system 2 can be viewed as a ``slave'' of system 1: when system 1 runs into difficulty, it is system 1 that decides to initiate system 2. Even during the execution of system 2, system 1 is ultimately in charge~\cite{kautz2022third}. In addition,  solutions discovered by system 2 can be readily available for later use by system 1. Thus, after a while, some problems, initially solvable only by resorting to system 2, can become manageable by system 1~\cite{booch2021thinking}. The consistent and effective use of system 2 can calibrate system 1, which, in turn, promotes system 2, leading to a feedback loop. As the characteristics of system 1 and system 2 are strikingly similar to those of the connectionist approach and the symbolic approach to AI, more and more AI researchers began to rethink the relation between the two traditions and recognize the value of NeSy.

   \vspace{-5pt}
\subsection{NeSy: Best-of-Both Worlds}\label{subsec32}
Rather than taking the motivation from the objective of achieving rational understanding and modeling of human cognition, the study of NeSy is also driven by a more technically motivated perspective -- combining numerical connectionist and symbolic logic approaches {in order to construct more powerful reasoning and learning machines for computer science applications.} The second motivation is based on the observation that connectionist techniques, especially modern DNNs, and symbolic approaches {complement each other with respect to their strengths and weaknesses.} In particular, connectionist techniques are good at discovering statistic patterns from raw data and are robust against noisy data. Hence they are effective in intuitive judgements, such as image classification. On the other hand, connectionist techniques are data hungry, and black boxes --  it is especially challenging to understand their decision-making processes. Alternatively, symbolic approaches {are excellent at principled judgements, such as logical reasoning}; they exhibit inherently high explainability and provide the ease of using powerful declarative languages for knowledge representation. Nevertheless, symbolic approaches are far less trainable and susceptible to out-of-domain brittleness. As a result, the integration of neural and symbolic approaches seems to be a natural step toward more powerful, trustworthy, and robust AI.

\newcommand{\cmark}{\ding{51}}%
\newcommand{\xmark}{\ding{55}}%

\begin{table*}
    \centering
    \caption{Summary of essential characteristics for reviewed NeSy methods (I). Please note Type 1 is presented here to show the input and output of a neural network can be made of symbols, albeit it may not qualify as NeSy from a rigorous perspective.}
           \vspace{-5pt}
    \label{table:nesy_methods}
    \begin{threeparttable}
        \resizebox{0.99\textwidth}{!}{
            \setlength\tabcolsep{16pt}
            \renewcommand\arraystretch{1.10}
            \begin{tabular}{|c|c||c|c|c|c|}
                \hline
                Neural-Symbolic & \multirow{2}{*}{Method} &Knowledge  & Knowledge&  \multicolumn{2}{c|}{Functionality}  \\
                \cline{5-6}
                Integration & &Representation & Embedding& Learning  & Reasoning  \\
                \hline
                \hline

                \multirow{3}{*}{{\color{gray}Symbolic Neuro Symbolic}}
                & {\color{gray}word2vec~\cite{mikolov2013efficient}}&{\color{gray} -} & {\color{gray}-}  & {\color{gray}$\checkmark$} &   \\
                & {\color{gray}Glove~\cite{pennington2014glove}} & {\color{gray}-} & {\color{gray}-}  & {\color{gray}$\checkmark$} &  \\
                &{\color{gray}GPT-3~\cite{brown2020language}} & {\color{gray}-} & {\color{gray}-}  & {\color{gray}$\checkmark$} &  \\
                \hline\hline
                \multirow{3}{*}{Symbolic[Neuro]}
                &AlphaGo~\cite{silver2016mastering}  & -& -  & $\checkmark$ & $\checkmark$  \\
                &NeSS~\cite{chen2020compositional} & -& -  &  & $\checkmark$  \\
                &PLANS~\cite{dang2020plans} & Programming Language & - & & $\checkmark$  \\
                &VisProg~\cite{gupta2023visual} & -& -    &$\checkmark$ &$\checkmark$  \\
                &HuggingGPT~\cite{shen2023hugginggpt}& -& -  &$\checkmark$  & $\checkmark$  \\
                &ViperGPT~\cite{surismenon2023vipergpt}& -& -  &$\checkmark$  & $\checkmark$  \\
               \hline
            \end{tabular}
        }
    \end{threeparttable}
       \vspace{-7pt}
\end{table*}

   \vspace{-5pt}
\subsection{Current Debate on AI}\label{subsec33}
Recent years have witnessed remarkable  breakthroughs in AI, brought by connectionist approaches and deep learning in particular. But researchers are also coming to realize that contemporary AI systems suffer from serious deficiencies in terms of, for example, data efficiency, comprehensibility, and compositional generalization~\cite{thompson2020computational}. This led to influential debates between famous researchers, which are about the underlying principles of AI. As a result, NeSy research gained renewed importance.

Specifically, the 2019 Montreal AI Debate between Yoshua Bengio and Gary Marcus~\cite{marcus2020next}, and the AAAI-2020 fireside conversation with Economics Nobel Laureate Daniel Kahneman and the 2018 Turing Award winners and deep learning pioneers Geoff Hinton, Yoshua Bengio, and Yann LeCun, brought new perspectives and concerns on the future of AI. In the debate between Yoshua Bengio
and Gary Marcus, Marcus emphasizes the importance of hybrid systems: ``$\cdots$~\textit{in order to get to robust artificial intelligence, we need to develop a framework for building systems that can
routinely acquire, represent, and manipulate abstract knowledge, with a focus on building systems that use that knowledge in the service of building, updating, and reasoning over complex, internal models of the external world}.'' Though Hinton agreed that ``\textit{we need those higher-level concepts to be grounded and have a distributed representation to achieve generalization}'', he also addressed that ``\textit{(numerical connectionist approaches) can get many of the attributes of symbols without the kind of explicit representations of them which has been the hallmark of classical AI}'' and that ``\textit{The reason why connectionists really wanted to depart from symbolic processing is because they thought that is wasn't a sufficiently rich kind of representation}.'' At AAAI-2020, Kahneman highlighted the importance of symbol manipulation in system 2: ``$\cdots$~\textit{as far as I'm concerned, system 1 certainly knows language $\cdots$ system 2 does involve certain manipulation of symbols}.'' Although there are disagreements about, for example, how to represent symbols in DNNs and how to achieve the hybrid of connectionism and symbolism, the thinkers, in broad strokes, are in agreement that new-generation AI systems ought to be able to handle high-level abstract concepts and to conduct sound reasoning.

\section{NeSy: Taxonomy and State of the Art}\label{sec4}
As already alluded to in the introduction, this section~is~devoted to a structured and comprehensive review of state-of-the-art$_{\!}$ NeSy$_{\!}$ algorithms.$_{\!}$ Sec.$_{\!}$~\ref{subsec41}$_{\!}$ details$_{\!}$ our$_{\!}$ taxonomy$_{\!}$~for NeSy, based on which we survey recent major research results in this area from {four} perspectives: neural-symbolic integration (Sec.$_{\!}$~\ref{subsec42}),  knowledge representation (Sec.$_{\!}$~\ref{subsec44}), knowledge embedding (Sec.$_{\!}$~\ref{subsec45}), and functionality (Sec.$_{\!}$~\ref{subsec46}). 

   \vspace{-5pt}
\subsection{Our Taxonomy for NeSy}\label{subsec41}
Our overall taxonomy for NeSy AI is mainly built upon the classification scheme proposed by Sebastian and Hitzler in 2005~\cite{bader2005dimensions}, but modified according to our specific focus and recent development tend in this field. Basically, our scheme has four main dimensions, namely \textit{neural-symbolic integration}, \textit{knowledge representation}, \textit{knowledge embedding}, and \textit{functionality}. Each dimension contains elements representing the notable properties of NeSy approaches.

The first dimension -- neural-symbolic integration -- categorizes NeSy systems according to the combination mode -- how the symbolic and neural parts are integrated as a hybrid. Along this dimension, we further adopt the classification schema recently introduced by Henry Kautz at AAAI-2020~\cite{kautz2022third}, which is influential and insightful. More details of this dimension will be given in Sec.~\ref{subsec42}.

For the second dimension -- knowledge representation, we focus on the symbolic aspect of the NeSy AI system. Depending on how the knowledge is represented, \ie, {symbolic} \textit{vs} {logic}, we can distinguish the systems, as discussed in Sec.~\ref{subsec44}.

The third dimension -- knowledge embedding -- considers at which component of the neural machine the symbolic knowledge is integrated into. We find that the integration can be made at every key part of the connectionist pipeline, namely data preprocessing, network training, network architecture, as well as final inference. Based on this insight, in Sec.~\ref{subsec45}, we make the categorization along this dimension.

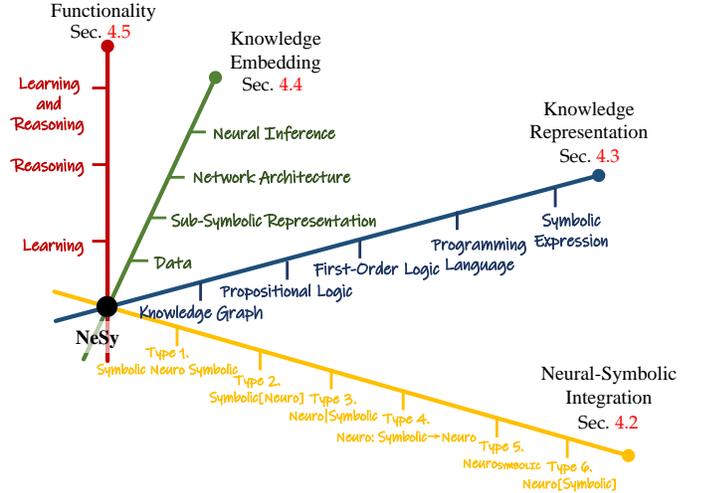
\begin{figure}[!bht]%
\centering
\includegraphics[width=0.49\textwidth]{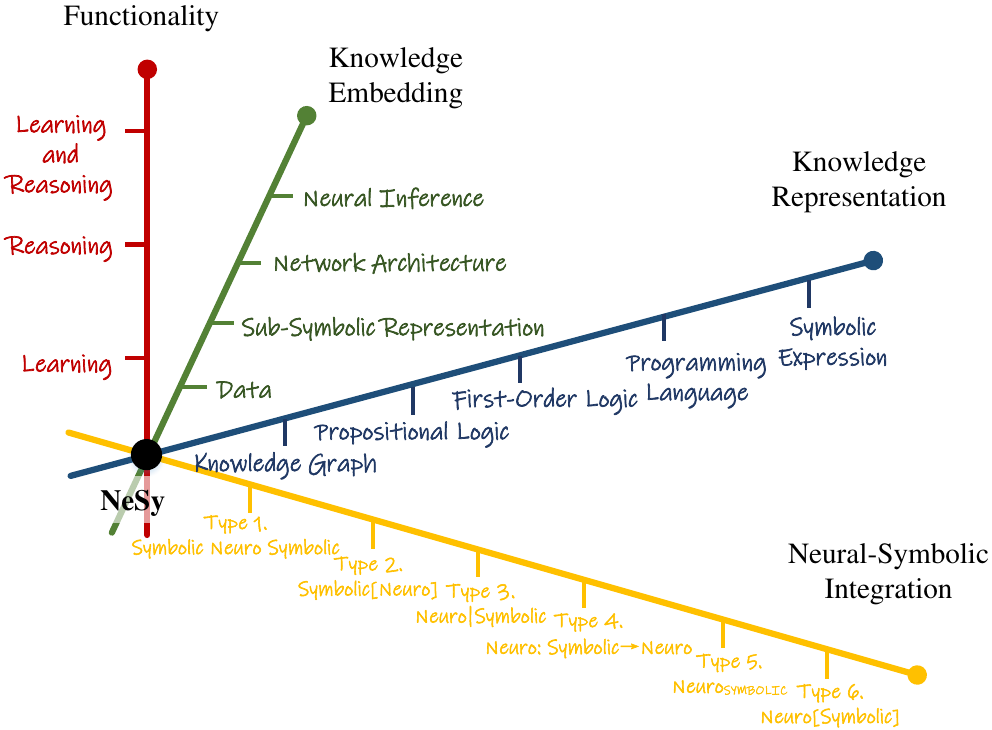}
\put(-230, 172){\scriptsize{Sec.~\ref{subsec46}}}
\put(-165, 152){\scriptsize{Sec.~\ref{subsec45}}}
\put(-45, 125){\scriptsize{Sec.~\ref{subsec44}}}
\put(-38, 24){\scriptsize{Sec.~\ref{subsec42}}}
   \vspace{-5pt}
\caption{Our four-dimensional taxonomy for NeSy. The dimensions are independent and not mutually exclusive, as explained in Sec.~\ref{subsec41}. Note that the axes indicate the order in which each section is structured, not the sequence of their evolution. For example, things can be represented by First-Order Logic that Programming Language may not be able to express, and vice versa.}\label{fig2}
   \vspace{-5pt}
\end{figure}

The forth dimension refers to the functionality of the NeSy system, namely whether it focuses more on machine learning or automated symbolic reasoning. More detailed discussions can be found in Sec.~\ref{subsec46}.

Note that the four dimensions are proposed to comprehensively describe the key characteristics of a NeSy system; they are independent and non-exclusive. Along with these four dimensions of our taxonomy,  we summarize the key features of recent remarkable works in this field in Table~\ref{table:nesy_methods}   and give detailed review below.

   \vspace{-9pt}
\subsection{Neural-Symbolic Integration}\label{subsec42}
With a good understanding of the reasons behind the need for integrating symbolic and connectionist approaches, it should turn next to the integration mode. Following the~rationale laid out in~\cite{kautz2022third}, we distinguish six types of NeSy AI systems:

\begin{figure}[!bht]%
\centering
   \vspace{-10pt}
\includegraphics[width=0.49\textwidth]{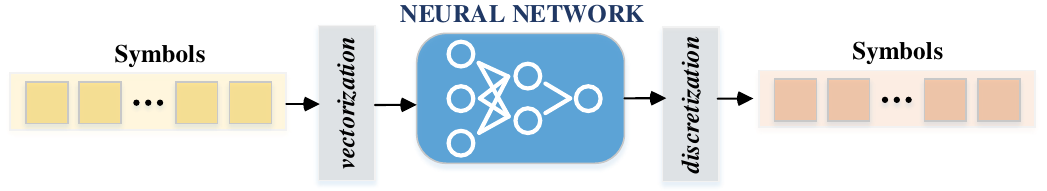}
   \vspace{-19pt}
\caption{Type~1: Symbolic Neuro Symbolic (also referred to as \textit{Neural Networks with Symbolic Input/Output}).}\label{fig2}
\end{figure}

\begin{table*}
    \centering
    \caption{Summary of essential characteristics for reviewed NeSy methods (Part II).}
           \vspace{-5pt}
    \label{table:nesy_methods2}
    \begin{threeparttable}
        \resizebox{0.99\textwidth}{!}{
            \setlength\tabcolsep{19pt}
            \renewcommand\arraystretch{1.1}
            \begin{tabular}{|c|c||c|c|c|c|c|}
                \hline
                Neural-Symbolic & \multirow{2}{*}{Method} &Knowledge & Knowledge &  \multicolumn{2}{c|}{Functionality}  \\
                \cline{5-6}
                Integration & &Representation & Embedding& Learning & Reasoning  \\
                \hline
                \hline
                \multirow{26}{*}{Neuro$\vert$Symbolic}
                &NS-VQA~\cite{yi2018neural} & Symbolic Expression & - & \checkmark &   \\
                &NSPS~\cite{parisotto2017neuro} &  Symbolic Expression & -   & \checkmark &   \\
                &NeRd~\cite{chen2019neural} & Symbolic Expression & - & \checkmark &   \\
                &Synth~\cite{nye2020learning} & Symbolic Expression& - & \checkmark &   \\
                &NSM~\cite{liang2017neural}& Symbolic Expression & -  & \checkmark &   \\
                &PS-GM~\cite{young2019learning} & - & -& \checkmark &   \\
                &NS-CL~\cite{mao2019neuro} & - & - & \checkmark &  \\
                &DSRL~\cite{garnelo2016towards} & -& - & \checkmark &  \\
                &CDSE~\cite{mou2017coupling} & - & -& \checkmark &  \\
                &NSCA~\cite{de2011neural} & Propositional Logic & -  & \checkmark &  \\
                &HOUDINI~\cite{valkov2018houdini} & Programming Language & - & \checkmark &   \\
                &PEORL~\cite{yang2018peorl} & Programming Language & - & \checkmark &  \\
                &SDRL~\cite{lyu2019sdrl} & Programming Language & -  & \checkmark &   \\
                &SORL~\cite{jin2022creativity} & Programming Language & - & \checkmark &   \\
                & RRN~\cite{hohenecker2020ontology} & First-order Logic & - & \checkmark &   \\
                &NLRL~\cite{jiang2019neural} & First-order Logic  & -& \checkmark &   \\
                &ABL~\cite{dai2019bridging} & First-order Logic & -& \checkmark & \checkmark  \\
                &Neural LP~\cite{yang2017differentiable} & First-order Logic& - &  & \checkmark  \\
                &NTPs~\cite{rocktaschel2017end} & First-order Logic& -  &  & \checkmark  \\
                &CTPs~\cite{minervini2020learning} & First-order Logic & -&  & \checkmark  \\
                &NLProlog~\cite{weber2019nlprolog} & First-order Logic& - &  & \checkmark  \\
                 &DeepProbLog~\cite{manhaeve2018deepproblog} & First-order Logic & - &  & \checkmark  \\
                &NeuroLog~\cite{tsamoura2021neural} & First-order Logic & - &  & \checkmark  \\
                &DiffLog~\cite{si2019synthesizing} & First-order Logic & -&  & \checkmark \\
                &TensorLog~\cite{cohen2016tensorlog} & First-order Logic & -&  & \checkmark  \\
                \hline
                \end{tabular}
        }
    \end{threeparttable}
        \vspace{-7pt}
\end{table*}

\noindent$\bullet$~\textbf{Type~1. Symbolic Neuro Symbolic} (also referred to as \textit{Neural Networks with Symbolic Input/Output}, Fig.~\ref{fig2}): This is, in Kautz's words, \textit{the current standard operating procedure} of deep learning methods in some application tasks where the input and output are symbols. For example, most current NLP systems, including large language models like GPT-3~\cite{brown2020language},  fall under this category (see Table~\ref{table:nesy_methods}); the input symbols are converted to vector embeddings by word2vec~\cite{mikolov2013efficient}, GloVe~\cite{pennington2014glove}, \etc,  and then processed by the neural models, whose output embeddings are further converted to the required symbolic category or sequence of symbols via a softmax operation. \redminor{This type is included by Kautz to emphasize that the input and output of a neural network can be made of symbols~\cite{garcez2020neurosymbolic}, \eg, {in the case of language translation, or graph classification.} Nowadays, some may argue that this type is a stretch to refer to as NeSy since it may encompass almost any deep learning model where the input and output spaces have clear human-interpretable semantics, \eg, tabular classification.}

\begin{figure}[!bht]%
\centering
   \vspace{-7pt}
\includegraphics[width=0.49\textwidth]{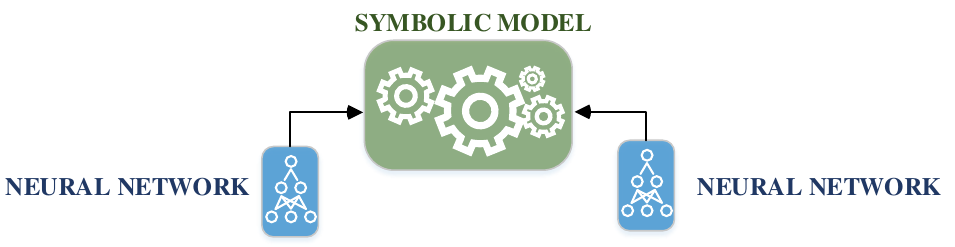}
   \vspace{-18pt}
\caption{Type~2: Symbolic[Neuro] (also referred to as \textit{Neuro Subroutines}).}\label{fig3}
   \vspace{-4pt}
\end{figure}

\noindent$\bullet$$_{\!}$~\textbf{Type~2. Symbolic$_{\!\!}$~[Neuro]} \redminor{(also referred to as \textit{Neuro Subroutines}}, Fig.~\ref{fig3}): Type~2 refers to hybrid but overall symbolic systems, where neural modules are internally used as subroutine within a symbolic problem solver. In a nutshell, the symbolic and neural parts are only loosely-coupled (see Table$_{\!}$~\ref{table:nesy_methods2}). Kautz includes DeepMind's AlphaGo$_{\!}$~\cite{silver2016mastering}$_{\!}$ as$_{\!}$ an$_{\!}$ example:$_{\!}$ {the$_{\!}$ problem$_{\!}$ solver$_{\!}$ is$_{\!}$ the$_{\!}$ Monte Carlo$_{\!}$ Tree$_{\!}$ Search$_{\!}$ algorithm$_{\!}$ and$_{\!}$ its$_{\!}$ heuristic$_{\!}$ evaluation$_{\!}$ func- 
tion is a neural network.} In$_{\!}$~\cite{chen2020compositional}, a NeSy model is designed to learn and generalize compositional rules. Based on a sequence-to-sequence (seq2seq) generation network, a symbolic stack machine is adopted for supporting recursion and sequence manipulation, and the execution trace is produced by a neural network. In$_{\!}$~\cite{dang2020plans},  a rule-based system, which uses
abstract concepts captured by a neural perception module as I/O specifications, is introduced for program synthesis from raw visual observations. Another example is recent large language model (LLM) based AI agents, \eg, VisProg$_{\!}$~\cite{gupta2023visual}, HuggingGPT$_{\!}$~\cite{shen2023hugginggpt}, and ViperGPT$_{\!}$~\cite{surismenon2023vipergpt}, which leverage LLMs to decompose a complex task into a sequence of sub-tasks that can be solved using off-the-shelf AI models.

\begin{table*}
    \centering
    \caption{Summary of essential characteristics for reviewed NeSy methods (Part III).}
           \vspace{-5pt}
    \label{table:nesy_methods3}
    \begin{threeparttable}
        \resizebox{0.99\textwidth}{!}{
            \setlength\tabcolsep{4pt}
            \renewcommand\arraystretch{1.1}
            \begin{tabular}{|c|c||c|c|c|c|c|}
                \hline
                Neural-Symbolic & \multirow{2}{*}{Method} &Knowledge  & Knowledge &  \multicolumn{2}{c|}{Functionality} \\
                \cline{5-6}
                Integration & &Representation& Embedding & Learning & Reasoning  \\
                \hline
                \hline
                \multirow{29}{*}{Neuro:\!~Symbolic}
                \multirow{29}{*}{$\rightarrow$\!~Neuro}
                & SD-VAE~\cite{dai2018syntax} & Programming Language & Data, Neural Inference& \checkmark & \\
                & Tree2tree~\cite{chen2018tree} & Programming Language &Data, Neural Inference& \checkmark & \\
                & JTAE~\cite{jin2018junction} & Knowledge Graph & Data& \checkmark & \\
                &EQNet~\cite{allamanis2017learning} & Symbolic Expression &Data& \checkmark & \\
                &DLSM~\cite{lample2019deep} & Symbolic Expression &Data & \checkmark & \\
                &NGS~\cite{li2020closed} & Symbolic Expression  &Data, Neural Inference& \checkmark & \checkmark \\
                &NeuralMath~\cite{arabshahi2018combining} & Symbolic Expression & Data & \checkmark & \\
                &IEP~\cite{johnson2017inferring} & Symbolic Expression & Sub-Symbolic Rep. & \checkmark &   \\
                &NSVR~\cite{amizadeh2020neuro} & First-order Logic & Sub-Symbolic Rep. & \checkmark &   \\
                &DrKIT~\cite{dhingra2019differentiable} & First-order Logic & Sub-Symbolic Rep. & \checkmark &   \\
                &Alps~\cite{dumancic2019learning} & First-order Logic & Sub-Symbolic Rep. &\checkmark &   \\
                &$\partial$ILP~\cite{evans2018learning} & First-order Logic & Sub-Symbolic Rep. &  & \checkmark  \\
                &NGCLP~\cite{zhang2018neural} & First-order Logic  & Sub-Symbolic Rep. & \checkmark & \checkmark\\
                &NMLNs~\cite{marra2021neural} & First-order Logic & Sub-Symbolic Rep. & \checkmark & \checkmark \\
                &NLIL~\cite{yang2019learn} & First-order Logic  & Network Architecture& \checkmark & \checkmark  \\
                &NSM~\cite{hudson2019learning} &  Knowledge Graph &Network Architecture& \checkmark & \\
                &XNMs~\cite{shi2019explainable} & Knowledge Graph &Network Architecture& \checkmark & \\
                &Prob-NMN~\cite{vedantam2019probabilistic}& Knowledge Graph &Network Architecture& \checkmark & \\
                &LENSR~\cite{xie2019embedding} & Propositional Logic, Knowledge Graph   &Sub-Symbolic Rep., Network Architecture& \checkmark &  \\
                &CNIF~\cite{wang2019learning}& Knowledge Graph & Network Architecture & \checkmark & \\
                &HHP~\cite{wang2020hierarchical} & Knowledge Graph & Network Architecture & \checkmark & \\
                 &KagNet~\cite{lin2019kagnet}& Knowledge Graph & Network Architecture & \checkmark & \\
                &GRHEK~\cite{lv2020graph}& Knowledge Graph & Network Architecture &\checkmark & \\
                &GraIL~\cite{teru2020inductive} & First-order Logic, Knowledge Graph & Data, Network Architecture& \checkmark & \\
                &KB-GAN~\cite{gu2019scene}& Knowledge Graph &Knowledge Graph& \checkmark &  \\
                &KRISP~\cite{marino2021krisp}& Knowledge Graph & Data, Network Architecture & \checkmark &  \\
                &SGN~\cite{cranmer2020discovering} & Knowledge Graph & Network Architecture & \checkmark &  \\
                &SGR~\cite{liang2018symbolic} & Knowledge Graph & Network Architecture & \checkmark & \\
                & SYMNET~\cite{garg2020symbolic} & Knowledge Graph & Network Architecture & \checkmark & \\
                &NSLMs~\cite{demeter2020just} & - & Network Architecture & \checkmark &  \\
                &\redminor{MultiplexNet}~\cite{hoernle2022multiplexnet} & First-order Logic & \redminor{Network Architecture}& \checkmark & \\
                &\redminor{C-HMCNN(h)}~\cite{giunchiglia2021multi,giunchiglia2020coherent} & Propositional Logic& \redminor{Network Architecture., Neural Inference} & \checkmark &  \\
                &\redminor{SPL}~\cite{ahmed2022semantic} & -& \redminor{Network Architecture, Neural Inference} & \checkmark & \checkmark \\
                &\redminor{CCN+}~\cite{giunchiglia2024ccn+} & \redminor{Propositional Logic}& \redminor{Network Architecture, Neural Inference} & \checkmark &  \\
                &\redminor{GeLaTo}~\cite{zhang2023tractable} & \redminor{-}& \redminor{Network Architecture, Neural Inference} &  &  \checkmark\\
                \hline
            \end{tabular}
        }
    \end{threeparttable}
       \vspace{-10pt}
\end{table*}

\noindent$\bullet$~\textbf{Type~3. Neuro$\vert$Symbolic} \redminor{(also referred to as \textit{Neural Learning + Symbolic Solver}}, Fig.~\ref{fig4}): Type~3 is also a hybrid system where the neural and symbolic parts focus on different but complementary tasks in a big pipeline. Kautz mentioned Type~3 and Type~2 ``\textit{differ in that the neuro part is a coroutine rather than a subroutine}.'' To eliminate ambiguity, here we further restrict Type~3 as the hybrid systems where the interaction between neural and symbolic parts can boost both the individual and collective performance. Therefore, the relation between neural and symbolic parts in Type~3 systems is collaboration, rather than only functional dependency in Type~2 (see Table~\ref{table:nesy_methods2}). For instance, \cite{dai2019bridging} presents an abductive learning framework, which conducts sub-symbolic perception learning and symbolic logic reasoning separately but interactively. Many deep
\begin{figure}[t]%
\centering
      \vspace{-4pt}
\includegraphics[width=0.31\textwidth]{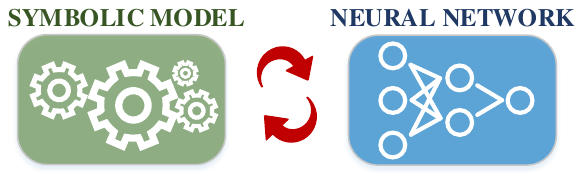}
      \vspace{-7pt}
\caption{Type~3: Neuro$\vert$Symbolic (also referred to as \textit{Neural Learning + Symbolic Solver}.)}\label{fig4}
      \vspace{-12pt}
\end{figure}
learning based program synthesis algorithms~\cite{parisotto2017neuro,chen2019neural,yi2018neural,nye2020learning,young2019learning,valkov2018houdini,yang2017differentiable} that leverage deep learning techniques to generate symbolic {programs}/rule systems satisfying high-level task specifications also fall in this category. Another notable case is~\cite{mao2019neuro}, where a neural perception module learns visual concepts and a symbolic reasoning module executes symbolic programs on the concept representations for question answering. The symbolic reasoning module provides feedback signals that support gradient-based optimization of the neural perception module. Recent efforts~\cite{yang2018peorl,garnelo2016towards,mou2017coupling,de2011neural,lyu2019sdrl,jin2022creativity,liang2017neural,jiang2019neural} in neural-symbolic reinforcement learning (RL) also belong to Type~3. For example, in~\cite{yang2018peorl}, symbolic planning are integrated into reinforcement learning (RL) for robust decision-making. Symbolic plans are used to guide task execution, and the task experiences are fed back for improving symbolic planning. Some other examples include Neural Theorem Provers (NTPs)~\cite{rocktaschel2017end}, Conditional Theorem Provers (CTPs)~\cite{minervini2020learning}, NLProlog~\cite{weber2019nlprolog}, DeepProbLog~\cite{manhaeve2018deepproblog}, NeuroLog~\cite{tsamoura2021neural}, DiffLog~\cite{si2019synthesizing}. Among them, a notable case is DeepProbLog~\cite{manhaeve2018deepproblog}, which adopts neural networks as predicates to compute the probabilities of probabilistic facts, and hence uses the inference mechanism of \redminor{ProbLog~\cite{de2007problog}}, a probabilistic logic programming language based on first-order logic, to compute the gradient of the desired loss.

\begin{figure}[t]%
\centering
\includegraphics[width=0.49\textwidth]{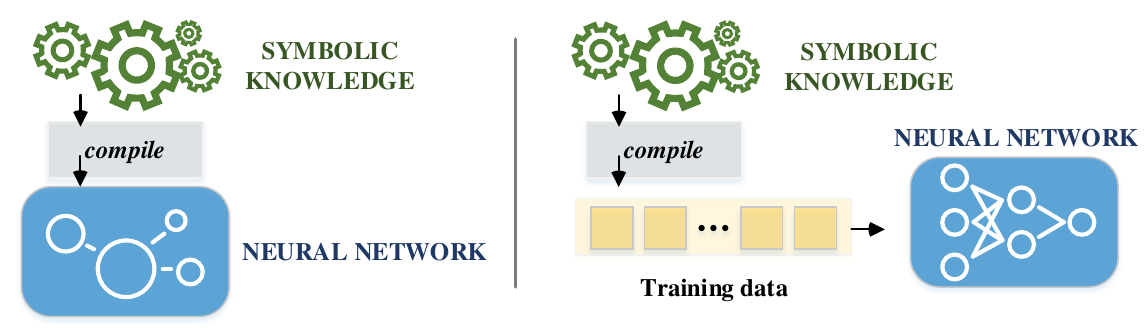}
   \vspace{-15pt}
\caption{Type~4: Neuro:\!~Symbolic\!~$\rightarrow$\!~Neuro (also referred to as \textit{Symbolic Compilation into Neural Topology}).}\label{fig5}
    \vspace{-13pt}
\end{figure}

\noindent$\bullet$~\textbf{Type~4. Neuro:\!~Symbolic\!~$\rightarrow$\!~Neuro} \redminor{(also referred to as \textit{Symbolic Compilation into Neural Topology}}, Fig.$_{\!}$~\ref{fig5}): In a TYPE~4 NeSy system,  symbolic rules/knowledge are compiled into the architecture or training regime of neural networks (see Table$_{\!}$~\ref{table:nesy_methods3}). For instance, there is a recent surge of interest in learning vector based representations of symbolic knowle- dge so as to naturally incorporate symbolic domain knowledge into$_{\!}$ connectionist$_{\!}$ architectures$_{\!}$~\cite{dumancic2019learning,allamanis2017learning,xie2019embedding,dai2018syntax,chen2018tree,jin2018junction,giunchiglia2021multi,giunchiglia2020coherent}.$_{\!}$ A  few$_{\!}$ neural-symbolic$_{\!}$ mathematics$_{\!}$ systems$_{\!}$ for$_{\!}$ equation$_{\!}$ solving$_{\!}$~\cite{lample2019deep,li2020closed} and verification~\cite{arabshahi2018combining} represent mathematical expressions as trees, which are used as training data.
A family of (visual) question answering models~\cite{johnson2017inferring,dhingra2019differentiable,evans2018learning,amizadeh2020neuro,hudson2019learning,shi2019explainable,vedantam2019probabilistic,zhang2018neural} generate and execute symbolic programs for answering questions, where the programs are implemented as fully differentiable operations and/or neural networks. \redminor{Some studies$_{\!}$~\cite{hoernle2022multiplexnet,ahmed2022semantic,giunchiglia2024ccn+} attempt to incorporate domain knowledge as logical constraints in the output layer to ensure that the output adheres to specified logical constraints, while in \cite{zhang2023tractable}, a tractable probabilistic model is proposed as the input layer to impose lexical constraints in autoregressive text generation models, \eg, GPT2~\cite{radford2019language}.}
A huge body of recent algorithms~\cite{wang2019learning,wang2020hierarchical,lin2019kagnet,lv2020graph,teru2020inductive,gu2019scene,marino2021krisp,cranmer2020discovering,liang2018symbolic,garg2020symbolic} leverage graph neural networks (GNNs) to embed entities and relations in external knowledge bases, so as to boost the performance in various applications tasks in computer vision and NLP. Broadly speaking, these methods fall into this type as suggested by Kautz, while some may argue that the reasoning ability of GNNs is rather weak.

\begin{table*}
    \centering
    \caption{Summary of essential characteristics for reviewed NeSy methods (Part IV).}
           \vspace{-5pt}
    \label{table:nesy_methods4}
    \begin{threeparttable}
        \resizebox{0.99\textwidth}{!}{
            \setlength\tabcolsep{12pt}
            \renewcommand\arraystretch{1.10}
            \begin{tabular}{|c|c||c|c|c|c|c|}
                \hline
                Neural-Symbolic & \multirow{2}{*}{Method} &Knowledge& Knowledge  &  \multicolumn{2}{c|}{Functionality}  \\
                \cline{5-6}
                Integration & &Representation & Embedding & Learning & Reasoning \\
                \hline\hline
                \multirow{19}{*}{Neuro$_{\textsc{Symbolic}}$}
                &LTN~\cite{serafini2016logic,ijcai2017p221,badreddine2022logic} & First-order Logic& Sub-Symbolic Rep. & \checkmark & \\
                & Logistic Circuits~\cite{liang2019learning} & First-order Logic& Sub-Symbolic Rep. & \checkmark & \\
                &LRI~\cite{demeester2016lifted} & First-order Logic & Sub-Symbolic Rep.& \checkmark & \\
                &HDNNLR~\cite{hu2016harnessing} & First-order Logic & Sub-Symbolic Rep.& \checkmark & \\
                &LFS~\cite{stewart2017label} & Propositional Logic& Sub-Symbolic Rep. & \checkmark & \\
                &Semantic Loss~\cite{xu2018semantic, ahmed2024pseudo} & Propositional Logic & Sub-Symbolic Rep.& \checkmark & \\
                &DANN~\cite{muralidhar2018incorporating} & Propositional Logic& Sub-Symbolic Rep. & \checkmark & \\
                &DFL~\cite{van2022analyzing} & First-order Logic & Sub-Symbolic Rep.& \checkmark & \\
                &LFIE~\cite{wang2020integrating} & First-order Logic & Sub-Symbolic Rep.& \checkmark & \\
                &ANNFL~\cite{li2019augmenting} & First-order Logic & Sub-Symbolic Rep.& \checkmark & \\
                
                &SBR~\cite{diligenti2017semantic} & First-order Logic & Sub-Symbolic Rep.& \checkmark & \\
                &LYRICS~\cite{marra2019lyrics} & First-order Logic & Sub-Symbolic Rep.& \checkmark &  \\
                &DL2~\cite{fischer2019dl2} & First-order Logic& Sub-Symbolic Rep. & \checkmark &  \\
                &HSS~\cite{li2022deep} &Knowledge Graph & Sub-Symbolic Rep., Neural Inference& \checkmark & \\
                &HMC~\cite{wehrmann2018hierarchical}&Knowledge Graph& Sub-Symbolic Rep. & \checkmark & \\
                &MBM~\cite{bertinetto2020making} &Knowledge Graph& Sub-Symbolic Rep. & \checkmark & \\
                \hline\hline
                \multirow{4}{*}{Neuro[Symbolic]}
                 &NLM~\cite{dong2018neural} & Propositional Logic & Network Architecture& \checkmark & \checkmark  \\
                &SATNet~\cite{wang2019satnet} & - & Network Architecture& \checkmark & \checkmark  \\
                &GVR~\cite{cai2017making} & - & Network Architecture& \checkmark & \checkmark  \\
                &{LogicSeg}~\cite{li2023logicseg} &First-order Logic & Sub-Symbolic Rep., Neural Inference& \checkmark & \checkmark  \\
                \hline
            \end{tabular}
        }
    \end{threeparttable}
        \vspace{-8pt}
\end{table*}

\begin{figure}[t]%
\centering
\includegraphics[width=0.49\textwidth]{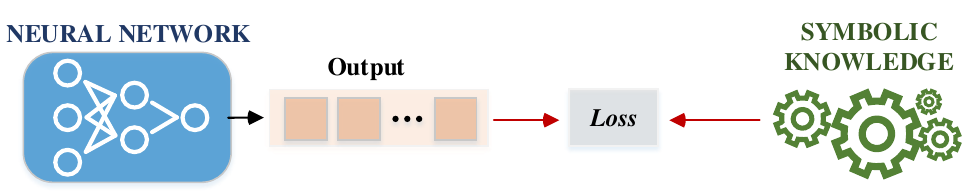}
    \vspace{-15pt}
\caption{Type~5: Neuro$_{\textsc{Symbolic}}$ (also referred to as \textit{Symbolic Integration in Loss Function}).}\label{fig6}
    \vspace{-13pt}
\end{figure}

\noindent$\bullet$~\textbf{Type~5. Neuro$_{\textsc{Symbolic}}$} \redminor{(also referred to as \textit{Symbolic Integration in Loss Function}}, Fig.~\ref{fig6}): {This type of NeSy systems turns symbolic knowledge into additional soft-constraints in the loss function used to train DNNs.} Thus, the knowledge is compiled into the weights of DNNs (see Table~\ref{table:nesy_methods4}). \redminor{Some other recent efforts in this direction include~\cite{demeester2016lifted,hu2016harnessing,stewart2017label,xu2018semantic,muralidhar2018incorporating,van2022analyzing,wang2020integrating,li2019augmenting,diligenti2017semantic,marra2019lyrics,fischer2019dl2,ahmed2024pseudo}}. \redminor{Logic Tensor Networks (LTNs)~\cite{serafini2016logic,ijcai2017p221,badreddine2022logic} and Logistic Circuits~\cite{liang2019learning},} as prominent examples, translate first-order logic formulae as fuzzy relations on real numbers for neural computing, so as to allow gradient based sub-symbolic learning. The core idea is to relax boolean first-order logic as soft fuzzy logic, which deals with reasoning that is approximate instead of fixed and exact. In fuzzy logic, variables have a truth degree that ranges in [0, 1]: zero and one meaning that the variable is \textit{false} and \textit{true} with certainty, respectively~\cite{novak2012mathematical}. LTNs approximate non-differentiable logic  connectives (\ie, $\land, \lor, \neg, \Rightarrow$), and quantifiers (\ie,  $\exists$, $\forall$)  with differentiable fuzzy logic operators~\cite{van2022analyzing}. In this way, logic rules can be embedded into network learning objective for end-to-end training. A trend of approaches consider class hierarchies when designing classifiers~\cite{wehrmann2018hierarchical,bertinetto2020making}, where the class hierarchies act as both classification targets and background knowledge. They design different training objective functions to encourage the coherence between the prediction
and the class hierarchy. For instance, in~\cite{li2022deep}, compositional relations over semantic hierarchies are cast as extra training targets for hierarchical scene parsing. \redminor{To efficiently compute losses, \cite{ahmed2024pseudo} proposes approximating the likelihood of the constraint on a local distribution centered around a model sample rather than enforcing the constraint on the entire distribution.}

\begin{figure}[t]%
\centering
\includegraphics[width=0.49\textwidth]{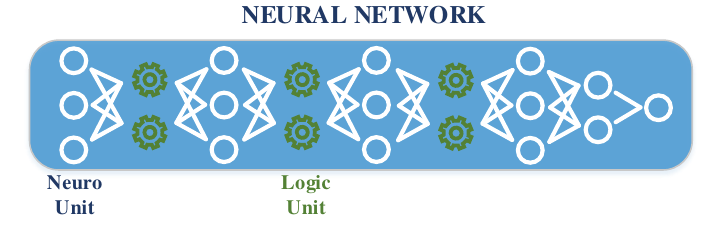}
   \vspace{-15pt}
\caption{Type~6: Neuro[Symbolic] (also referred to as \textit{Full Hybridization of Neural and Symbolic Components}).}\label{fig7}
   \vspace{-10pt}
\end{figure}

\noindent$\bullet$~\textbf{Type~6. Neuro[Symbolic]} \redminor{(also referred to as \textit{Full Hybridization of Neural and Symbolic Components}}, Fig.~\ref{fig7}): Type~6 system, which
 Kautz$_{\!}$ believes$_{\!}$ ``\textit{has$_{\!}$ the$_{\!}$ greatest$_{\!}$ potential$_{\!}$ to$_{\!}$ combine$_{\!}$ the$_{\!}$ strengths of logic-based and neural-based AI}'', are fully-integrated systems that directly embed a symbolic reasoning engine inside a neural engine.
By imitating logical reasoning with tensor calculus, a line of approaches learn the execution of symbo- lic$_{\!}$ operations$_{\!}$ through$_{\!}$ neural$_{\!}$ networks$_{\!}$~\cite{dong2018neural,wang2019satnet,cai2017making},$_{\!}$ which,$_{\!}$ to
some extent, can be classified into Type$_{\!}$~6  (see Table$_{\!}$~\ref{table:nesy_methods4}).~Yet,
their$_{\!}$ symbolic$_{\!}$ reasoning$_{\!}$ ability$_{\!}$ is$_{\!}$ still$_{\!}$ relatively$_{\!}$ weak.$_{\!}$ Kautz
views Type~6 methods as computational models of Kahneman's system~1 and system~2 and further addresses that Type~6 methods should be capable of combinatorial reasoning. From Kautz's viewpoint, it seems that there is no NeSy approach to-date that can truly meet the standard of Type~6.

    \vspace{-5pt}
\subsection{Knowledge Representation}\label{subsec44}
After clarifying and categorizing the main ways~in which symbolic and deep learning approaches are integrated together in this area, we turn next to symbolic knowledge, based on which symbol manipulation/logical calculus can be carried out. Understanding symbolic knowledge serves as the cornerstone of a NeSy system. Hence a new categorization dimension for NeSy approaches emerges purely from the perspective of how symbolic knowledge is represented. As illustrated in Fig.~\ref{fig8}, the representation approaches for symbolic knowledge can be classified into five main groups: knowledge graph, propositional logic, first-order logic,  programming language, and symbolic expression.  Hence this section is structured according to such these different categories of knowledge representations.

\begin{figure*}[!bht]%
\centering
\includegraphics[width=0.99\textwidth]{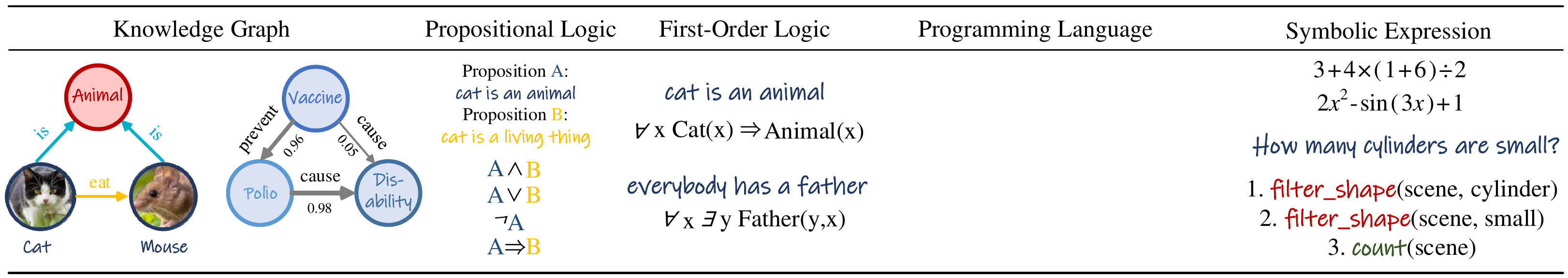}
\put(-223,62){\fontsize{5.7pt}{5.7pt}\ttfamily\selectfont{\textbf{(\textit{machine}} \color{codedefine}{\textbf{lookalgo}}}}
\put(-217,55){\fontsize{5.7pt}{5.7pt}\ttfamily\selectfont {\textbf{(\textit{state}} \color{codedefine}{\textbf{lookleft}}} }
\put(-211,48){\fontsize{5.7pt}{5.7pt}\ttfamily\selectfont {\textbf{(\textit{running}} \textbf{[robot move:[:msg|}} }
\put(-207,41){\fontsize{5.7pt}{5.7pt}\ttfamily\selectfont {\textbf{angular z: search]]))}} }
\put(-217,34){\fontsize{5.7pt}{5.7pt}\ttfamily\selectfont{\textbf{(\textit{state}} \color{codedefine}{\textbf{returnleft}}} }
\put(-211,27){\fontsize{5.7pt}{5.7pt}\ttfamily\selectfont {\textbf{(\textit{running}} \textbf{[robot move:[:msg|}} }
\put(-207,20){\fontsize{5.7pt}{5.7pt}\ttfamily\selectfont {\textbf{angular z: search negated]]))}} }
\put(-207,13){\fontsize{5.7pt}{5.7pt}\ttfamily\selectfont {\textbf{......}} }
    \vspace{-5pt}
\caption{Illustrative overview of symbolic knowledge representations in NeSy (see Sec.~\ref{subsec44}). }\label{fig8}
    \vspace{-10pt}
\end{figure*}

\noindent$\bullet$~\textbf{Knowledge Graph:} Knowledge graphs, as a popular and effective tool for knowledge representation, contain a large amount of entities and the relationships between them. Knowledge graphs are typically directed labeled graphs, formed by representing entities -- \eg, \textit{people}, \textit{places},  \textit{things} -- as nodes, and relations between entities -- \eg, ``\textit{is a friend of}'', ``\textit{is located in}'', ``\textit{is a}'' -- as edges. They contain facts that are represented as ``SPO'' triples: (Subject, Predicate, Object) where Subject and Object are entities and Predicate is the relation between them. Edges are directed from subject to object, and edge labels represent different types of relations. In an unweighted graph, all edges have the same weight. In a weighted graph, each edge is associated with a number representing its weight. The edge weight  quantifies the strength and the sign of the corresponding relationship between nodes. Please refer to the first two examples in Fig.~\ref{fig8}. A considerable body of works~\cite{hudson2019learning,shi2019explainable,vedantam2019probabilistic,xie2019embedding,wang2019learning,wang2020hierarchical,lin2019kagnet,lv2020graph,teru2020inductive,gu2019scene,marino2021krisp,cohen2020scalable,cranmer2020discovering,liang2018symbolic,li2022deep,wehrmann2018hierarchical,bertinetto2020making,demeter2020just} in computer vision and NLP fields build (weak) NeSy systems upon knowledge graphs. Their knowledge graphs  are frequently built upon our world knowledge. Since we humans understand the world by components, graphs are naturally used to represent relations between visual entities in the field of computer vision. Many famous computer vision datasets, such as~ImageNet~\cite{deng2009imagenet}, and Cityscapes~\cite{cordts2016cityscapes}, are released with structured/hierarchical annotation. For example, the ImageNet labels are organized according to WordNet~\cite{miller1995wordnet}. A family of visual parsing algorithms are developed for interpreting the \textit{part-of} (compositional) relations in common visual scenes or human-/object-centric visual stimuli~\cite{li2022deep,wang2019learning,wang2020hierarchical}. These methods fall into a broader field of machine learning, called \textit{hierarchical classification}, which is devoted to class taxonomy aware classification~\cite{wehrmann2018hierarchical,giunchiglia2020coherent,bertinetto2020making,giunchiglia2021multi}.

\noindent$\bullet$~\textbf{Propositional Logic:} Logical statements provide a flexible declarative language for formalizing knowledge about facts and dependencies, hence playing an important role for the integration of
prior knowledge into connectionist architectures. Propositional logic, also known as boolean logic or sometimes zeroth-order logic, is the simplest form of logic where all the statements are made by propositions. A pro-
position is a declarative statement that is either \textit{true} or \textit{false}. Propositional logic
studies the logical relationships between propositions which are connected via logical connectives. Typically, logical connectives (or operators), including Conjunction (``$\wedge$''), Disjunction (``$\vee$''), Negation (``$\neg$''), and Implication (``$\Rightarrow$''), are used to create compound propositions or represent a sentence logically. Propositional Logic allows for translating ordinary language statements (\ie,$_{\!}$ \texttt{IF}$_{\!}$\!~$A$\!~\texttt{THEN}  $B$) into formal logic rules ($A\!\Rightarrow\!B$). In propositional logic,  simple statements -- statements that contain no other statement as a part -- are treated as indivisible wholes. Hence, propositional logic does not deal with logical relationships and properties that involve smaller parts of statements, such as$_{\!}$ the$_{\!}$ subject$_{\!}$ and$_{\!}$ predicate$_{\!}$ of$_{\!}$ a$_{\!}$ statement.$_{\!}$ Due$_{\!}$ to$_{\!}$ the$_{\!}$ simplicity$_{\!}$ of$_{\!}$ propositional$_{\!}$ logic,$_{\!}$ many$_{\!}$ early$_{\!}$ NeSy$_{\!}$ systems,$_{\!}$~such as~\cite{towell1994knowledge,garcez1999connectionist}, consider the symbolic knowledge in the form of propositional logic. Recent work in this direction includes \cite{xu2018semantic,stewart2017label,xie2019embedding,muralidhar2018incorporating,ahmed2024pseudo}.  For instance, \cite{xu2018semantic} derives differentiable semantic loss from constraints expressed in propositional logic, for improving the performance in semi-supervised classification. In the ROAD-R dataset~\cite{giunchiglia2022jml}, logical constraints are incorporated into agents, actions and locations, which reflect real-world conditions, \eg, given an autonomous driving scenario containing a traffic light (TL), one such requirement is that a traffic light cannot simultaneously display both red and green signals. This rule can be expressed in propositional logic as: $\neg$~RedTL$~\vee$~$\neg$~GreenTL. In \cite{muralidhar2018incorporating}, domain knowledge regarding \textit{monotonic} relationships between process variables are incorporated into DNN's training. Considering a function $h(x)\!=\!y$ such that $x_1\!>x_2\!\Rightarrow\!h(x_1)\!>\!h(x_2)$. Then $x_1, x_2$ and $h(x_1), h(x_2)$ are said to share a monotonic relationship.

\noindent$\bullet$~\textbf{First-Order Logic:} Propositional logic is a finitary system that only involves a finite number of propositions and does not require sophisticated symbol manipulation operations, \ie, substitution and unification, which are needed for nested terms. Thus, it is relatively easy to implement propositional logic programs using DNNs~\cite{bader2005dimensions}. However, the expressive power of propositional logic is rather limited, since it cannot express assertions about elements of a structure.  First-order logic, also called quantified logic or predicate logic, is an extension to propositional logic and more powerful. First-order logic can express the relationship between objects by allowing \textit{variables} in predicates bound by \textit{quantifiers}. Specifically, first-order logic  augments propositional logic with two new linguistic features, \textit{viz.} variables and quantifiers. Variables are introduced to refer to objects of a certain type (\ie, domain of discourse) and can be substituted by a specific object. The universal quantifier (``$\forall$'') and existential quantifier (``$\exists$'') allow us to quantify over objects (see examples in Fig.~\ref{fig8}). A few solutions have emerged to enable DNNs to represent first-order logic. However, most of these solutions~\cite{evans2018learning,rocktaschel2017end,si2019synthesizing,yang2019learn,hohenecker2020ontology,dumancic2019learning,teru2020inductive,hu2016harnessing} can only handle restricted fragments of first-order logic. For example, some NeSy systems turn to Datalog~\cite{evans2018learning,rocktaschel2017end,si2019synthesizing,hohenecker2020ontology} for logic reasoning, or leverage GNNs to reason over local subgraph structure for inductive relation prediction~\cite{teru2020inductive}, or regularize distributed representations via domain-specific logic rules~\cite{hu2016harnessing}. To capture the full expressive power of first-order logic, some approaches~\cite{donadello2017logic,diligenti2017semantic,liang2019learning,li2023logicseg} use fuzzy logic to translate prior knowledge, expressed as a set of first-order logic clauses, into extra training objectives. In~\cite{yang2017differentiable,cohen2016tensorlog}, first-order logic rules are compiled into differentiable operations. Another group of approaches~\cite{qu2019probabilistic,zhang2019efficient,marra2020relational,marra2021neural} use first-order logic to generate a random field, based on Markov logic networks~\cite{richardson2006markov}.
Some other approaches \cite{demeester2016lifted,manhaeve2018deepproblog,weber2019nlprolog} adopt Prolog~\cite{clocksin2003programming}, a logic programming language, for knowledge representation. It shall be noted here that, due to the conflict between the infinitary nature of first-order logic -- allowing the use of function symbols as language primitives, and the finiteness of DNNs~\cite{bader2005dimensions}, it is much harder to model first-order logic in a connectionist setting compared with propositional logic.

\noindent$\bullet$~\textbf{Programming Language:} Programming language, such as logic language Prolog \cite{clocksin2003programming} and action language \textbf{$\mathcal{BC}$} \cite{lee2013action},~is a family of formal language used for writing computer~pro- grams and communicating with machines. Typically, they consist$_{\!}$ of$_{\!}$ \textit{syntax}$_{\!}$ and$_{\!}$ \textit{semantics},$_{\!}$ where$_{\!}$ \textit{syntax}$_{\!}$ represents$_{\!}$ rules that define the combinations of symbols and \textit{semantics} assigns computational meaning to valid strings formulated with respect to the \textit{syntax}. A set of NeSy methods \cite{yi2018neural,parisotto2017neuro,chen2019neural,nye2020learning,young2019learning,valkov2018houdini,lyu2019sdrl,jin2022creativity} store knowledge in programs to execute. For$_{\!}$ example,$_{\!}$ \cite{lyu2019sdrl,jin2022creativity}$_{\!}$ formulate$_{\!}$ domain$_{\!}$ knowledge$_{\!}$ in$_{\!}$ ac- tion language \textbf{$\mathcal{BC}$} to perform long-term planning, \cite{valkov2018houdini} performs a type-directed search over the library composed of parameterized programs defined in the HOUDINI language for concept reusing in other tasks. Note that for the NeSy systems that adopt Datalog or Prolog -- a subset of first-order logic$_{\!}$ --$_{\!}$ for$_{\!}$ knowledge$_{\!}$ representation,$_{\!}$ they$_{\!}$ are$_{\!}$ classified$_{\!}$~into the group of first-order logic, along the dimension of know- ledge representation.

\noindent$\bullet$~\textbf{Symbolic Expression:} Symbolic expression here roughly refers to other types of knowledge representation other than those mentioned above. Representative examples include mathematical expressions and specific symbolic sequences generated from some informal symbolic systems with self-defined rules. For instance,  in \cite{allamanis2017learning}, the source symbolic strings can be arbitrary forms of algebraic or logic expressions. In \cite{lample2019deep,li2020closed,arabshahi2018combining}, learning and reasoning are  conducted in conjunction with mathematical equations, which are typically translated into a syntax tree according to the grammatical or structural knowledge.  Apart from that, \cite{chen2019neural,nye2020learning,liang2017neural} decompose the generation of complex program into multiple predefined operators and combine them together afterwards, which improves the accuracy and can be applied to different domains by simply extending the set of symbolic operator. This kind of compositionality is also a notable case of NeSy in symbolic reasoning.

      \vspace{-8pt}
\subsection{Knowledge Embedding}\label{subsec45}
After studying how the knowledge is represented in NeSy, we next focus on the dimension of knowledge embedding, which addresses the question of where in the neural network based connectionist solutions the symbolic knowledge is embedded. Answering this question renders us a more profound understanding of the integration of symbolic knowledge and neural networks in modern NeSy systems. Our literature survey revealed that modern NeSy solutions are able to embed symbolic knowledge into training data, sub-symbolic representation, connectionist architecture, and neural inference, corresponding to the key element of the connectionist pipeline. Note that Type~1, Type~2, and Type~3 NeSy systems are not discussed here, as they either do not take symbolic knowledge into consideration (Type~1) or adopt an independent symbolic model for exploiting knowledge (Type~2 and Type~3). Whereas for Type~4, Type~5, and Type~6 NeSy systems, the knowledge can be simultaneously integrated into different components of the neural pipeline. 

\noindent$\bullet$~\textbf{Data:} A natural strategy to embed knowledge into connectionist approaches is to straightforwardly embed it in the$_{\!}$ structure$_{\!}$ of$_{\!}$ data.$_{\!}$ A$_{\!}$ prominent$_{\!}$ approach$_{\!}$ is$_{\!}$ to$_{\!}$ translate$_{\!}$ symbolic$_{\!}$ expressions,$_{\!}$ such$_{\!}$ as$_{\!}$ computer$_{\!}$ programs$_{\!}$~\cite{dai2018syntax,chen2018tree}, molecular$_{\!}$ structures$_{\!}$~\cite{dai2018syntax,jin2018junction},$_{\!}$ mathematical$_{\!}$ expressions \cite{allamanis2017learning,li2020closed},$_{\!}$ logic$_{\!}$ formulae$_{\!}$~\cite{xie2019embedding},$_{\!}$ into$_{\!}$ a structured (typically tree-/graph-organized) symbolic sequence, with respect to the$_{\!}$ corresponding$_{\!}$ grammars,$_{\!}$ semantics,$_{\!}$ and/or$_{\!}$ the$_{\!}$ relational  structure of the knowledge. The advantage of this knowledge embedding strategy is the tremendous relief~of burden on the engineering of network architecture and training objective$_{\!}$ --$_{\!}$ off-the-shelf$_{\!}$ seq2seq$_{\!}$ networks$_{\!}$ and$_{\!}$ GNNs$_{\!}$ can be directly applied. A few Type~4 NeSy systems adopt this strategy~\cite{dai2018syntax,chen2018tree,dai2018syntax,jin2018junction,allamanis2017learning,li2020closed,xie2019embedding}. However, such simple strategy has its limits for embedding complex knowledge.

\noindent$\bullet$~\textbf{Sub-Symbolic$_{\!}$ Representation:$_{\!}$} Type$_{\!}$~5$_{\!}$ NeSy$_{\!}$ systems$_{\!}$ embed symbolic knowledge into the distributed representation,  by means of training objectives that are specialized to the knowledge. A common way of building such  knowledge-specialized training objectives is to make discrete symbolic operations differentiable~\cite{li2023logicseg}. Designing appropriate loss functions for distributed encoding of symbolic knowledge is appealing as it does not require architectural change to the connectionist models or extra load of preprocessing the input data. However, it appears to be particular challenging, for example, when encoding highly abstract symbolic knowledge, such as compositional generalization \cite{parisotto2017neuro,chen2019neural,nye2020learning,johnson2017inferring}. Moreover, there is no guarantee that distributed knowledge embedding can always lead to valid outputs that are coherent with the symbolic knowledge.

\begin{figure*}[t]%
    \centering
    \includegraphics[width=0.99\textwidth]{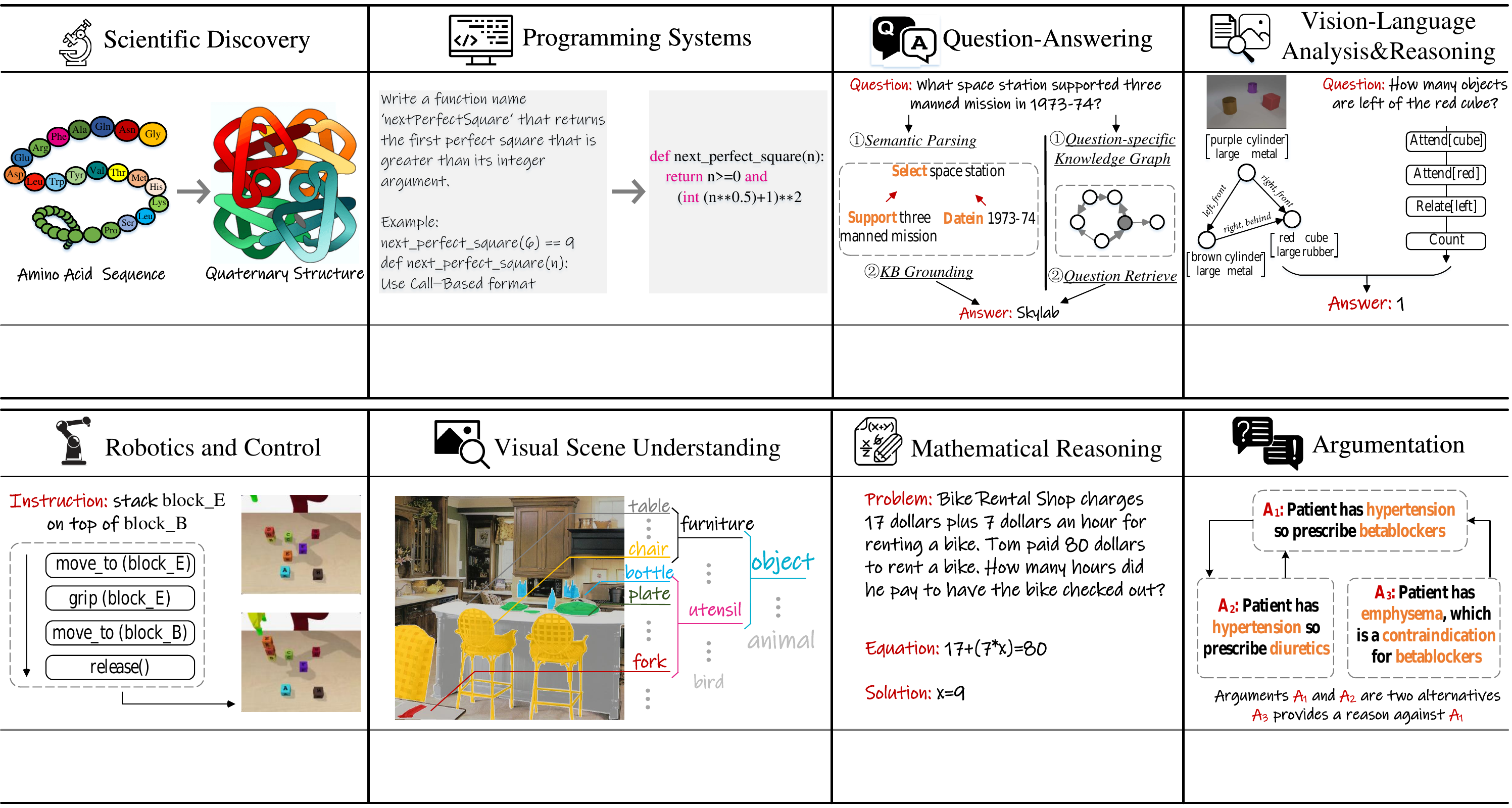}
    \put(-510,155){\fontsize{7.5pt}{7.5pt}\selectfont{\cite{segler2017neural},\cite{dai2018syntax}, \cite{cranmer2020discovering},\cite{udrescu2020ai},\cite{shah2020learning},\cite{jumper2021highly},}}
    \put(-500,145){\fontsize{7.5pt}{7.5pt}\selectfont{\cite{sacha2021molecule},\cite{zhong2023retrosynthesis},\cite{tseng2022automatic},\cite{segler2018planning},\cite{dai2019retrosynthesis}}}
    \put(-358,150){\small{\cite{murali2017neural},\cite{balog2017deepcoder},\cite{mukherjee2021neural},\cite{ellis2018learning}}}
    \put(-218,155){\small{\cite{liang2017neural},\cite{verga2021adaptable}, \cite{bosselut2021dynamic},\cite{ma2021knowledge},}}
    \put(-208,145){\small{\cite{gupta2019neural},\cite{ravishankar2022two},\cite{ye2022rng}}}
    \put(-112,155){\fontsize{6.7pt}{6.7pt}\selectfont{\cite{andreas2016neural},\cite{mao2019neuro},\cite{amizadeh2020neuro},\cite{yi2018neural},\cite{johnson2017inferring}, \cite{shi2019explainable}, \cite{vedantam2019probabilistic},}}
    \put(-110,145){\fontsize{6.7pt}{6.7pt}\selectfont{\cite{hudson2018compositional},\cite{saqur2020multimodal},\cite{andreas2016learning}, \cite{hu2017learning},\cite{mao2018neuro}, \cite{mascharka2018transparency}}}
    \put(-510,13){\small{\cite{yang2018peorl},\cite{sun2021neuro},\cite{silver2022learning},\cite{zhu2021hierarchical},\cite{xu2018neural}}}
    \put(-365,17){\small{\cite{donadello2017logic},\cite{li2022deep},\cite{xie2019embedding},\cite{wang2019learning}, \cite{wang2020hierarchical},
                   }}
    \put(-362,7){\small{
                   \cite{bertinetto2020making},\cite{li2023logicseg},\cite{gupta2023visual},\cite{shen2023hugginggpt},\cite{surismenon2023vipergpt}}}
    \put(-220,19){\fontsize{6.7pt}{6.7pt}\selectfont{\cite{rocktaschel2017end},\cite{minervini2020learning},\cite{lample2019deep},\cite{liu2019tree},\cite{xie2019goal},\cite{zhang2020graph},}}
    \put(-223,9){\fontsize{6.7pt}{6.7pt}\selectfont{\cite{paliwal2020graph},\cite{lin2021hms},\cite{qin2021neural},\cite{li2022seeking},\cite{fawzi2022discovering}, \cite{li2023softened}}}
    \put(-110,19){\fontsize{6.7pt}{6.7pt}\selectfont{\cite{atkinson2017towards},\cite{proietti2023roadmap},\cite{d2005value},\cite{garcez2014neural},\cite{cocarascu2019extracting},\cite{graves2012long},}}
    \put(-100,9){\fontsize{6.7pt}{6.7pt}\selectfont{\cite{riveret2020neuro},\cite{capobianco2011argument},\cite{carrera2015systematic},\cite{gao2016argumentation},\cite{monte2024argumentation}}}
        \vspace{-5pt}
    \caption{Summary of typical applications that have been benefited from NeSy (see Sec.~\ref{sec5}). }    \label{table:nesy_application}
        \vspace{-8pt}
    \end{figure*}

\noindent$\bullet$~\textbf{Network Architecture:} Another common way to integrate knowledge into DNNs is to design the network architecture to reflect the structure of the knowledge.  For instance, GNNs are widely adopted for capturing the complex relations in knowledge graphs and graph-structured symbolic expressions~\cite{wang2019learning,wang2020hierarchical,lin2019kagnet,lv2020graph,teru2020inductive,gu2019scene,marino2021krisp,cohen2020scalable,cranmer2020discovering,liang2018symbolic,asai2021learning,garg2020symbolic}. In~\cite{wang2019learning} and~\cite{wang2020hierarchical}, composi- tional$_{\!}$ relations$_{\!}$ between$_{\!}$ visual$_{\!}$ entities$_{\!}$ are$_{\!}$ explicitly$_{\!}$ encoded into differentiable network layers/models. Although this strategy is adopted by many Type~3 systems and is believed as the key building block for Type~6 systems, it requires significant engineering efforts in neural architecture design.

\noindent$\bullet$~\textbf{Neural Inference:} Embedding symbolic knowledge into network feedforward inference is also a feasible way, which imposes explicit constraints to force the final hypothesis to agree with the knowledge. For instance, \cite{dai2018syntax,chen2018tree,li2020closed} generate both syntactically and semantically correct predictions of molecular structures by parsing a feasible path from a tree-structured knowledge space. \cite{li2023logicseg} packages logical constraints into an iterative process and injected into the DNNs in a form of several matrix multiplications, so as to bind
logic reasoning into network feed-forward prediction. \cite{giunchiglia2021multi,giunchiglia2020coherent} ensure that no hierarchy violation happens in the hierarchical multi-label classification at inference time. \cite{ahmed2022semantic} combines exact probabilistic inference with logical reasoning, guaranteeing that predictions are always consistent with constraints during inference for structured output prediction. \cite{giunchiglia2024ccn+} applies multiple inference rules that ensure compliance with requirements, given the available knowledge about propositional logic requirements and datapoints. \cite{luetal2021neurologic,luetal2022neurologic} unveil effective algorithms using logic constraints that empower neural language models to produce articulate text while meeting intricate lexical requirements.

    \vspace{-5pt}
\subsection{Functionality} \label{subsec46}
It is clear that the ultimate goal of NeSy is to implement~a powerful$_{\!}$ AI$_{\!}$ system$_{\!}$ with$_{\!}$ combined$_{\!}$
capabilities$_{\!}$ of$_{\!}$ both$_{\!}$~data-driven learning and knowledge-driven reasoning. However, most existing NeSy systems are either good at learning or good at reasoning, but rarely both~\cite{bader2005dimensions}. To better understand the strengths and weaknesses of the systems, we examine the current NeSy systems from another dimension – core functionality. This dimension reflects whether the systems focus more on statistical learning or on symbolic reasoning.

\noindent$\bullet$~\textbf{Learning:} Some Type 3 NeSy systems, like those neural-symbolic RL approaches~\cite{yang2018peorl,garnelo2016towards,mou2017coupling,lyu2019sdrl,jin2022creativity}, and the vast majority of Type 4 and Type 5 systems usually exhibit strong learning ability, but are relatively weak at logic reasoning. For example, neural-symbolic RL approaches~\cite{yang2018peorl,garnelo2016towards,mou2017coupling,lyu2019sdrl,jin2022creativity} and visual reasoning algorithms~\cite{johnson2017inferring,dhingra2019differentiable,amizadeh2020neuro,hudson2019learning,shi2019explainable,vedantam2019probabilistic}  of Type 4 are typically limited to a small set of pre-defined and simple programs/operations, and the sequences of the programs are usually generated through DNNs. For those Type 4 systems based on knowledge embedding~\cite{wang2019learning,wang2020hierarchical,lin2019kagnet,lv2020graph,teru2020inductive,gu2019scene,marino2021krisp} or training regime modification~\cite{lample2019deep,li2020closed,arabshahi2018combining} and Type 5 systems~\cite{donadello2017logic,liang2019learning,demeester2016lifted,hu2016harnessing,diligenti2017semantic,stewart2017label,xu2018semantic,marra2019lyrics,fischer2019dl2,li2022deep} that integrate logical knowledge as additional constraints in the loss function, they pay more attention to symbolic knowledge embedding, rather than performing logic reasoning. As the symbolic knowledge is only implicitly encoded into the weights of DNNs, they struggle with explicit reasoning and their explainability is also weak.

\noindent$\bullet$$_{\!}$~\textbf{Reasoning:}$_{\!}$ Generally$_{\!}$ speaking,$_{\!}$ most$_{\!}$ Type$_{\!}$~2$_{\!}$ NeSy$_{\!}$ models \cite{chen2020compositional,dang2020plans}$_{\!}$ and$_{\!}$ a$_{\!}$ few$_{\!}$ Type$_{\!}$ 3$_{\!}$ systems$_{\!}$~\cite{rocktaschel2017end,minervini2020learning,manhaeve2018deepproblog,weber2019nlprolog,si2019synthesizing},$_{\!}$ that are built upon  statistical relational learning and logic program- ming,$_{\!}$ retain$_{\!}$ the$_{\!}$ main$_{\!}$ focus$_{\!}$ on$_{\!}$ the$_{\!}$ manipulation$_{\!}$ of$_{\!}$ the$_{\!}$ sym- bols and thus yield relatively strong reasoning ability rather than statistical learning. In particular, for the Type~2 NeSy systems~\cite{chen2020compositional,dang2020plans}, the neural part is only involved as a submodule, while the whole system acts as a symbolic model. For those logical-programming-based Type 3 systems~\cite{rocktaschel2017end,minervini2020learning,manhaeve2018deepproblog,weber2019nlprolog,si2019synthesizing}, they allow for (differentiable) logical inference over probabilistic evidence from neural
networks; however, their scalability is typically limited. \redminor{In a related vein, informed machine learning~\cite{von2021informed,dash2022review} focus in particular on how to integrate prior knowledge into DNNs. On the one hand, the findings about the  reasoning capability of NeSy methods may stimulate the potential of models learned by informed machine learning. On the other hand, the advances in informed machine learning (\eg, knowledge source, representation, and integration) offer invaluable references for the development of NeSy systems.}

\noindent$\bullet$~\textbf{Reasoning and Learning:} Despite the recent progress,~it~is still hard  to achieve a compact NeSy system that has both strong logic reasoning and expressive statistic learning abilities. In the sense of Kautz's vision, Type 6 symbolic systems could have such combined abilities. However, there are only a$_{\!}$ few$_{\!}$ models$_{\!}$~\cite{cai2017making,dong2018neural,wang2019satnet}$_{\!}$ can$_{\!}$ be$_{\!}$ barely$_{\!}$ recognized$_{\!}$ as$_{\!}$ Type$_{\!}$ 6.

\begin{table*}
    \centering
    \redminor{
    \caption{Summary of some widely-used PyPI packages for NeSy methods.}
    \label{table:pip_packge}
    \begin{threeparttable}
        \resizebox{\textwidth}{!}{
            \setlength\tabcolsep{8pt}
            \renewcommand\arraystretch{1.10}
            \begin{tabular}{|c|c|c|c|c|}
                \hline
                Package Name & Key Feature & Dependency & Quick Installation & Project Link \\
                \hline
                \hline
                \multirow{2}{*}{LTN} & \multirow{2}{*}{Differentiable first-order logic} & PyTorch & pip install LTNtorch & \url{https://pypi.org/project/LTNtorch/} \\
                & & TensorFlow & pip install ltn & \url{https://pypi.org/project/ltn/} \\
                \hline
                Pylon & Programmatic constraint specification & PyTorch & pip install pylon-lib & \url{https://pypi.org/project/pylon-lib/} \\
                \hline
                PiShield & Background knowledge injection & PyTorch & pip install pishield & \url{https://pypi.org/project/pishield/} \\
                \hline
                DeepProbLog & Probabilistic Logic Programming & PyTorch & pip install deepproblog & \url{https://pypi.org/project/deepproblog/} \\
                \hline
                KENN2 &  Universally quantified FOL clauses & TensorFlow & pip install KENN2 & \url{https://pypi.org/project/KENN2/} \\
                \hline
                PyNeuraLogic & Differentiable logic programming & Java Backend & pip install neuralogic & \url{https://pypi.org/project/neuralogic/} \\
                \hline
                torch-explain & Explainable deep learning models & PyTorch & pip install torch-explain & \url{https://pypi.org/project/torch-explain/} \\
                \hline
            \end{tabular}
        }
    \end{threeparttable}
    }
\end{table*}

\vspace{-2pt}
\section{Application Areas and Tasks}\label{sec5}

With the ambitious goal and recent rapid progress of NeSy research, various novel applications and tasks have emerged across different domains (Fig.~\ref{table:nesy_application}), such as computer vision, natural language processing, robotics, and even other scientific disciplines.    In  this section, we showcase some of the prominent application examples that illustrate the potential and impact of NeSy. However, we note that due to the high diversity of the application scenarios and the significant difference among different NeSy systems, it is not feasible to provide a comprehensive and fair evaluation framework for all the NeSy systems under a unified setting. \redminor{Furthermore, we provide a variety of popular PyPI packages that can assist in building more complex NeSy systems. (see Table\!~\ref{table:pip_packge}).}

\noindent$\bullet$~\textbf{Scientific$_{\!}$ Discovery:$_{\!}$} Scientific$_{\!}$ discovery$_{\!}$ typically$_{\!}$ requires

\noindent algorithms that discover scientific hypotheses or concepts from data, and respect physical constraints and domain-specific knowledge that are known to hold in the world. Moreover, the algorithms should better interpret and explain how they come up with their solutions and convey their insights to human scientists~\cite{krenn2022scientific}. As a result, scientific discovery poses great challenges for current pure data-driven AI techniques, yet serves as a good testbed for NeSy.

Several$_{\!}$ recent$_{\!}$ studies$_{\!}$ showed$_{\!}$ the$_{\!}$ extraction$_{\!}$ of$_{\!}$~symbolic models from experimental data of mechanical systems$_{\!}$~\cite{udrescu2020ai} and$_{\!}$ in$_{\!}$ astronomy$_{\!}$~\cite{cranmer2020discovering}.$_{\!}$ For$_{\!}$ instance,$_{\!}$ the$_{\!}$ authors$_{\!}$ of$_{\!}$~\cite{cranmer2020discovering}$_{\!}$ apply a symbolic regression technique to a GNN model that is trained on cosmological dark matter data, and demonstrate that explicit physical relations can be discovered in the form of analytic formula.  For protein structure prediction, \cite{jumper2021highly} applies the gradient descent algorithm to uncover the graph structure of proteins in 3D space, where the edges of the graph are determined by the proximity of residues. Moreover, some recent works \cite{shah2020learning,tseng2022automatic} employ neuro-symbolic programming to analyze the behavior of laboratory animals, such as classifying sequential animal behaviors, clustering animal behaviors in an interpretable way, and representing expert knowledge in a reusable domain-specific language and more general domain-level labeling functions. In addi- tion,$_{\!}$ NeSy$_{\!}$ techniques$_{\!}$ are$_{\!}$ applied$_{\!}$ to$_{\!}$ retrosynthesis$_{\!}$ and$_{\!}$~reac- tion$_{\!}$ prediction$_{\!}$ in$_{\!}$ organic$_{\!}$ chemistry~\cite{segler2017neural,sacha2021molecule,zhong2023retrosynthesis}.$_{\!}$ For$_{\!}$ example,$_{\!}$ \cite{segler2018planning}$_{\!}$ combines$_{\!}$ Monte$_{\!}$ Carlo$_{\!}$ tree$_{\!}$ search$_{\!}$ with$_{\!}$ an$_{\!}$ expansion$_{\!}$ policy$_{\!}$~network to discover retrosynthetic routes. \cite{dai2018syntax} adds syntax~and semantics checking during molecule synthesis. \cite{dai2019retrosynthesis} devises a conditional graph logic network to learn when to apply rules$_{\!}$ from$_{\!}$ reaction$_{\!}$ templates,$_{\!}$ implicitly$_{\!}$ considering$_{\!}$ both$_{\!}$ the chemical and strategic feasibility of the resulting reaction.

\noindent$\bullet$~\textbf{Programming Systems:} Program synthesis is another important application domain of NeSy. The goal is to automati- cally generate programs from high-level task specifications. The$_{\!}$ specifications$_{\!}$ are$_{\!}$ typically$_{\!}$ hard$_{\!}$ logical$_{\!}$ constraints,$_{\!}$~for example,$_{\!}$ test$_{\!}$ cases$_{\!}$ that need to be satisfied exactly, pre-postcondition pairs, or temporal logic formulas. The pro- grams$_{\!}$ are$_{\!}$ structured,$_{\!}$ symbolic$_{\!}$  expressions$_{\!}$ that$_{\!}$ follow$_{\!}$ the$_{\!}$ syntax of a domain-specific language. NeSy tools are more suitable for this domain than purely neural alternatives, by virtue of their modularity and use of symbolic primitives.

To combine neural learning with the formal, logic constraints of programming languages, NeSy based programmers~\cite{murali2017neural,balog2017deepcoder,ellis2018learning,mukherjee2021neural} typically learn input and context-specific heuristics from large-scale data and use such heuristics to guide search-based symbolic methods to guarantee soundness. For example, given some tokens that indicate the desired program functionality, such as API calls, types, or keywords, \cite{murali2017neural} generates strongly typed Java-like source code in two steps. First, it learns neural models to produce sketches of programs, which$_{\!}$ are$_{\!}$ abstract$_{\!}$ representations$_{\!}$~of program syntax that omit low-level details. Second, it concretizes the sketches into type-safe programs using a combinatorial search procedure.  \cite{ellis2018learning} utilizes a combination~of DNNs and stochastic search to parse drawings into symbolic specifications; these specifications are then fed into a general-purpose program synthesis engine to infer a structured graphics program. \cite{mukherjee2021neural}  introduces the concept of neurosymbolic attribute grammars, which combine a stochastic context-free grammar with semantic attributes computed by static program analysis. The neural network learns to condition its generation actions on these attributes, which provide useful semantic clues and long-distance dependencies.

\noindent$\bullet$~\textbf{Question-Answering:} Question-answering (QA) is a long-standing AI task that aims to build intelligent systems~that can automatically answer questions from humans in natural$_{\!}$  language,$_{\!}$  typically$_{\!}$  with$_{\!}$  the$_{\!}$  aid$_{\!}$  of$_{\!}$  a$_{\!}$  knowledge$_{\!}$ source$_{\!}$  com- posed of unstructured text corpora and/or structured concepts. Answering \textit{complex} questions that~involve$_{\!}$ multiple$_{\!}$ subjects,$_{\!}$ compound$_{\!}$ relations,$_{\!}$ and$_{\!}$ numerical$_{\!}$ operations is a grand challenge in QA. To address this challenge, NeSy based QA systems have been recently developed, typically following either a \textit{semantic parsing} paradigm or a \textit{knowledge embedding}  paradigm. Semantic parsing-based methods~\cite{liang2017neural,gupta2019neural,ravishankar2022two,ye2022rng} learn to translate a question into a symbolic~logic form by conducting semantic and syntactic analysis and then derive the answer by executing the parsed logic form against$_{\!}$ the$_{\!}$ knowledge$_{\!}$ source.$_{\!}$ For$_{\!}$ example,$_{\!}$ \cite{liang2017neural} adopts a neural seq2seq model that maps language utterances to  programs
and utilizes a key-variable memory to save and~reuse intermediate$_{\!}$ execution$_{\!}$ results$_{\!}$ for$_{\!}$ supporting$_{\!}$ language$_{\!}$ compositionality and complex semantics. Then a symbolic Lisp interpreter is used to perform program execution over the knowledge source, and helps find good programs by pruning the search space. Knowledge embedding-based methods~\cite{bosselut2021dynamic,verga2021adaptable,ma2021knowledge} learn and store neural representations of the knowledge source and then retrieve the answers from the stored neural form of the knowledge source considering the information conveyed in the questions. For example,  \cite{verga2021adaptable} builds a \textit{fact memory} that encodes the entities of a knowledge source as numerical vectors and provides a contextualized reference for a neural language model to create answers. Overall, semantic parsing based NeSy QA systems can produce a more interpretable reasoning process by generating expressive logic forms. However, they heavily rely on the design of the logic form and parsing algorithm, which turns out to be
the bottleneck of performance improvement. As a comparison, knowledge embedding based NeSy QA systems enjoy more benefits of end-to-end training but lack traceable reasoning.

\noindent$\bullet$~\textbf{Vision-Language Analysis and Reasoning:}  NeSy techni- ques have been also successfully applied to vision-language analysis and reasoning tasks (\eg, visual question-answering (VQA) and visual grounding), which often require comprehensive understanding and reasoning over both the visual and linguistic modalities. VQA is a challenging task that is concerned with answering questions based on visual content. Existing NeSy based VQA systems focusing on parsing questions and visual scenes into structured representations for cross-modality reasoning. For textual semantic parsing, Andreas \etal~\cite{andreas2016neural} proposed a Neural Module~Network that interprets questions as executable programs composed of learnable neural modules that can be directly applied to images.  A module is typically implemented by the neural attention operation and corresponds$_{\!}$ to$_{\!}$ a$_{\!}$ certain$_{\!}$ atomic$_{\!}$~rea- soning$_{\!}$ step,$_{\!}$ such$_{\!}$ as$_{\!}$ \textit{recognizing$_{\!}$ objects},$_{\!}$ \textit{classifying$_{\!}$ colors},$_{\!}$ \etc.$_{\!}$ This$_{\!}$ pioneering$_{\!}$ work$_{\!}$ inspired$_{\!}$ many$_{\!}$ subsequent$_{\!}$ studies$_{\!}$ \cite{andreas2016learning,hu2017learning,johnson2017inferring,hudson2018compositional,mao2018neuro,mascharka2018transparency,yi2018neural,vedantam2019probabilistic,amizadeh2020neuro}. In the context of visual semantic parsing, a few NeSy based VQA systems~\cite{hudson2019learning,shi2019explainable,mao2019neuro,saqur2020multimodal} utilize scene graphs as structured, symbolic representations of visual scenes, and derive answers by graphical reasoning. As for visual grounding, which studies how to localize objects in visual scenes based on language descriptions, Hsu \etal~\cite{hsu2023ns3d} combine large language-to-code models with modular neural networks to parse natural language into symbolic programs for 3D visual reasoning.

\noindent$\bullet$~\textbf{Robotics$_{\!}$ and$_{\!}$ Control:$_{\!}$} Robots$_{\!}$ are$_{\!}$ complex$_{\!}$ systems$_{\!}$ with mechanical elements and controllers. To build~an~autono- mous embodied$_{\!}$ system,$_{\!}$ we$_{\!}$ are$_{\!}$ supposed$_{\!}$ to$_{\!}$ design$_{\!}$ suitable policies that ensure the system operates within reasonable mechanical constraints. Moreover, safety and data efficiency are also crucial for constructing the system. So far, many NeSy based autonomous systems~\cite{yang2018peorl,xu2018neural,sun2021neuro,zhu2021hierarchical,silver2022learning} have been developed, where high-level, symbolic planning is generated to guide low-level RL for task execution  and$_{\!}$ learning.$_{\!}$ They$_{\!}$ recycle$_{\!}$ the$_{\!}$ common$_{\!}$ practice$_{\!}$ in$_{\!}$ this$_{\!}$ field that$_{\!}$ decision-making$_{\!}$ in$_{\!}$ robotics$_{\!}$ environments$_{\!}$ can$_{\!}$ be$_{\!}$ decomposed into a high level (\ie, \textit{what to do}) and a low level~(\ie, \textit{how to do it}). Another major source of their idea can be traced back to the notion of hierarchical RL \cite{ryan2002using,sutton2018reinforcement} which seeks to impose the task structure onto the learned policy. In particular, \cite{silver2022learning} learns parameterized polices in combination with operators and samplers, which are packaged into modular neuro-symbolic skills and easily reused in new tasks. \cite{zhu2021hierarchical} combines geometric and symbolic scene graphs as a two-level abstraction of manipulation scenes and leverages GNNs for predicting high-level task plans and low-level motions. In \cite{sun2021neuro}, a program search method is proposed for autonomous driving decision module design, where differentiable neuro-symbolic programs that specify all the behaviors for reactive and deliberative autonomous driving are synthesized and can be end-to-end trained with the whole system. As a result, more interpretable and transparent decision-making process can be delivered.

\noindent$\bullet$~\textbf{Visual Scene Understanding:} In the context of interpreting high-level semantics from visual perception, there are a set of NeSy models~\cite{donadello2017logic,li2022deep,wang2019learning,xie2019embedding,wang2020hierarchical,bertinetto2020making,li2023logicseg}~that  seek to exploit external symbolic knowledge regarding the relations between visual semantics and structured properties of novel objects to improve the robustness and performance. For example, in LTN~\cite{donadello2017logic}, \textit{part-of} relations between objects are formalized in the form of first-order logic, and converted into differentiable training objectives for end-to-end object classification learning. In HSS~\cite{li2022deep}, semantic concepts and their complex meronymy relations are organized into a tree/directed acyclic graph, from which a set of con-
 strains are derived for boosting network training for hierar- chical semantic segmentation. In~\cite{xie2019embedding}, symbolic knowledge about visual relations are expressed as propositional~formu- lae and embedded onto a manifold via a GNN. The authors$_{\!}$
 also introduce semantic regularization and heterogeneous node embedding to enhance the semantic fidelity and expressiveness of the embeddings for visual relation understanding. Li \etal \cite{li2023logicseg} proposed LogicSeg, a NeSy based visual semantic parser that formalizes the complex \textit{meronymy} and \textit{exclusion} relations among symbolic concepts as first-order logic rules. After fuzzy logic-based continuous relaxa-
 tion, the logical formulae are grounded onto data and neural computational graphs for end-to-end network training and encapsulated into an iterative optimization process for network feed-forward inference. Recently, LLM based AI agents \cite{gupta2023visual,shen2023hugginggpt,surismenon2023vipergpt} are developed for solving complex$_{\!}$ vision$_{\!}$
tasks.$_{\!}$ They$_{\!}$ employ$_{\!}$ LLMs$_{\!}$ to$_{\!}$ automatically$_{\!}$ crate$_{\!}$ plans and$_{\!}$ execute$_{\!}$ the$_{\!}$ plan$_{\!}$ by$_{\!}$ systematically invoking external tools (\eg, off-the-shelf specialized models) to get the solution.

\noindent$\bullet$~\textbf{Mathematical Reasoning:} As a distinct and specialized capability inherent in humans, mathematical reasoning has garnered substantial attention within AI community. This multifaceted skill encompasses linguistic reasoning, visual reasoning, common sense reasoning, logical reasoning, numerical reasoning, and symbolic reasoning~\cite{faldu2021towards}. Human approach to understanding and solving mathematical problems is not primarily rooted in experience and evidence, but on the basis of learning, inferring, and applying laws, axioms, and symbolic manipulation rules~\cite{saxton2018analysing}. The structured and reasoning-heavy nature of mathematical problems enables the construction of NeSy based solvers~\cite{li2023softened}. For mathematical problem solving, \cite{liu2019tree,xie2019goal} introduced tree structured decoder to explicitly explore the abstract syntax tree of mathematical expressions, and stimulated many follow-up efforts~\cite{zhang2020graph,qin2021neural,lin2021hms,li2022seeking}. For theorem proving, \cite{paliwal2020graph} considers syntax trees of formulas as graphs and apply message-passing for higher-order proof search. For handwritten formula evaluation, \cite{li2023softened} models the symbol solution states as a Boltzmann distribution, avoiding expensive state searching and facilitating mutually beneficial interactions between network training and symbolic reasoning. For discovering faster matrix multiplication algorithms, \cite{fawzi2022discovering} makes use of a Monte Carlo tree search planning pro- cedure, aided by DNNs.

\noindent\redminor{$\bullet$~\textbf{Argumentation:} Argumentation, a distinct cognitive capacity of humans, plays a pivotal role in managing diverse mental attitudes and navigating situations characterized by incomplete or inconsistent information~\cite{atkinson2017towards}. The argumentation framework, which represent arguments and the relationships between them as a directed graph, excels in capturing and generalizing various forms of non-monotonic reasoning~\cite{dung1995acceptability,bondarenko1997abstract}, essential for dealing with rules and exceptions that give rise to logical conflicts~\cite{proietti2023roadmap}. The combination of this framework with the powerful computational capability of the neural networks has led to advancements in computational argumentation. While early research in this field was predominantly built upon symbolistic models~\cite{bench2003persuasion}, recent NeSy based methods have made significant strides. These include translating argumentation networks into neural networks~\cite{d2005value}, exploring correspondences between argumentation frameworks and neural networks~\cite{garcez2014neural}, and using networks to extract argumentation frameworks for subsequent symbolic and argumentative reasoning, \eg, ADA~\cite{cocarascu2019extracting} mines argumentation frameworks from text reviews and reason with them to provide movies recommendations. Furthermore, reasoning over argumentation frameworks can be integrated into neural networks as a form of inductive bias, as exemplified by argumentation Boltzmann machines~\cite{riveret2020neuro}. Recent developments in this field have also explored explanation robustness, such as counterfactual explanations~\cite{stepin2021survey,jiang2023formalising} in neural networks, and multi-agent argumentation systems~\cite{capobianco2011argument,carrera2015systematic,gao2016argumentation,monte2024argumentation}, which combine argumentation frameworks with distributed reasoning for decision-making.}

    \vspace{-5pt}
\section{Performance Comparison}\label{sec5.5}
\redminor{NeSy has contributed to improving performance across various tasks, \eg, generating and refining samples in semi- and weakly-supervised scenarios~\cite{liang2023logic,zhou2018brief}, enhancing structured prediction for reasoning-heavy tasks~\cite{xie2019goal,zhong2023retrosynthesis}, and enforcing hard constraints in networks for semantic parsing~\cite{li2023logicseg}.}
To offer more empirical insights, in this section we tabulate 
the performance of some of the NeSy algorithms discussed before. As NeSy has become a quite broad research field that covers various application tasks, and the algorithmic design is often highly customized for each task, it is infeasible to compare all the NeSy algorithms on a common task or dataset. Therefore, we select three representative tasks~of NeSy, based on our review in Sec.~\ref{sec5}, for performance evaluation.   The performance scores are either obtained from our own implementation or collected from the original papers.

\begin{table}[t]
    \centering
    \caption{\textbf{Quantitative retrosynthesis prediction results} (\S\ref{sec5.51}) on {USPTO-50K}$_{\!}$~\cite{schneider2016s} \texttt{test}, in terms of Top-\textit{k} Exact match accuracy. (The three best scores are marked in \textcolor{red}{\textbf{red}}, \textcolor{blue}{\textbf{blue}}, and \textcolor{green}{\textbf{green}}, respectively.  These notes also apply to the other tables.)}
        \vspace{-5pt}
    \small
    \resizebox{0.99\linewidth}{!}{
       \setlength\tabcolsep{4pt}{}
       \renewcommand\arraystretch{1.05}
       \begin{tabular}{z{73}y{35}|c||cccccc}
          \hline\thickhline
          \rowcolor{mygray}
          \multicolumn{2}{c|}{Method} &NeSy & \textit{k}\!~=\!~1 &  3 & 5 & 10 & 20 & 50  \\ \hline\hline
          Seq2seq~\cite{liu2017retrosynthetic}\!&\!\!\!\pub{Chem~Eur~J17}\!& &  37.4 & 52.4 & 57.0 & 61.7 & 65.9 & 70.7  \\
          GTA~\cite{seo2021gta}\!&\!\!\!\pub{AAAI21}\!& & 51.1 & 67.6 &  74.8 &  81.6 & - & -  \\
          RetroPrime~\cite{wang2021retroprime}\!&\!\!\!\pub{Chem~Eng~J21}\!& &  51.4 & \textcolor{green}{\textbf{70.8}} & 74.0 &  76.1 & - &  -  \\
          Dual-TF~\cite{sun2021towards}\!&\!\!\!\pub{NeurIPS21}\!& &\textcolor{blue}{\textbf{53.3}} &  69.7 & 73.0 &  75.0 & - & -  \\
          GraphRetro~\cite{somnath2021learning}\!&\!\!\!\pub{NeurIPS21}\!& & \textcolor{green}{\textbf{53.7}} & 68.3 & 72.2 &  75.5 & - & -  \\
          NSR~\cite{segler2017neural}\!&\!\!\!\pub{Chem~Eur~J17}\!&$\checkmark$& 44.4 & 65.3 & 72.4 & 78.9 & \textcolor{blue}{\textbf{82.2}} & 83.1  \\
          MEGAN~\cite{sacha2021molecule}\!&\!\!\!\pub{JCIM21}\!&$\checkmark$& 48.1 & \textcolor{blue}{\textbf{70.7}} & \textcolor{green}{\textbf{78.4}} & \textcolor{green}{\textbf{86.1}} & \textcolor{red}{\textbf{90.3}} & \textcolor{red}{\textbf{93.2}}  \\
          GLN~\cite{dai2019retrosynthesis}\!&\!\!\!\pub{NeurIPS19}\!&$\checkmark$& 52.5 & 69.0 & \textcolor{blue}{\textbf{75.6}} & \textcolor{blue}{\textbf{83.7}} & \textcolor{green}{\textbf{89.0}} & \textcolor{blue}{\textbf{92.4}}  \\
          Graph2Edits~\cite{zhong2023retrosynthesis}\!&\!\!\!\pub{Nat~Commun23}\!&$\checkmark$& \textcolor{red}{\textbf{55.1}} & \textcolor{red}{\textbf{77.3}} & \textcolor{red}{\textbf{83.4}} & \textcolor{red}{\textbf{89.4}} & - & \textcolor{green}{\textbf{92.7}}  \\
          \hline
       \end{tabular}
    }
    \label{table:retro}
 \vspace{-1.em}
 \end{table}
 
    \vspace{-5pt}
\subsection{Performance$_{\!}$ Benchmarking:$_{\!}$ Retrosynthesis Prediction}\label{sec5.51}
As a fundamental task in organic synthesis, retrosynthesis prediction aims to predict the reactants given a core product.

\noindent$\bullet$~\textbf{Dataset:$_{\!}$} We adopt the widely-used retrosynthesis prediction dataset USPTO-50K \cite{schneider2016s} for evaluation. USPTO-50K comprises about $50,000$ reactions with precise atom mappings between reactants and products. The 80\%/10\%/10\% of the total $50$K reactions are set as \texttt{train}/\texttt{val}/\texttt{test} data. For fair comparison, all the experiments are conducted without knowing the reaction class in advance.

\noindent$\bullet$~\textbf{Benchmarking Algorithms:} For thorough assessment, we involve four Nesy based retrosynthesis algorithms (\ie, NSR$_{\!}$~\cite{segler2017neural},$_{\!}$ GLN$_{\!}$~\cite{dai2019retrosynthesis},$_{\!}$ MEGAN$_{\!}$~\cite{sacha2021molecule},$_{\!}$ Graph2Edits$_{\!}$~\cite{zhong2023retrosynthesis}), as well as five purely neural methods (\ie, Seq2seq$_{\!}$~\cite{liu2017retrosynthetic},$_{\!}$ GTA \cite{seo2021gta},$_{\!}$ RetroPrime$_{\!}$~\cite{wang2021retroprime},$_{\!}$ Dual-TF$_{\!}$~\cite{sun2021towards},$_{\!}$ GraphRetro$_{\!}$~\cite{somnath2021learning}).

\noindent$\bullet$~\textbf{Evaluation$_{\!}$ Metric:$_{\!}$} As$_{\!}$ standard,$_{\!}$ Top-\textit{k}$_{\!}$ exact$_{\!}$ match$_{\!}$ accuracy is used as the evaluation metric. It is computed as the ratio that one of the Top-\textit{k} predicted results exactly match the ground truth, where \textit{k} ranges from \{1, 3, 5, 10, 20, 50\}.

\noindent$\bullet$~\textbf{Result:} As shown in Table$_{\!}$~\ref{table:retro}, the newly proposed Nesy based solution (\ie, Graph2Edits$_{\!}$~\cite{zhong2023retrosynthesis}) has a clear advantage over conventional neural models such as Dual-TF \cite{sun2021towards} and GraphRetro$_{\!}$~\cite{somnath2021learning}, yielding improvements of \textbf{1.8}\% and \textbf{1.4}\% on Top-1 Exact match accuracy. This confirms the efficacy of data-and knowledge-driven methods in organic chemistry.

\begin{table*}[t]
   \centering
   \caption{\textbf{Quantitative visual semantic parsing results} (\S\ref{sec5.53}) on {PASCAL-Person-Part}$_{\!}$~\cite{xia2017joint} \texttt{val}, in terms of mIoU. For fairness, all the compared models use Swin-S as the backbone. }
       \vspace{-5pt}
   \small
   \resizebox{0.99\textwidth}{!}{
      \setlength\tabcolsep{5pt}{}
      \renewcommand\arraystretch{1.05}
      \begin{tabular}{z{75}y{20}|c||cccccccccc|ccc}
         \hline\thickhline
         \rowcolor{mygray}
         \multicolumn{2}{c|}{Method} &NeSy & Head &  Torso & U-Arm & L-Arm & U-Leg & L-Leg & U-Body & L-Body & F-Body & B.G. & $\text{mIoU}^3$$\uparrow$& $\text{mIoU}^2$$\uparrow$& $\text{mIoU}^1$$\uparrow$ \\ \hline\hline
         DeepLabV3+~\cite{chen2018encoder}\!\!\!&\!\!\!\pub{ECCV18}\!& & 87.02 & 72.02 & 60.37 & 57.36 & 53.54 & 48.52 & 90.07 & 65.88 & 93.02 & 96.07 & 94.55 & 84.01 &  67.84 \\
         PCNet~\cite{zhang2020part}\!\!\!&\!\!\!\pub{CVPR20}\!&& 90.04 & 76.89 & 69.11 & 68.40 & 60.78 & 60.14 & 94.02 & 68.71 & 96.50& 96.78 & 96.64 & 86.47 & 74.59 \\
         CrossSeg~\cite{wang2021exploring}\!\!\!&\!\!\!\pub{ICCV21}\!&& 90.05 & 75.62 & 68.58 & 66.25 & 58.21 & 57.94 & 92.67 & 67.54 & 96.11& 96.17 & 96.14 & 85.46 & 73.26 \\
         ProtoSeg~\cite{zhou2022rethinking}\!\!\!&\!\!\!\pub{CVPR22}\!&& 90.09 & 77.20 & 69.16 & 68.44 & 60.89 & 60.13 & 93.64 & 68.71 & 96.44& 96.82 & 96.63 & 86.39 & 74.68 \\
         Mask2Former~\cite{cheng2022masked}\!\!\!&\!\!\!\pub{CVPR22}\!&&  \textcolor{blue}{\textbf{90.21}} & 78.26 & 70.14 & 69.51 & 60.73 & 60.31 & 94.52 & 69.23 & 96.80& 97.04 & 96.92 & 86.93 & 75.17 \\
GMMSeg~\cite{lianggmmseg}\!\!\!&\!\!\!\pub{NeurIPS22}\!&& 90.15 & \textcolor{blue}{\textbf{79.10}} & \textcolor{green}{\textbf{70.99}} & 69.43 & \textcolor{green}{\textbf{61.32}} & 60.42 &93.37 & 70.52 & 97.10& 97.02 & 97.06 & 87.17 & 75.49 \\
{ClustSeg}~\cite{liang2023clustseg}\!\!\!&\!\!\!\pub{ICML23}&& \textcolor{green}{\textbf{90.20}} & \textcolor{green}{\textbf{78.94}} & \textcolor{blue}{\textbf{71.23}} & \textcolor{green}{\textbf{69.69}} & \textcolor{blue}{\textbf{61.87}} & \textcolor{blue}{\textbf{61.39}} & \textcolor{green}{\textbf{94.52}} & \textcolor{green}{\textbf{70.81}} & \textcolor{green}{\textbf{97.20}}& \textcolor{blue}{\textbf{97.23}} & \textcolor{green}{\textbf{97.26}} & \textcolor{green}{\textbf{87.52}} & \textcolor{blue}{\textbf{75.79}} \\
            CNIF~\cite{wang2019learning}\!\!\!&\!\!\!\pub{ICCV19}\!&$\checkmark$& 88.02 & 72.91 & 64.31 & 63.52 & 55.61 & 54.96 & 91.82 & 66.56 & 94.33 & 96.02 & 95.18 & 84.80 & 70.76\\
         HHP~\cite{wang2020hierarchical}\!\!\!&\!\!\!\pub{CVPR20}\!&$\checkmark$ & 89.73 & 75.22 & 66.87 & 66.21 & 58.69 & 58.17 & 93.44 & 68.02 & 96.77 & 96.94 & 96.86 & 86.13 & 73.12 \\
            {HSSN}~\cite{li2022deep}\!\!\!&\!\!\!\pub{CVPR22}\!&$\checkmark$& 90.19  &  78.72 & 70.67& \textcolor{blue}{\textbf{69.71}}& 61.15& \textcolor{green}{\textbf{60.44}}& \textcolor{blue}{\textbf{95.86}} & \textcolor{blue}{\textbf{71.56}} & \textcolor{blue}{\textbf{98.20}} & \textcolor{green}{\textbf{97.18}} & \textcolor{blue}{\textbf{97.69}} & \textcolor{blue}{\textbf{88.20}} & \textcolor{green}{\textbf{75.44}} \\
         LogicSeg~\cite{li2023logicseg}\!\!\!&\!\!\!\pub{ICCV23}\!&$\checkmark$& \textcolor{red}{\textbf{90.23}} & \textcolor{red}{\textbf{79.56}} & \textcolor{red}{\textbf{71.44}} & \textcolor{red}{\textbf{70.52}} & \textcolor{red}{\textbf{62.26}} & \textcolor{red}{\textbf{61.46}} & \textcolor{red}{\textbf{95.97}} & \textcolor{red}{\textbf{72.51}} & \textcolor{red}{\textbf{98.43}}& \textcolor{red}{\textbf{97.35}} & \textcolor{red}{\textbf{97.89}} & \textcolor{red}{\textbf{88.61}} & \textcolor{red}{\textbf{76.12}} \\
         \hline
      \end{tabular}
   }
   \label{table:ppp}
\vspace{-1.em}
\end{table*}

    \vspace{-5pt}
\subsection{$_{\!\!\!}$Performance$_{\!}$ Benchmarking:$_{\!}$ Visual$_{\!}$ Semantic$_{\!}$ Parsing$_{\!\!\!\!\!\!}$}\label{sec5.53}
Visual semantic parsing, \ie, interpreting high-level semantic concepts of visual stimuli
at pixel level, is a fundamental and challenging task in the field of computer vision.

\noindent$\bullet$~\textbf{Dataset:$_{\!}$} We$_{\!}$ select$_{\!}$ {PASCAL-Person-Part}$_{\!}$~\cite{xia2017joint},$_{\!}$ a$_{\!}$ widely-used dataset for visual semantic parsing, to evaluate the performance. PASCAL-Person-Part consists of $1,716$/$1,817$ images for \texttt{train}/\texttt{test}. It provides dense annotations for $20$ fine-grained human parts (\eg, \texttt{head}, \texttt{left-arm}) from which a three-layer label hierarchy can be derived: the fine-grained parts belong to two superclasses, \texttt{upper-body} and \texttt{lower-body}, which are further merged into \texttt{full-body}.

\noindent$\bullet$~\textbf{Benchmarking Algorithms:} For performance comparison, we involve four NeSy based structured visual parsers~(\ie, CNIF$_{\!}$~\cite{wang2019learning},$_{\!}$ HHP$_{\!}$~\cite{wang2020hierarchical},$_{\!}$ {HSSN}$_{\!}$~\cite{li2022deep},$_{\!}$ LogicSeg$_{\!}$~\cite{li2023logicseg})$_{\!}$ which$_{\!}$ ex-
ploit the three-level human semantic hierarchy. For a comprehensive evaluation, we also include a group of hierarchy-agnostic segmentation algorithms$_{\!}$ (\ie,$_{\!}$ DeepLabV3+$_{\!}$~\cite{chen2018encoder},
PCNet$_{\!}$~\cite{zhang2020part}, CrossSeg$_{\!}$~\cite{wang2021exploring}, ProtoSeg$_{\!}$~\cite{zhou2022rethinking}, Mask2Former \cite{cheng2022masked}, GMMSeg$_{\!}$~\cite{lianggmmseg}, {ClustSeg}$_{\!}$~\cite{liang2023clustseg}), whose segmentation results on coarse-grained semantics are simply obtained by merging the predictions of the corresponding subclasses.

\noindent$\bullet$~\textbf{Evaluation Metric:} As customary, we employ the mean
intersection-over-union (mIoU) for evaluation. As in~\cite{li2022deep,li2023logicseg}, we further report the average score for each hierarchy level $l$ (denoted as mIoU$^l$) for detailed analysis.

\noindent$\bullet$~\textbf{Result:} Table~\ref{table:ppp} demonstrates that, the newest NeSy based visual semantic parser, \ie,  LogicSeg$_{\!}$~\cite{li2023logicseg}, achieves superior performance over {ClustSeg}~\cite{liang2023clustseg}, the current top-leading purely neural solution,
by \textbf{0.63}\%/\textbf{1.11}\%/\textbf{0.32}\% over the three semantic levels, in terms of mIoU. This suggests the great potential of integrating symbolic reasoning and sub-symbolic learning in large-scale machine perception.

\begin{table}[t]
	\centering
	\small
	\caption{{\textbf{Quantitative MWP solving results} (\S\ref{sec5.54}) on Math23K~\cite{wang2017deep} in terms of answer accuracy. * denotes 5-fold cross-validation.}}
    \vspace{-5pt}
	\small
	\resizebox{0.99\columnwidth}{!}{
		\setlength\tabcolsep{6pt}
		\renewcommand\arraystretch{1.05}
		\begin{tabular}{z{75}y{22}|c||cc}
			\hline\thickhline
			\rowcolor{mygray}
			\multicolumn{2}{c|}{Method}  &NeSy &Accuracy (\%)$\uparrow$ &Accuracy* (\%)$\uparrow$ \\ \hline\hline
DNS~\cite{wang2017deep}\!\!\!&\!\!\!\pub{EMNLP17}\!  & &-  &58.1\\
Math-EN~\cite{wang2018translating}\!\!\!&\!\!\!\pub{EMNLP18}\!  & & 66.7 &-\\
T-RNN~\cite{wang2019template}\!\!\!&\!\!\!\pub{AAAI19}\!  & & 66.9 &-\\
GROUP-ATT~\cite{li2019modeling}\!\!\!&\!\!\!\pub{ACL19}\!  & & 69.5 &66.9\\
TSD~\cite{liu2019tree}\!\!\!&\!\!\!\pub{EMNLP19}\!  &$\checkmark$ & 69.0 &-\\
GTS~\cite{xie2019goal}\!\!\!&\!\!\!\pub{IJCAI19}\!  &$\checkmark$ & 75.6 &74.3\\
Graph2Tree~\cite{zhang2020graph}\!\!\!&\!\!\!\pub{ACL20}\!  &$\checkmark$ & \textcolor{blue}{\textbf{77.4}} &75.5\\
NSS~\cite{qin2021neural}\!\!\!&\!\!\!\pub{ACL21}\!  &$\checkmark$ &- &75.7\\
HMS~\cite{lin2021hms}\!\!\!&\!\!\!\pub{AAAI21}\!  &$\checkmark$ & \textcolor{green}{\textbf{76.1}} &-\\
BERT-Tree~\cite{li2022seeking}\!\!\!&\!\!\!\pub{ACL22}\!  &$\checkmark$ & \textcolor{red}{\textbf{82.4}} &-\\
			\hline
		\end{tabular}
	}
	\label{table:MWP}
 \vspace{-1.em}
\end{table}

    \vspace{-2pt}
\subsection{Performance Benchmarking: Math Word Problem Solving}\label{sec5.54}
The task of solving math word problems (MWPs) is~to automatically answer a mathematical question that is described in natural language. MWP solving is an important natural language understanding task that requires logical reasoning over the quantities presented in the context to compute the numerical answer.

\noindent$\bullet$~\textbf{Dataset:$_{\!}$} Math23K$_{\!}$~\cite{wang2017deep},$_{\!}$ a$_{\!}$ large-scale$_{\!}$ MWP$_{\!}$ dataset,$_{\!}$ is$_{\!}$~used in$_{\!}$ our$_{\!}$ experiments.$_{\!}$ Math23K$_{\!}$ has$_{\!}$ a$_{\!}$ total$_{\!}$ of$_{\!}$ 23,161$_{\!}$ real math word problems for elementary school students with prob- lem$_{\!}$ descriptions,$_{\!}$ structured$_{\!}$ equations$_{\!}$ and$_{\!}$ answers.$_{\!}$ The$_{\!}$ pro- blems are crawled
from multiple online education websites and solved by one-unknown-variable linear expressions.

\noindent$\bullet$~\textbf{Benchmarking Algorithms:} We compare the performance

\noindent of$_{\!}$ ten$_{\!}$ famous$_{\!}$ MWP$_{\!}$ solvers;$_{\!}$ four$_{\!}$ of them are purely based on neural networks, namely DNS$_{\!}$~\cite{wang2017deep}, Math-EN$_{\!}$~\cite{wang2018translating}, T-RNN$_{\!}$~\cite{wang2019template}, GROUP-ATT$_{\!}$~\cite{li2019modeling}. The remaining six solvers, namely TSD~\cite{liu2019tree}, GTS~\cite{xie2019goal}, Graph2Tree~\cite{zhang2020graph}, NSS~\cite{qin2021neural}, HMS~\cite{lin2021hms}, BERT-Tree~\cite{li2022seeking}, are NeSy based.

\noindent$\bullet$~\textbf{Evaluation Metric:} Here the standard evaluation metric in MWP solving, namely answer accuracy, is adopted.

\noindent$\bullet$~\textbf{Result:} Table~\ref{table:MWP} shows that NeSy based solvers generally outperform the four neural competitors. This proves the efficacy of NeSy in dealing with reasoning-heavy symbolic problems. Moreover, with the aid of pre-trained BERT, BERT-Tree~\cite{li2022seeking} provides impressive performance, suggesting the power of combining NeSy with LLMs.

    \vspace{-5pt}
\section{Open Challenges}\label{sec6}
While recent years have witnessed remarkable progress in NeSy, there still exist several open challenges to overcome.

\noindent$\bullet$~\textbf{Scalability:} Current NeSy systems are still struggling~with large-scale$_{\!}$ symbolic/logic$_{\!}$ reasoning.$_{\!}$ First,$_{\!}$ the$_{\!}$ increasing$_{\!}$~ex- pressivity of symbolic/logic rules~\cite{bianchi2019capabilities,bianchi2019complementing,eberhart2020completion}, such as the~inclusion of universal quantification over variables~\cite{bianchi2019complementing}, and the complex syntax in higher-order logic, typically comes with growing computational complexity. Second, the frequent use of symbolic knowledge also impedes the applicability of NeSy systems in large-scale applications in the wild~\cite{mao2018neuro,chaturvedi2019fuzzy}, since grounding massive symbolic knowledge on real-world examples is time-consuming.  Third, collecting large-scale symbolic knowledge, especially in specific domains, is often difficult and expensive. Finally, while recent NeSy systems are relatively easy to make a full use of rich data with the aid of modern connectionist tools, it is less clear whether they can indeed exhibit the desirable features, such as sound reasoning, out-of-distribution generalization, data-efficient learning, transparency, and transferability to new domains, at a large scale. \redminor{These features are promised by NeSy's symbolic aspect, but their realization in the light of real-world complexity requires improved scalability, which has attracted recent researches~\cite{van2023nesi,maene2024soft,li2024neuro} and warrants further investigations.}

\noindent$\bullet$~\textbf{Compositional$_{\!}$ Generalization:$_{\!}$} As$_{\!}$ discussed$_{\!}$ in$_{\!}$~Sec.$_{\!}$~\ref{subsec31}, compositionality, a central aspect of human intelligence,~is among the most desirable characterizations that NeSy sys- tems$_{\!}$ are$_{\!}$ expected$_{\!}$ to$_{\!}$ offer.$_{\!}$ It$_{\!}$ requires$_{\!}$ systematical$_{\!}$ decom- position$_{\!}$ and$_{\!}$ recombination$_{\!}$ of$_{\!}$ the$_{\!}$ learned$_{\!}$ knowledge,$_{\!}$ so~as to$_{\!}$ generalize$_{\!}$ to$_{\!}$ novel$_{\!}$ reasoning$_{\!}$ problems.$_{\!}$ Whilst$_{\!}$ a$_{\!}$ few  attempts~\cite{chen2020compositional,kim2021improving} have been made for algorithmic implementation of compositional generalization within the NeSy framework, they are primarily specialized for toy language games. \redminor{As a result, despite early studies exploring the principle of compositionality~\cite{garcez2006connectionist,garcez2007connectionist,garcez2014neural} and commonsense reasoning~\cite{proietti2023roadmap}, achieving human-like compositionality in NeSy systems, such as the comprehensive use of various forms of logic (\eg, modal, temporal, commonsense, epistemic) and different types of knowledge (\eg, declarative, procedural, causal, and relational) for generalization and reasoning, as well as applying these compositional skills to solve real-world problems, still remains a challenge.}

\noindent$\bullet$~$_{\!}$\textbf{Automated$_{\!}$ Knowledge$_{\!}$ Acquisition:$_{\!}$} Symbolic knowledge$_{\!}$ serves$_{\!}$ as the foundation for developing a NeSy system; it influences the quality and scope of the system's reasoning capability.$_{\!}$ Nevertheless,$_{\!}$ most$_{\!}$ modern$_{\!}$ NeSy$_{\!}$ systems$_{\!}$ simply take$_{\!}$ the$_{\!}$ knowledge$_{\!}$ for$_{\!}$ granted,$_{\!}$ ignoring$_{\!}$ two$_{\!}$ crucial$_{\!}$ issues:  i)$_{\!}$~how$_{\!}$ to$_{\!}$ acquire$_{\!}$ domain-specific$_{\!}$ knowledge$_{\!}$ that$_{\!}$ is$_{\!}$ required for the system; and ii) how to ensure the completeness and adequacy$_{\!}$ of$_{\!}$ the$_{\!}$ knowledge$_{\!}$ for$_{\!}$ supporting$_{\!}$ the$_{\!}$ system's$_{\!}$ fun- ctionality.$_{\!}$~Given$_{\!}$ the$_{\!}$ aforementioned$_{\!}$ challenge$_{\!}$ of$_{\!}$ \textit{scalability},$_{\!}$ knowledge acquisition seems a bottleneck in the process~of developing$_{\!}$ NeSy$_{\!}$ systems$_{\!}$ in$_{\!}$ large-scale$_{\!}$ and$_{\!}$ real-world$_{\!}$ application scenarios. This calls for the automatic acquisition of knowledge (preferably, from different data sources). This is also closely relevant to the concept of \textit{learning to reason}~\cite{khardon1997learning}, which studies the entire process of learning a knowledge base representation from examples, and then reasoning with that knowledge base by querying with similar examples. In fact, automated knowledge acquisition, which is essentially a problem in modeling a human expert's introspective capabilities, has experienced a long-period history of development in the field of AI~\cite{felgenbaum1977art}. Despite recent efforts in this direction are mainly restricted to the construction of concep- tual knowledge graphs~\cite{ji2021survey}, we believe the integration~of
automated knowledge acquisition, symbolic reasoning, and sub-symbolic$_{\!}$ learning$_{\!}$ will$_{\!}$ lead$_{\!}$ to$_{\!}$ more$_{\!}$ complete$_{\!}$ and$_{\!}$ powerful NeSy$_{\!}$ systems.$_{\!}$ Such$_{\!}$ an$_{\!}$ integration touches many fundamental challenges and aspects of NeSy and even AI, such as: i) which kind of knowledge representation is more favored for building NeSy systems; ii) how to abstract knowledge from data; and  iii) how to build a close and mutual-feedback loop between deductive reasoning and inductive learning, \ie, grounding knowledge onto data to guide the practice, and updating knowledge according to the practical results.

\noindent$\bullet$~\textbf{Recursive Neuro[Symbolic] Engine:} Another invaluable research direction is the construction of a Neuro[Symbolic] engine (\ie, the Type 6 NeSy system elaborated in Sec.~\ref{subsec42}) that can deeply embed a symbolic reasoning engine inside a neural sub-symbolic engine. Unlike existing Type 1-5 NeSy systems and modern connectionist machines, such a Neuro[Symbolic] engine explores the mechanism of human intelligence more deeply: how neural activations, which are sub-symbolic and widely distributed in the human brain, give rise to complex behaviors that are symbolic, such as language and logical reasoning. In such a Neuro[Symbolic] engine, the neural part shall be trained with the guidance of the symbolic component's reasoning results, which are derived from the symbolic knowledge, and recursively, the symbolic component shall be evolved by updating its knowledge according to the neural component's feedback, which are induced from data. This loop is closely related to the aforementioned challenge of automated knowledge acquisition. In addition, the Neuro[Symbolic] engine provides a computational realization of Kahneman's System 1 and System 2 theory 1 of cognition, which distinguishes between fast, intuitive, and unconscious System 1 thinking and slow, deliberate, and conscious System 2 thinking. Hence it can achieve both types of thinking and leverage their strengths.

\noindent$\bullet$~\textbf{Testbed for Metacognitive Skills of NeSy:} From a practical perspective, though NeSy systems is widely regarded as one of the most  promising avenues towards human-like AI~\cite{garcez2022neural,booch2021thinking}, the main strands of NeSy's applications are still limited to a handful of tasks (\cf, Sec.~\ref{sec5}). Many of the application tasks are placed in simulators, designed around limited proof-of-concept settings, or with small examples, in contrast to the large vision we hold onto the metacognitive capabilities of human beings, such as productivity, systematicity, compositionality and inferential
coherence of mental thought~\cite{fodor1988connectionism}, causal and counterfactual thinking~\cite{roese1997counterfactual}, deductive reasoning~\cite{liang2022visual}, interpretability~\cite{pagin2010compositionality,smolensky2022neurocompositionala}, \etc, as well as their extensive, daily use.
To advance NeSy towards this vision, we need more challenging and appropriate playgrounds that seek fundamental progress of NeSy in mastering human metacognitive skills. Some promising domains for such benchmarks include social robotics, health informatics, hardware/software specification, and scientific problems in genomics, chemistry, and astronomy, where both large amounts of data and knowledge are available and the discovery of scientific hypotheses is needed.

\noindent$\bullet$~\textbf{NeSy$_{\!}$ in$_{\!}$ the$_{\!}$ Big$_{\!}$ Model$_{\!}$ Era:$_{\!}$} The$_{\!}$ community$_{\!}$ has$_{\!}$ recently witnessed remarkable
progress fueled by large AI models. Large AI models exhibit emergent abilities (\eg, in-context learning, chain-of-thought reasoning), and can accomplish diverse tasks in a zero-shot fashion or with the aid of a few
 examples, akin to human beings. Albeit these astonishing abilities, it is becoming increasingly clear that large
AI models still suffer several deficiencies, such as their pronounced opacity, insatiable demand for data and computational resources, and tendency to generate nonsensical or unfaithful content, known as ``hallucination''. These drawbacks reveal their inherent biases, lack of real-world understanding, and weakness in generalizing or reasoning beyond their scope. These are intrinsic limitations of connectionist models and exacerbated by the heightened sophistication and scale of$_{\!}$ large$_{\!}$ models.$_{\!}$ With$_{\!}$ regard$_{\!}$ to$_{\!}$ this,$_{\!}$ it$_{\!}$ is$_{\!}$ appealing$_{\!}$ to$_{\!}$ explore the integration of large AI models and symbolic techniques. Such integration can address the limitations of big neural models and empower the symbolic part with the massive implicit knowledge encoded by the big models, hence stepping closer towards artificial general intelligence. The recent emerge of LLM based AI agents\!~\cite{yang2024doraemongpt} that can automatically compose external tools for real-world task solving supports this view, although they only achieve loose neural-symbolic integration (see Sec.~\ref{subsec42} Type~2). In short, developing NeSy with big AI models is a promising and challenging direction that requires dense collaboration across
different AI fields.

    \vspace{-5pt}
\section{Conclusions}\label{sec7}
{Though having a long history, NeSy remained a rather niche topic until recently when landmark advances in machine learning -- pushed by the wave of deep learning -- caused increasing interest in forming the bridge between neural and symbolic methods.  In this work, we conducted a large-scale and up-to-date survey of the rapidly growing area, from five perspectives: \textbf{i)} A historical point of view -- we provide a brief review of early research results of NeSy; \textbf{ii)} A motivation point of view -- we clarify two major driving forces behind the field as well as the recent AI debate which promotes the research activity in NeSy; \textbf{iii)} A methodological point of view -- we classify and analyze the contemporary NeSy systems from four dimensions: neural-symbolic integration, knowledge representation, knowledge embedding, and functionality; \textbf{iv)} An application point of view -- we outline several key application areas including scientific discovery, programming systems, question-answering, vision-language analysis and reasoning, robotics and control, visual scene understanding, and mathematical reasoning; and \textbf{v)} An experimental point of view -- we providing a
performance benchmarking of several NeSy methods on three representative application tasks including retrosynthesis prediction, visual semantic parsing, and math word problem
solving.  In the end, we discuss outstanding challenges and areas for future research. Although a strong NeSy system is still far from achieved, given the significant progress in AI over the past decade, we remain optimistic about the future and believe NeSy is a
promising direction for the development of the next generation of AI.



%
%
%
%
%

\ifCLASSOPTIONcaptionsoff
  \newpage
\fi



%
	\vspace{-4pt}
{\small
\bibliographystyle{IEEEtran}

\bibliography{egbib}

\begin{thebibliography}{100}
\providecommand{\url}[1]{#1}
\csname url@samestyle\endcsname
\providecommand{\newblock}{\relax}
\providecommand{\bibinfo}[2]{#2}
\providecommand{\BIBentrySTDinterwordspacing}{\spaceskip=0pt\relax}
\providecommand{\BIBentryALTinterwordstretchfactor}{4}
\providecommand{\BIBentryALTinterwordspacing}{\spaceskip=\fontdimen2\font plus
\BIBentryALTinterwordstretchfactor\fontdimen3\font minus
  \fontdimen4\font\relax}
\providecommand{\BIBforeignlanguage}[2]{{%
\expandafter\ifx\csname l@#1\endcsname\relax
\typeout{** WARNING: IEEEtran.bst: No hyphenation pattern has been}%
\typeout{** loaded for the language `#1'. Using the pattern for}%
\typeout{** the default language instead.}%
\else
\language=\csname l@#1\endcsname
\fi
#2}}
\providecommand{\BIBdecl}{\relax}
\BIBdecl

\bibitem{yang2021multiple}
Y.~Yang, Y.~Zhuang, and Y.~Pan, ``Multiple knowledge representation for big
  data artificial intelligence: framework, applications, and case studies,''
  \emph{Frontiers of Information Technology \& Electronic Engineering},
  vol.~22, no.~12, pp. 1551--1558, 2021.

\bibitem{haugeland1989artificial}
J.~Haugeland, \emph{Artificial Intelligence: The very idea}.\hskip 1em plus
  0.5em minus 0.4em\relax MIT press, 1989.

\bibitem{zhang2021neural}
J.~Zhang, B.~Chen, L.~Zhang, X.~Ke, and H.~Ding, ``Neural, symbolic and
  neural-symbolic reasoning on knowledge graphs,'' \emph{AI Open}, vol.~2, pp.
  14--35, 2021.

\bibitem{lecun2015deep}
Y.~LeCun, Y.~Bengio, and G.~Hinton, ``Deep learning,'' \emph{Nature}, vol. 521,
  no. 7553, pp. 436--444, 2015.

\bibitem{rumelhart1985learning}
D.~E. Rumelhart, G.~E. Hinton, and R.~J. Williams, ``Learning internal
  representations by error propagation,'' California Univ San Diego La Jolla
  Inst for Cognitive Science, Tech. Rep., 1985.

\bibitem{smolensky2022neurocompositionala}
P.~Smolensky, R.~T. McCoy, R.~Fernandez, M.~Goldrick, and J.~Gao,
  \emph{Neurocompositional computing in human and machine intelligence: A
  tutorial}.\hskip 1em plus 0.5em minus 0.4em\relax Microsoft Technical Report
  MSR-TR-2022, 2022.

\bibitem{smolensky2022neurocompositionalb}
P.~Smolensky, R.~McCoy, R.~Fernandez, M.~Goldrick, and J.~Gao,
  ``Neurocompositional computing: From the central paradox of cognition to a
  new generation of ai systems,'' \emph{AI Magazine}, vol.~43, no.~3, pp.
  308--322, 2022.

\bibitem{fodor1988connectionism}
J.~A. Fodor and Z.~W. Pylyshyn, ``Connectionism and cognitive architecture: A
  critical analysis,'' \emph{Cognition}, vol.~28, no. 1-2, pp. 3--71, 1988.

\bibitem{mcculloch1943logical}
W.~S. McCulloch and W.~Pitts, ``A logical calculus of the ideas immanent in
  nervous activity,'' \emph{The Bulletin of Mathematical Biophysics}, vol.~5,
  no.~4, pp. 115--133, 1943.

\bibitem{lake2017building}
B.~M. Lake, T.~D. Ullman, J.~B. Tenenbaum, and S.~J. Gershman, ``Building
  machines that learn and think like people,'' \emph{Behavioral and brain
  sciences}, vol.~40, 2017.

\bibitem{marcus2018deep}
G.~Marcus, ``Deep learning: A critical appraisal,'' \emph{arXiv preprint
  arXiv:1801.00631}, 2018.

\bibitem{valiant2003three}
L.~G. Valiant, ``Three problems in computer science,'' \emph{Journal of the
  ACM}, vol.~50, no.~1, pp. 96--99, 2003.

\bibitem{garcez2019neural}
A.~Garcez, M.~Gori, L.~Lamb, L.~Serafini, M.~Spranger, and S.~Tran,
  ``Neural-symbolic computing: An effective methodology for principled
  integration of machine learning and reasoning,'' \emph{Journal of Applied
  Logics}, vol.~6, no.~4, pp. 611--632, 2019.

\bibitem{garcez2022neural}
T.~R. Besold, A.~d’Avila Garcez, S.~Bader, H.~Bowman, P.~Domingos,
  P.~Hitzler, K.-U. K{\"u}hnberger, L.~C. Lamb, P.~M.~V. Lima, L.~de~Penning
  \emph{et~al.}, ``Neural-symbolic learning and reasoning: A survey and
  interpretation,'' \emph{Neuro-Symbolic Artificial Intelligence: The State of
  the Art}, vol. 342, p.~1, 2022.

\bibitem{donadello2017logic}
I.~Donadello, L.~Serafini, and A.~d'Avila Garcez, ``Logic tensor networks for
  semantic image interpretation,'' in \emph{International Joint Conferences on
  Artificial Intelligence}, 2017, pp. 1596--1602.

\bibitem{zhou2021cascaded}
T.~Zhou, S.~Qi, W.~Wang, J.~Shen, and S.-C. Zhu, ``Cascaded parsing of
  human-object interaction recognition,'' \emph{IEEE Trans. Pattern Anal. Mach.
  Intell.}, vol.~44, no.~6, pp. 2827--2840, 2021.

\bibitem{andreas2016neural}
J.~Andreas, M.~Rohrbach, T.~Darrell, and D.~Klein, ``Neural module networks,''
  in \emph{Proc. IEEE Conf. Comput. Vis. Pattern Recognit.}, 2016, pp. 39--48.

\bibitem{mao2019neuro}
J.~Mao, C.~Gan, P.~Kohli, J.~B. Tenenbaum, and J.~Wu, ``The neuro-symbolic
  concept learner: Interpreting scenes, words, and sentences from natural
  supervision,'' in \emph{Proc. Int. Conf. Learn. Representations}, 2019.

\bibitem{amizadeh2020neuro}
S.~Amizadeh, H.~Palangi, A.~Polozov, Y.~Huang, and K.~Koishida,
  ``Neuro-symbolic visual reasoning: Disentangling ``visual'' from
  ``reasoning'','' in \emph{Proc. ACM Int. Conf. Mach. Learn.}, 2020, pp.
  279--290.

\bibitem{wang2021hierarchical}
W.~Wang, T.~Zhou, S.~Qi, J.~Shen, and S.-C. Zhu, ``Hierarchical human semantic
  parsing with comprehensive part-relation modeling,'' \emph{IEEE Trans.
  Pattern Anal. Mach. Intell.}, vol.~44, no.~07, pp. 3508--3522, 2021.

\bibitem{li2022deep}
L.~Li, T.~Zhou, W.~Wang, J.~Li, and Y.~Yang, ``Deep hierarchical semantic
  segmentation,'' in \emph{Proc. IEEE Conf. Comput. Vis. Pattern Recognit.},
  2022, pp. 1246--1257.

\bibitem{arabshahi2021conversational}
F.~Arabshahi, J.~Lee, M.~Gawarecki, K.~Mazaitis, A.~Azaria, and T.~Mitchell,
  ``Conversational neuro-symbolic commonsense reasoning,'' in \emph{AAAI
  Conference on Artificial Intelligence}, 2021, pp. 4902--4911.

\bibitem{belle2020symbolic}
V.~Belle, ``Symbolic logic meets machine learning: A brief survey in infinite
  domains,'' in \emph{International Conference on Scalable Uncertainty
  Management}, 2020, pp. 3--16.

\bibitem{lamb2021graph}
L.~C. Lamb, A.~d'Avila Garcez, M.~Gori, M.~O. Prates, P.~H. Avelar, and M.~Y.
  Vardi, ``Graph neural networks meet neural-symbolic computing: A survey and
  perspective,'' in \emph{International Joint Conferences on Artificial
  Intelligence}, 2021, pp. 4877--4884.

\bibitem{yu2021survey}
D.~Yu, B.~Yang, D.~Liu, and H.~Wang, ``A survey on neural-symbolic systems,''
  \emph{arXiv preprint arXiv:2111.08164}, 2021.

\bibitem{ijcai2022p767}
E.~Giunchiglia, M.~C. Stoian, and T.~Lukasiewicz, ``Deep learning with logical
  constraints,'' in \emph{International Joint Conferences on Artificial
  Intelligence}, 2022, pp. 5478--5485.

\bibitem{giunchiglia2022deep}
------, ``Deep learning with logical constraints,'' in \emph{International
  Joint Conferences on Artificial Intelligence}, 2022.

\bibitem{bader2005dimensions}
S.~Bader and P.~Hitzler, ``Dimensions of neural-symbolic integration -- a
  structured survey,'' \emph{arXiv preprint cs/0511042}, 2005.

\bibitem{garcez2008neural}
A.~S. Garcez, L.~C. Lamb, and D.~M. Gabbay, \emph{Neural-symbolic cognitive
  reasoning}.\hskip 1em plus 0.5em minus 0.4em\relax Springer Berlin
  Heidelberg, 2009.

\bibitem{shi2020neural}
S.~Shi, H.~Chen, W.~Ma, J.~Mao, M.~Zhang, and Y.~Zhang, ``Neural logic
  reasoning,'' in \emph{International Conference on Information \& Knowledge
  Management}, 2020, pp. 1365--1374.

\bibitem{towell1990refinement}
G.~G. Towell, J.~W. Shavlik, M.~O. Noordewier \emph{et~al.}, ``Refinement of
  approximate domain theories by knowledge-based neural networks,'' in
  \emph{Proceedings of the National Conference on Artificial Intelligence},
  1990, pp. 861--866.

\bibitem{pollack1990recursive}
J.~B. Pollack, ``Recursive distributed representations,'' \emph{Artificial
  Intelligence}, vol.~46, no. 1-2, pp. 77--105, 1990.

\bibitem{shastri1993simple}
L.~Shastri and V.~Ajjanagadde, ``From simple associations to systematic
  reasoning: A connectionist representation of rules, variables and dynamic
  bindings using temporal synchrony,'' \emph{Behavioral and Brain Sciences},
  vol.~16, no.~3, pp. 417--451, 1993.

\bibitem{holldobler1991towards}
S.~H{\"o}lldobler, Y.~Kalinke, F.~W. Ki \emph{et~al.}, ``Towards a new
  massively parallel computational model for logic programming,'' in
  \emph{ECAI'94 workshop on Combining Symbolic and Connectioninst Processing},
  1991.

\bibitem{garcez1999connectionist}
A.~Garcez and G.~Zaverucha, ``The connectionist inductive learning and logic
  programming system,'' \emph{Applied Intelligence Journal}, vol.~11, no.~1,
  pp. 59--77, 1999.

\bibitem{towell1994knowledge}
G.~G. Towell and J.~W. Shavlik, ``Knowledge-based artificial neural networks,''
  \emph{Artificial intelligence}, vol.~70, no. 1-2, pp. 119--165, 1994.

\bibitem{plate1995holographic}
T.~A. Plate, ``Holographic reduced representations,'' \emph{IEEE Transactions
  on Neural Networks}, vol.~6, no.~3, pp. 623--641, 1995.

\bibitem{cloete2000knowledge}
I.~Cloete and J.~M. Zurada, \emph{Knowledge-based neurocomputing}.\hskip 1em
  plus 0.5em minus 0.4em\relax MIT press, 2000.

\bibitem{browne2001connectionist}
A.~Browne and R.~Sun, ``Connectionist inference models,'' \emph{Neural
  Networks}, vol.~14, no.~10, pp. 1331--1355, 2001.

\bibitem{garcez2002neural}
A.~S.~d. Garcez, K.~Broda, D.~M. Gabbay \emph{et~al.}, \emph{Neural-symbolic
  learning systems: foundations and applications}.\hskip 1em plus 0.5em minus
  0.4em\relax Springer Science \& Business Media, 2002.

\bibitem{deacon1997co}
T.~W. Deacon, ``The co-evolution of language and the brain,'' \emph{W.W.
  Norton, New York}, vol.~2, 1997.

\bibitem{russell2016artificial}
S.~J. Russell and P.~Norvig, \emph{Artificial Intelligence: A modern approach},
  3rd~ed.\hskip 1em plus 0.5em minus 0.4em\relax Pearson, 2009.

\bibitem{horst2003computational}
S.~Horst, ``The computational theory of mind,'' 2003.

\bibitem{hitzler2022neuro}
P.~Hitzler, A.~Eberhart, M.~Ebrahimi, M.~K. Sarker, and L.~Zhou,
  ``Neuro-symbolic approaches in artificial intelligence,'' \emph{National
  Science Review}, vol.~9, no.~6, 2022.

\bibitem{marcus2020next}
G.~Marcus, ``The next decade in {AI}: Four steps towards robust artificial
  intelligence,'' \emph{arXiv preprint arXiv:2002.06177}, 2020.

\bibitem{maruyama2020symbolic}
Y.~Maruyama, ``Symbolic and statistical theories of cognition: towards
  integrated artificial intelligence,'' in \emph{International Conference on
  Software Engineering and Formal Methods}, 2020, pp. 129--146.

\bibitem{dantsin2001complexity}
E.~Dantsin, T.~Eiter, G.~Gottlob, and A.~Voronkov, ``Complexity and expressive
  power of logic programming,'' \emph{ACM Computing Surveys}, vol.~33, no.~3,
  pp. 374--425, 2001.

\bibitem{mitchell1997machine}
T.~M. Mitchell, \emph{Machine learning}.\hskip 1em plus 0.5em minus 0.4em\relax
  New York: McGraw-hill, 1997, vol.~1, no.~9.

\bibitem{wang2024visual}
W.~Wang, Y.~Yang, and Y.~Pan, ``Visual knowledge in the big model era:
  Retrospect and prospect,'' \emph{Frontiers of Information Technology \&
  Electronic Engineering}, 2024.

\bibitem{churchland1994computational}
P.~S. Churchland and T.~J. Sejnowski, \emph{The computational brain}.\hskip 1em
  plus 0.5em minus 0.4em\relax MIT press, 1994.

\bibitem{kiparsky1969syntactic}
P.~Kiparsky and J.~F. Staal, ``Syntactic and semantic relations in panini,''
  \emph{Foundations of Language}, pp. 83--117, 1969.

\bibitem{janssen2012compositionality}
T.~M. Janssen, ``Compositionality: Its historic context,'' \emph{The Oxford
  Handbook of Compositionality}, pp. 19--46, 2012.

\bibitem{szabo2012case}
Z.~Szab{\'o}, ``The case for compositionality,'' \emph{The Oxford Handbook of
  Compositionality}, vol.~64, p.~80, 2012.

\bibitem{pagin2010compositionality}
P.~Pagin and D.~Westerst{\aa}hl, ``Compositionality i: Definitions and
  variants,'' \emph{Philosophy Compass}, vol.~5, no.~3, pp. 250--264, 2010.

\bibitem{smolensky1988proper}
P.~Smolensky, ``On the proper treatment of connectionism,'' \emph{Behavioral
  and Brain Sciences}, vol.~11, no.~1, pp. 1--23, 1988.

\bibitem{booch2021thinking}
G.~Booch, F.~Fabiano, L.~Horesh, K.~Kate, J.~Lenchner, N.~Linck, A.~Loreggia,
  K.~Murgesan, N.~Mattei, F.~Rossi \emph{et~al.}, ``Thinking fast and slow in
  {AI},'' in \emph{AAAI Conference on Artificial Intelligence}, 2021, pp.
  15\,042--15\,046.

\bibitem{kahneman2011thinking}
D.~Kahneman, \emph{Thinking, fast and slow}.\hskip 1em plus 0.5em minus
  0.4em\relax Farrar, Strauss and Giroux, 2011.

\bibitem{kautz2022third}
H.~Kautz, ``The third {AI} summer: {AAAI} {R}obert s. {E}ngelmore memorial
  lecture,'' \emph{AI Magazine}, vol.~43, no.~1, pp. 93--104, 2022.

\bibitem{mikolov2013efficient}
T.~Mikolov, K.~Chen, G.~Corrado, and J.~Dean, ``Efficient estimation of word
  representations in vector space,'' \emph{arXiv preprint arXiv:1301.3781},
  2013.

\bibitem{pennington2014glove}
J.~Pennington, R.~Socher, and C.~D. Manning, ``Glove: Global vectors for word
  representation,'' in \emph{Proceedings of the Conference on Empirical Methods
  in Natural Language Processing}, 2014, pp. 1532--1543.

\bibitem{brown2020language}
T.~Brown, B.~Mann, N.~Ryder, M.~Subbiah, J.~D. Kaplan, P.~Dhariwal,
  A.~Neelakantan, P.~Shyam, G.~Sastry, A.~Askell \emph{et~al.}, ``Language
  models are few-shot learners,'' in \emph{Proc. Advances Neural Inf. Process.
  Syst}, 2020, pp. 1877--1901.

\bibitem{silver2016mastering}
D.~Silver, A.~Huang, C.~J. Maddison, A.~Guez, L.~Sifre, G.~Van Den~Driessche,
  J.~Schrittwieser, I.~Antonoglou, V.~Panneershelvam, M.~Lanctot \emph{et~al.},
  ``Mastering the game of go with deep neural networks and tree search,''
  \emph{Nature}, vol. 529, no. 7587, pp. 484--489, 2016.

\bibitem{chen2020compositional}
X.~Chen, C.~Liang, A.~W. Yu, D.~Song, and D.~Zhou, ``Compositional
  generalization via neural-symbolic stack machines,'' in \emph{Proc. Advances
  Neural Inf. Process. Syst}, 2020, pp. 1690--1701.

\bibitem{dang2020plans}
R.~Dang-Nhu, ``Plans: Neuro-symbolic program learning from videos,'' in
  \emph{Proc. Advances Neural Inf. Process. Syst}, 2020, pp. 22\,445--22\,455.

\bibitem{gupta2023visual}
T.~Gupta and A.~Kembhavi, ``Visual programming: Compositional visual reasoning
  without training,'' in \emph{Proc. IEEE Conf. Comput. Vis. Pattern
  Recognit.}, 2023, pp. 14\,953--14\,962.

\bibitem{shen2023hugginggpt}
Y.~Shen, K.~Song, X.~Tan, D.~Li, W.~Lu, and Y.~Zhuang, ``Hugginggpt: Solving ai
  tasks with chatgpt and its friends in huggingface,'' in \emph{Proc. Advances
  Neural Inf. Process. Syst}, 2023.

\bibitem{surismenon2023vipergpt}
D.~Sur\'is, S.~Menon, and C.~Vondrick, ``Vipergpt: Visual inference via python
  execution for reasoning,'' in \emph{Proc. IEEE Int. Conf. Comput. Vis.},
  2023.

\bibitem{thompson2020computational}
N.~C. Thompson, K.~Greenewald, K.~Lee, and G.~F. Manso, ``The computational
  limits of deep learning,'' \emph{arXiv preprint arXiv:2007.05558}, 2020.

\bibitem{yi2018neural}
K.~Yi, J.~Wu, C.~Gan, A.~Torralba, P.~Kohli, and J.~B. Tenenbaum,
  ``Neural-symbolic {VQA}: Disentangling reasoning from vision and language
  understanding,'' in \emph{Proc. Advances Neural Inf. Process. Syst}, 2018,
  pp. 1039--1050.

\bibitem{parisotto2017neuro}
E.~Parisotto, A.-r. Mohamed, R.~Singh, L.~Li, D.~Zhou, and P.~Kohli,
  ``Neuro-symbolic program synthesis,'' in \emph{Proc. Int. Conf. Learn.
  Representations}, 2017.

\bibitem{chen2019neural}
X.~Chen, C.~Liang, A.~W. Yu, D.~Zhou, D.~Song, and Q.~V. Le, ``Neural symbolic
  reader: Scalable integration of distributed and symbolic representations for
  reading comprehension,'' in \emph{Proc. Int. Conf. Learn. Representations},
  2019.

\bibitem{nye2020learning}
M.~Nye, A.~Solar-Lezama, J.~Tenenbaum, and B.~M. Lake, ``Learning compositional
  rules via neural program synthesis,'' in \emph{Proc. Advances Neural Inf.
  Process. Syst}, 2020, pp. 10\,832--10\,842.

\bibitem{liang2017neural}
C.~Liang, J.~Berant, Q.~Le, K.~Forbus, and N.~Lao, ``Neural symbolic machines:
  Learning semantic parsers on freebase with weak supervision,'' in
  \emph{Proceedings of the Annual Meeting of the Association for Computational
  Linguistics}, 2017.

\bibitem{young2019learning}
H.~Young, O.~Bastani, and M.~Naik, ``Learning neurosymbolic generative models
  via program synthesis,'' in \emph{Proc. ACM Int. Conf. Mach. Learn.}, 2019,
  pp. 7144--7153.

\bibitem{garnelo2016towards}
M.~Garnelo, K.~Arulkumaran, and M.~Shanahan, ``Towards deep symbolic
  reinforcement learning,'' in \emph{arXiv preprint arXiv:1609.05518}, 2016.

\bibitem{mou2017coupling}
L.~Mou, Z.~Lu, H.~Li, and Z.~Jin, ``Coupling distributed and symbolic execution
  for natural language queries,'' in \emph{Proc. ACM Int. Conf. Mach. Learn.},
  2017, pp. 2518--2526.

\bibitem{de2011neural}
H.~L.~H. de~Penning, A.~S.~d. Garcez, L.~C. Lamb, and J.-J.~C. Meyer, ``A
  neural-symbolic cognitive agent for online learning and reasoning,'' in
  \emph{International Joint Conferences on Artificial Intelligence}, 2011, pp.
  1653--1658.

\bibitem{valkov2018houdini}
L.~Valkov, D.~Chaudhari, A.~Srivastava, C.~Sutton, and S.~Chaudhuri, ``Houdini:
  Lifelong learning as program synthesis,'' in \emph{Proc. Advances Neural Inf.
  Process. Syst}, 2018, pp. 8701--8712.

\bibitem{yang2018peorl}
F.~Yang, D.~Lyu, B.~Liu, and S.~Gustafson, ``{PEORL}: Integrating symbolic
  planning and hierarchical reinforcement learning for robust
  decision-making,'' in \emph{International Joint Conferences on Artificial
  Intelligence}, 2018, pp. 4860--4866.

\bibitem{lyu2019sdrl}
D.~Lyu, F.~Yang, B.~Liu, and S.~Gustafson, ``Sdrl: interpretable and
  data-efficient deep reinforcement learning leveraging symbolic planning,'' in
  \emph{AAAI Conference on Artificial Intelligence}, 2019, pp. 2970--2977.

\bibitem{jin2022creativity}
M.~Jin, Z.~Ma, K.~Jin, H.~H. Zhuo, C.~Chen, and C.~Yu, ``Creativity of ai:
  Automatic symbolic option discovery for facilitating deep reinforcement
  learning,'' in \emph{AAAI Conference on Artificial Intelligence}, 2022, pp.
  7042--7050.

\bibitem{hohenecker2020ontology}
P.~Hohenecker and T.~Lukasiewicz, ``Ontology reasoning with deep neural
  networks,'' \emph{Journal of Artificial Intelligence Research}, vol.~68, pp.
  503--540, 2020.

\bibitem{jiang2019neural}
Z.~Jiang and S.~Luo, ``Neural logic reinforcement learning,'' in \emph{Proc.
  ACM Int. Conf. Mach. Learn.}, 2019, pp. 3110--3119.

\bibitem{dai2019bridging}
W.-Z. Dai, Q.~Xu, Y.~Yu, and Z.-H. Zhou, ``Bridging machine learning and
  logical reasoning by abductive learning,'' in \emph{Proc. Advances Neural
  Inf. Process. Syst}, 2019, pp. 2815--2826.

\bibitem{yang2017differentiable}
F.~Yang, Z.~Yang, and W.~W. Cohen, ``Differentiable learning of logical rules
  for knowledge base reasoning,'' in \emph{Proc. Advances Neural Inf. Process.
  Syst}, 2017, pp. 2316--2325.

\bibitem{rocktaschel2017end}
T.~Rockt{\"a}schel and S.~Riedel, ``End-to-end differentiable proving,'' in
  \emph{Proc. Advances Neural Inf. Process. Syst}, 2017, pp. 3791--3803.

\bibitem{minervini2020learning}
P.~Minervini, S.~Riedel, P.~Stenetorp, E.~Grefenstette, and T.~Rockt{\"a}schel,
  ``Learning reasoning strategies in end-to-end differentiable proving,'' in
  \emph{Proc. ACM Int. Conf. Mach. Learn.}, 2020, pp. 6938--6949.

\bibitem{weber2019nlprolog}
L.~Weber, P.~Minervini, J.~M{\"u}nchmeyer, U.~Leser, and T.~Rockt{\"a}schl,
  ``Nlprolog: Reasoning with weak unification for question answering in natural
  language,'' in \emph{Annual Meeting of the Association for Computational
  Linguistics}, 2019, pp. 6151--6161.

\bibitem{manhaeve2018deepproblog}
R.~Manhaeve, S.~Dumancic, A.~Kimmig, T.~Demeester, and L.~D. Raedt,
  ``Deepproblog: neural probabilistic logic programming,'' in \emph{Proc.
  Advances Neural Inf. Process. Syst}, 2018, pp. 3753--3763.

\bibitem{tsamoura2021neural}
E.~Tsamoura, T.~Hospedales, and L.~Michael, ``Neural-symbolic integration: A
  compositional perspective,'' in \emph{AAAI Conference on Artificial
  Intelligence}, 2021, pp. 5051--5060.

\bibitem{si2019synthesizing}
X.~Si, M.~Raghothaman, K.~Heo, and M.~Naik, ``Synthesizing datalog programs
  using numerical relaxation,'' in \emph{International Joint Conferences on
  Artificial Intelligence}, 2019, pp. 6117--6124.

\bibitem{cohen2016tensorlog}
W.~W. Cohen, ``Tensorlog: A differentiable deductive database,'' \emph{arXiv
  preprint arXiv:1605.06523}, 2016.

\bibitem{garcez2020neurosymbolic}
A.~d. Garcez and L.~C. Lamb, ``Neurosymbolic {AI}: the 3rd wave,'' \emph{arXiv
  preprint arXiv:2012.05876}, 2020.

\bibitem{dai2018syntax}
H.~Dai, Y.~Tian, B.~Dai, S.~Skiena, and L.~Song, ``Syntax-directed variational
  autoencoder for structured data,'' in \emph{Proc. Int. Conf. Learn.
  Representations}, 2018.

\bibitem{chen2018tree}
X.~Chen, C.~Liu, and D.~Song, ``Tree-to-tree neural networks for program
  translation,'' in \emph{Proc. Advances Neural Inf. Process. Syst}, 2018, pp.
  2552--2562.

\bibitem{jin2018junction}
W.~Jin, R.~Barzilay, and T.~Jaakkola, ``Junction tree variational autoencoder
  for molecular graph generation,'' in \emph{Proc. ACM Int. Conf. Mach.
  Learn.}, 2018, pp. 2323--2332.

\bibitem{allamanis2017learning}
M.~Allamanis, P.~Chanthirasegaran, P.~Kohli, and C.~Sutton, ``Learning
  continuous semantic representations of symbolic expressions,'' in \emph{Proc.
  ACM Int. Conf. Mach. Learn.}, 2017, pp. 80--88.

\bibitem{lample2019deep}
G.~Lample and F.~Charton, ``Deep learning for symbolic mathematics,'' in
  \emph{Proc. Int. Conf. Learn. Representations}, 2019.

\bibitem{li2020closed}
Q.~Li, S.~Huang, Y.~Hong, Y.~Chen, Y.~N. Wu, and S.-C. Zhu, ``Closed loop
  neural-symbolic learning via integrating neural perception, grammar parsing,
  and symbolic reasoning,'' in \emph{Proc. ACM Int. Conf. Mach. Learn.}, 2020,
  pp. 5884--5894.

\bibitem{arabshahi2018combining}
F.~Arabshahi, S.~Singh, and A.~Anandkumar, ``Combining symbolic expressions and
  black-box function evaluations in neural programs,'' in \emph{Proc. Int.
  Conf. Learn. Representations}, 2018.

\bibitem{johnson2017inferring}
J.~Johnson, B.~Hariharan, L.~Van Der~Maaten, J.~Hoffman, L.~Fei-Fei,
  C.~Lawrence~Zitnick, and R.~Girshick, ``Inferring and executing programs for
  visual reasoning,'' in \emph{Proc. IEEE Int. Conf. Comput. Vis.}, 2017, pp.
  2989--2998.

\bibitem{dhingra2019differentiable}
B.~Dhingra, M.~Zaheer, V.~Balachandran, G.~Neubig, R.~Salakhutdinov, and W.~W.
  Cohen, ``Differentiable reasoning over a virtual knowledge base,'' in
  \emph{Proc. Int. Conf. Learn. Representations}, 2019.

\bibitem{dumancic2019learning}
S.~Dumancic, T.~Guns, W.~Meert, and H.~Blockeel, ``Learning relational
  representations with auto-encoding logic programs,'' in \emph{International
  Joint Conferences on Artificial Intelligence}, 2019, pp. 6081--6087.

\bibitem{evans2018learning}
R.~Evans and E.~Grefenstette, ``Learning explanatory rules from noisy data,''
  \emph{Journal of Artificial Intelligence Research}, vol.~61, pp. 1--64, 2018.

\bibitem{zhang2018neural}
L.~Zhang, G.~Rosenblatt, E.~Fetaya, R.~Liao, W.~Byrd, M.~Might, R.~Urtasun, and
  R.~Zemel, ``Neural guided constraint logic programming for program
  synthesis,'' in \emph{Proc. Advances Neural Inf. Process. Syst}, 2018, pp.
  1744--1753.

\bibitem{marra2021neural}
G.~Marra and O.~Ku{\v{z}}elka, ``Neural markov logic networks,''
  \emph{Uncertainty in Artificial Intelligence}, pp. 908--917, 2021.

\bibitem{yang2019learn}
Y.~Yang and L.~Song, ``Learn to explain efficiently via neural logic inductive
  learning,'' in \emph{Proc. Int. Conf. Learn. Representations}, 2019.

\bibitem{hudson2019learning}
D.~Hudson and C.~D. Manning, ``Learning by abstraction: The neural state
  machine,'' in \emph{Proc. Advances Neural Inf. Process. Syst}, 2019, pp.
  5903--5916.

\bibitem{shi2019explainable}
J.~Shi, H.~Zhang, and J.~Li, ``Explainable and explicit visual reasoning over
  scene graphs,'' in \emph{Proc. IEEE Conf. Comput. Vis. Pattern Recognit.},
  2019, pp. 8376--8384.

\bibitem{vedantam2019probabilistic}
R.~Vedantam, K.~Desai, S.~Lee, M.~Rohrbach, D.~Batra, and D.~Parikh,
  ``Probabilistic neural symbolic models for interpretable visual question
  answering,'' in \emph{Proc. ACM Int. Conf. Mach. Learn.}, 2019, pp.
  6428--6437.

\bibitem{xie2019embedding}
Y.~Xie, Z.~Xu, M.~S. Kankanhalli, K.~S. Meel, and H.~Soh, ``Embedding symbolic
  knowledge into deep networks,'' in \emph{Proc. Advances Neural Inf. Process.
  Syst}, 2019, pp. 4233--4243.

\bibitem{wang2019learning}
W.~Wang, Z.~Zhang, S.~Qi, J.~Shen, Y.~Pang, and L.~Shao, ``Learning
  compositional neural information fusion for human parsing,'' in \emph{Proc.
  IEEE Int. Conf. Comput. Vis.}, 2019, pp. 5703--5713.

\bibitem{wang2020hierarchical}
W.~Wang, H.~Zhu, J.~Dai, Y.~Pang, J.~Shen, and L.~Shao, ``Hierarchical human
  parsing with typed part-relation reasoning,'' in \emph{Proc. IEEE Conf.
  Comput. Vis. Pattern Recognit.}, 2020, pp. 8929--8939.

\bibitem{lin2019kagnet}
B.~Y. Lin, X.~Chen, J.~Chen, and X.~Ren, ``Kagnet: Knowledge-aware graph
  networks for commonsense reasoning,'' in \emph{Conference on Empirical
  Methods in Natural Language Processing}, 2019, pp. 2829--2839.

\bibitem{lv2020graph}
S.~Lv, D.~Guo, J.~Xu, D.~Tang, N.~Duan, M.~Gong, L.~Shou, D.~Jiang, G.~Cao, and
  S.~Hu, ``Graph-based reasoning over heterogeneous external knowledge for
  commonsense question answering,'' in \emph{AAAI Conference on Artificial
  Intelligence}, 2020, pp. 8449--8456.

\bibitem{teru2020inductive}
K.~Teru, E.~Denis, and W.~Hamilton, ``Inductive relation prediction by subgraph
  reasoning,'' in \emph{Proc. ACM Int. Conf. Mach. Learn.}, 2020, pp.
  9448--9457.

\bibitem{gu2019scene}
J.~Gu, H.~Zhao, Z.~Lin, S.~Li, J.~Cai, and M.~Ling, ``Scene graph generation
  with external knowledge and image reconstruction,'' in \emph{Proc. IEEE Conf.
  Comput. Vis. Pattern Recognit.}, 2019, pp. 1969--1978.

\bibitem{marino2021krisp}
K.~Marino, X.~Chen, D.~Parikh, A.~Gupta, and M.~Rohrbach, ``Krisp: Integrating
  implicit and symbolic knowledge for open-domain knowledge-based vqa,'' in
  \emph{Proc. IEEE Conf. Comput. Vis. Pattern Recognit.}, 2021, pp.
  14\,111--14\,121.

\bibitem{cranmer2020discovering}
M.~Cranmer, A.~Sanchez~Gonzalez, P.~Battaglia, R.~Xu, K.~Cranmer, D.~Spergel,
  and S.~Ho, ``Discovering symbolic models from deep learning with inductive
  biases,'' in \emph{Proc. Advances Neural Inf. Process. Syst}, 2020, pp.
  17\,429--17\,442.

\bibitem{liang2018symbolic}
X.~Liang, Z.~Hu, H.~Zhang, L.~Lin, and E.~P. Xing, ``Symbolic graph reasoning
  meets convolutions,'' in \emph{Proc. Advances Neural Inf. Process. Syst},
  2018, pp. 1858--1868.

\bibitem{garg2020symbolic}
S.~Garg, A.~Bajpai \emph{et~al.}, ``Symbolic network: generalized neural
  policies for relational mdps,'' in \emph{Proc. ACM Int. Conf. Mach. Learn.},
  2020, pp. 3397--3407.

\bibitem{demeter2020just}
D.~Demeter and D.~Downey, ``Just add functions: A neural-symbolic language
  model,'' in \emph{AAAI Conference on Artificial Intelligence}, 2020, pp.
  7634--7642.

\bibitem{hoernle2022multiplexnet}
N.~Hoernle, R.~M. Karampatsis, V.~Belle, and K.~Gal, ``Multiplexnet: Towards
  fully satisfied logical constraints in neural networks,'' in \emph{AAAI
  Conference on Artificial Intelligence}, 2022, pp. 5700--5709.

\bibitem{giunchiglia2021multi}
E.~Giunchiglia and T.~Lukasiewicz, ``Multi-label classification neural networks
  with hard logical constraints,'' \emph{Journal of Artificial Intelligence
  Research}, vol.~72, pp. 759--818, 2021.

\bibitem{giunchiglia2020coherent}
------, ``Coherent hierarchical multi-label classification networks,'' in
  \emph{Proc. Advances Neural Inf. Process. Syst}, 2020, pp. 9662--9673.

\bibitem{ahmed2022semantic}
K.~Ahmed, S.~Teso, K.-W. Chang, G.~Van~den Broeck, and A.~Vergari, ``Semantic
  probabilistic layers for neuro-symbolic learning,'' in \emph{Proc. Advances
  Neural Inf. Process. Syst}, 2022, pp. 29\,944--29\,959.

\bibitem{giunchiglia2024ccn+}
E.~Giunchiglia, A.~Tatomir, M.~C. Stoian, and T.~Lukasiewicz, ``{CCN+}: A
  neuro-symbolic framework for deep learning with requirements,''
  \emph{International Journal of Approximate Reasoning}, p. 109124, 2024.

\bibitem{zhang2023tractable}
H.~Zhang, M.~Dang, N.~Peng, and G.~Van~den Broeck, ``Tractable control for
  autoregressive language generation,'' in \emph{International Conference on
  Machine Learning}, 2023, pp. 40\,932--40\,945.

\bibitem{de2007problog}
L.~De~Raedt, A.~Kimmig, and H.~Toivonen, ``Problog: A probabilistic prolog and
  its application in link discovery.'' in \emph{International Joint Conferences
  on Artificial Intelligence}, 2007, pp. 2462--2467.

\bibitem{radford2019language}
A.~Radford, J.~Wu, R.~Child, D.~Luan, D.~Amodei, I.~Sutskever \emph{et~al.},
  ``Language models are unsupervised multitask learners,'' \emph{OpenAI blog},
  vol.~1, no.~8, p.~9, 2019.

\bibitem{serafini2016logic}
L.~Serafini and A.~d. Garcez, ``Logic tensor networks: Deep learning and
  logical reasoning from data and knowledge,'' in \emph{Proceedings of the 11th
  International Workshop on Neural-Symbolic Learning and Reasoning}, 2016.

\bibitem{ijcai2017p221}
I.~Donadello, L.~Serafini, and A.~d'Avila Garcez, ``Logic tensor networks for
  semantic image interpretation,'' in \emph{International Joint Conferences on
  Artificial Intelligence}, 2017, pp. 1596--1602.

\bibitem{badreddine2022logic}
S.~Badreddine, A.~d. Garcez, L.~Serafini, and M.~Spranger, ``Logic tensor
  networks,'' \emph{Artificial Intelligence}, vol. 303, p. 103649, 2022.

\bibitem{liang2019learning}
Y.~Liang and G.~Van~den Broeck, ``Learning logistic circuits,'' in \emph{AAAI
  Conference on Artificial Intelligence}, 2019, pp. 4277--4286.

\bibitem{demeester2016lifted}
T.~Demeester, T.~Rockt{\"a}schel, and S.~Riedel, ``Lifted rule injection for
  relation embeddings,'' in \emph{Conference on Empirical Methods in Natural
  Language Processing}, 2016, pp. 1389--1399.

\bibitem{hu2016harnessing}
Z.~Hu, X.~Ma, Z.~Liu, E.~Hovy, and E.~Xing, ``Harnessing deep neural networks
  with logic rules,'' in \emph{Proceedings of the Annual Meeting of the
  Association for Computational Linguistics}, 2016, pp. 2410--2420.

\bibitem{stewart2017label}
R.~Stewart and S.~Ermon, ``Label-free supervision of neural networks with
  physics and domain knowledge,'' in \emph{AAAI Conference on Artificial
  Intelligence}, 2017, pp. 2576--2582.

\bibitem{xu2018semantic}
J.~Xu, Z.~Zhang, T.~Friedman, Y.~Liang, and G.~Broeck, ``A semantic loss
  function for deep learning with symbolic knowledge,'' in \emph{Proc. ACM Int.
  Conf. Mach. Learn.}, 2018, pp. 5502--5511.

\bibitem{ahmed2024pseudo}
K.~Ahmed, K.-W. Chang, and G.~Van~den Broeck, ``A pseudo-semantic loss for
  autoregressive models with logical constraints,'' in \emph{Proc. Advances
  Neural Inf. Process. Syst}, 2023, pp. 18\,325--18\,340.

\bibitem{muralidhar2018incorporating}
N.~Muralidhar, M.~R. Islam, M.~Marwah, A.~Karpatne, and N.~Ramakrishnan,
  ``Incorporating prior domain knowledge into deep neural networks,'' in
  \emph{IEEE International Conference on Big Data}, 2018, pp. 36--45.

\bibitem{van2022analyzing}
E.~van Krieken, E.~Acar, and F.~van Harmelen, ``Analyzing differentiable fuzzy
  logic operators,'' \emph{Artificial Intelligence}, vol. 302, p. 103602, 2022.

\bibitem{wang2020integrating}
W.~Wang and S.~J. Pan, ``Integrating deep learning with logic fusion for
  information extraction,'' in \emph{AAAI Conference on Artificial
  Intelligence}, 2020, pp. 9225--9232.

\bibitem{li2019augmenting}
T.~Li and V.~Srikumar, ``Augmenting neural networks with first-order logic,''
  in \emph{Proceedings of the Annual Meeting of the Association for
  Computational Linguistics}, 2019.

\bibitem{diligenti2017semantic}
M.~Diligenti, M.~Gori, and C.~Sacca, ``Semantic-based regularization for
  learning and inference,'' \emph{Artificial Intelligence}, vol. 244, pp.
  143--165, 2017.

\bibitem{marra2019lyrics}
G.~Marra, F.~Giannini, M.~Diligenti, and M.~Gori, ``Lyrics: A general interface
  layer to integrate logic inference and deep learning,'' in \emph{Joint
  European Conference on Machine Learning and Knowledge Discovery in
  Databases}, 2019, pp. 283--298.

\bibitem{fischer2019dl2}
M.~Fischer, M.~Balunovic, D.~Drachsler-Cohen, T.~Gehr, C.~Zhang, and M.~Vechev,
  ``Dl2: Training and querying neural networks with logic,'' in \emph{Proc. ACM
  Int. Conf. Mach. Learn.}, 2019, pp. 1931--1941.

\bibitem{wehrmann2018hierarchical}
J.~Wehrmann, R.~Cerri, and R.~Barros, ``Hierarchical multi-label classification
  networks,'' in \emph{Proc. ACM Int. Conf. Mach. Learn.}, 2018, pp.
  5075--5084.

\bibitem{bertinetto2020making}
L.~Bertinetto, R.~Mueller, K.~Tertikas, S.~Samangooei, and N.~A. Lord, ``Making
  better mistakes: Leveraging class hierarchies with deep networks,'' in
  \emph{Proc. IEEE Conf. Comput. Vis. Pattern Recognit.}, 2020, pp.
  12\,506--12\,515.

\bibitem{dong2018neural}
H.~Dong, J.~Mao, T.~Lin, C.~Wang, L.~Li, and D.~Zhou, ``Neural logic
  machines,'' in \emph{Proc. Int. Conf. Learn. Representations}, 2018.

\bibitem{wang2019satnet}
P.-W. Wang, P.~Donti, B.~Wilder, and Z.~Kolter, ``Satnet: Bridging deep
  learning and logical reasoning using a differentiable satisfiability
  solver,'' in \emph{Proc. ACM Int. Conf. Mach. Learn.}, 2019, pp. 6545--6554.

\bibitem{cai2017making}
J.~Cai, R.~Shin, and D.~Song, ``Making neural programming architectures
  generalize via recursion,'' in \emph{Proc. Int. Conf. Learn.
  Representations}, 2017.

\bibitem{li2023logicseg}
L.~Li, W.~Wang, and Y.~Yang, ``Logicseg: Parsing visual semantics with neural
  logic learning and reasoning,'' in \emph{ICCV}, 2023.

\bibitem{novak2012mathematical}
V.~Nov{\'a}k, I.~Perfilieva, and J.~Mockor, \emph{Mathematical principles of
  fuzzy logic}.\hskip 1em plus 0.5em minus 0.4em\relax Springer Science \&
  Business Media, 2012, vol. 517.

\bibitem{cohen2020scalable}
W.~W. Cohen, H.~Sun, R.~A. Hofer, and M.~Siegler, ``Scalable neural methods for
  reasoning with a symbolic knowledge base,'' in \emph{Proc. Int. Conf. Learn.
  Representations}, 2020.

\bibitem{deng2009imagenet}
J.~Deng, W.~Dong, R.~Socher, L.-J. Li, K.~Li, and L.~Fei-Fei, ``Imagenet: A
  large-scale hierarchical image database,'' in \emph{Proc. IEEE Conf. Comput.
  Vis. Pattern Recognit.}, 2009, pp. 248--255.

\bibitem{cordts2016cityscapes}
M.~Cordts, M.~Omran, S.~Ramos, T.~Rehfeld, M.~Enzweiler, R.~Benenson,
  U.~Franke, S.~Roth, and B.~Schiele, ``The cityscapes dataset for semantic
  urban scene understanding,'' in \emph{Proc. IEEE Conf. Comput. Vis. Pattern
  Recognit.}, 2016.

\bibitem{miller1995wordnet}
G.~A. Miller, ``Wordnet: a lexical database for english,'' \emph{Communications
  of the ACM}, vol.~38, no.~11, pp. 39--41, 1995.

\bibitem{giunchiglia2022jml}
E.~Giunchiglia, M.~C. Stoian, S.~Khan, F.~Cuzzolin, and T.~Lukasiewicz,
  ``Road-{R}: The autonomous driving dataset with logical requirements,''
  \emph{Machine Learning}, 2023.

\bibitem{qu2019probabilistic}
M.~Qu and J.~Tang, ``Probabilistic logic neural networks for reasoning,'' in
  \emph{Proc. Advances Neural Inf. Process. Syst}, 2019, pp. 7712--7722.

\bibitem{zhang2019efficient}
Y.~Zhang, X.~Chen, Y.~Yang, A.~Ramamurthy, B.~Li, Y.~Qi, and L.~Song,
  ``Efficient probabilistic logic reasoning with graph neural networks,'' in
  \emph{Proc. Int. Conf. Learn. Representations}, 2019.

\bibitem{marra2020relational}
G.~Marra, M.~Diligenti, F.~Giannini, M.~Gori, and M.~Maggini, ``Relational
  neural machines,'' in \emph{Proceedings of the European Conference on
  Artificial Intelligence}, 2020.

\bibitem{richardson2006markov}
M.~Richardson and P.~Domingos, ``Markov logic networks,'' \emph{Machine
  Learning}, vol.~62, no.~1, pp. 107--136, 2006.

\bibitem{clocksin2003programming}
W.~F. Clocksin and C.~S. Mellish, \emph{Programming in PROLOG}.\hskip 1em plus
  0.5em minus 0.4em\relax Springer Science \& Business Media, 2003.

\bibitem{lee2013action}
J.~Lee, V.~Lifschitz, and F.~Yang, ``Action language bc: Preliminary report.''
  in \emph{International Joint Conferences on Artificial Intelligence}, 2013,
  pp. 983--989.

\bibitem{asai2021learning}
M.~Asai and C.~Muise, ``Learning neural-symbolic descriptive planning models
  via cube-space priors: the voyage home (to strips),'' in \emph{International
  Joint Conferences on Artificial Intelligence}, 2021, pp. 2676--2682.

\bibitem{luetal2021neurologic}
X.~Lu, P.~West, R.~Zellers, R.~Le~Bras, C.~Bhagavatula, and Y.~Choi,
  ``{N}euro{L}ogic decoding: (un)supervised neural text generation with
  predicate logic constraints,'' in \emph{Proceedings of NAACL-HLT}, 2021, pp.
  4288--4299.

\bibitem{luetal2022neurologic}
X.~Lu, S.~Welleck, P.~West, L.~Jiang, J.~Kasai, D.~Khashabi, R.~Le~Bras,
  L.~Qin, Y.~Yu, R.~Zellers, N.~A. Smith, and Y.~Choi, ``{N}euro{L}ogic
  {A}*esque decoding: Constrained text generation with lookahead heuristics,''
  in \emph{Proceedings of NAACL-HLT}, 2022, pp. 780--799.

\bibitem{segler2017neural}
M.~H. Segler and M.~P. Waller, ``Neural-symbolic machine learning for
  retrosynthesis and reaction prediction,'' \emph{Chemistry--A European
  Journal}, vol.~23, no.~25, pp. 5966--5971, 2017.

\bibitem{udrescu2020ai}
S.-M. Udrescu and M.~Tegmark, ``{AI} {F}eynman: A physics-inspired method for
  symbolic regression,'' \emph{Science Advances}, vol.~6, no.~16, p. 2631,
  2020.

\bibitem{shah2020learning}
A.~Shah, E.~Zhan, J.~Sun, A.~Verma, Y.~Yue, and S.~Chaudhuri, ``Learning
  differentiable programs with admissible neural heuristics,'' in \emph{Proc.
  Advances Neural Inf. Process. Syst}, 2020, pp. 4940--4952.

\bibitem{jumper2021highly}
J.~Jumper, R.~Evans, A.~Pritzel, T.~Green, M.~Figurnov, O.~Ronneberger,
  K.~Tunyasuvunakool, R.~Bates, A.~{\v{Z}}{\'\i}dek, A.~Potapenko
  \emph{et~al.}, ``Highly accurate protein structure prediction with
  alphafold,'' \emph{Nature}, vol. 596, no. 7873, pp. 583--589, 2021.

\bibitem{sacha2021molecule}
M.~Sacha, M.~B{\l}az, P.~Byrski, P.~Dabrowski-Tumanski, M.~Chrominski,
  R.~Loska, P.~W{\l}odarczyk-Pruszynski, and S.~Jastrzebski, ``Molecule edit
  graph attention network: modeling chemical reactions as sequences of graph
  edits,'' \emph{Journal of Chemical Information and Modeling}, vol.~61, no.~7,
  pp. 3273--3284, 2021.

\bibitem{zhong2023retrosynthesis}
W.~Zhong, Z.~Yang, and C.~Y.-C. Chen, ``Retrosynthesis prediction using an
  end-to-end graph generative architecture for molecular graph editing,''
  \emph{Nature Communications}, vol.~14, no.~1, p. 3009, 2023.

\bibitem{tseng2022automatic}
A.~Tseng, J.~J. Sun, and Y.~Yue, ``Automatic synthesis of diverse weak
  supervision sources for behavior analysis,'' in \emph{Proc. IEEE Conf.
  Comput. Vis. Pattern Recognit.}, 2022, pp. 2211--2220.

\bibitem{segler2018planning}
M.~H. Segler, M.~Preuss, and M.~P. Waller, ``Planning chemical syntheses with
  deep neural networks and symbolic ai,'' \emph{Nature}, vol. 555, no. 7698,
  pp. 604--610, 2018.

\bibitem{dai2019retrosynthesis}
H.~Dai, C.~Li, C.~Coley, B.~Dai, and L.~Song, ``Retrosynthesis prediction with
  conditional graph logic network,'' in \emph{Proc. Advances Neural Inf.
  Process. Syst}, 2019, pp. 8872--8882.

\bibitem{murali2017neural}
V.~Murali, L.~Qi, S.~Chaudhuri, and C.~Jermaine, ``Neural sketch learning for
  conditional program generation,'' in \emph{Proc. Int. Conf. Learn.
  Representations}, 2018.

\bibitem{balog2017deepcoder}
M.~Balog, A.~Gaunt, M.~Brockschmidt, S.~Nowozin, and D.~Tarlow, ``Deepcoder:
  Learning to write programs,'' in \emph{Proc. Int. Conf. Learn.
  Representations}, 2017.

\bibitem{mukherjee2021neural}
R.~Mukherjee, Y.~Wen, D.~Chaudhari, T.~Reps, S.~Chaudhuri, and C.~Jermaine,
  ``Neural program generation modulo static analysis,'' in \emph{Proc. Advances
  Neural Inf. Process. Syst}, 2021, pp. 18\,984--18\,996.

\bibitem{ellis2018learning}
K.~Ellis, D.~Ritchie, A.~Solar-Lezama, and J.~Tenenbaum, ``Learning to infer
  graphics programs from hand-drawn images,'' in \emph{Proc. Advances Neural
  Inf. Process. Syst}, 2018, pp. 6062--6071.

\bibitem{verga2021adaptable}
P.~Verga, H.~Sun, L.~B. Soares, and W.~Cohen, ``Adaptable and interpretable
  neural memory over symbolic knowledge,'' in \emph{North American Chapter of
  the Association for Computational Linguistics}, 2021, pp. 3678--3691.

\bibitem{bosselut2021dynamic}
A.~Bosselut, R.~Le~Bras, and Y.~Choi, ``Dynamic neuro-symbolic knowledge graph
  construction for zero-shot commonsense question answering.'' in \emph{AAAI
  Conference on Artificial Intelligence}, 2021, pp. 4923--4931.

\bibitem{ma2021knowledge}
K.~Ma, F.~Ilievski, J.~Francis, Y.~Bisk, E.~Nyberg, and A.~Oltramari,
  ``Knowledge-driven data construction for zero-shot evaluation in commonsense
  question answering,'' in \emph{AAAI Conference on Artificial Intelligence},
  2021, pp. 13\,507--13\,515.

\bibitem{gupta2019neural}
N.~Gupta, K.~Lin, D.~Roth, S.~Singh, and M.~Gardner, ``Neural module networks
  for reasoning over text,'' in \emph{Proc. Int. Conf. Learn. Representations},
  2020.

\bibitem{ravishankar2022two}
S.~Ravishankar, J.~Thai, I.~Abdelaziz, N.~Mihindukulasooriya, T.~Naseem,
  P.~Kapanipathi, G.~Rossiello, and A.~Fokoue, ``A two-stage approach towards
  generalization in knowledge base question answering,'' in \emph{Conference on
  Empirical Methods in Natural Language Processing}, 2022.

\bibitem{ye2022rng}
X.~Ye, S.~Yavuz, K.~Hashimoto, Y.~Zhou, and C.~Xiong, ``Rng-kbqa: Generation
  augmented iterative ranking for knowledge base question answering,'' in
  \emph{Proceedings of the Annual Meeting of the Association for Computational
  Linguistics}, 2022, pp. 6032--6043.

\bibitem{hudson2018compositional}
D.~A. Hudson and C.~D. Manning, ``Compositional attention networks for machine
  reasoning,'' in \emph{Proc. Int. Conf. Learn. Representations}, 2018.

\bibitem{saqur2020multimodal}
R.~Saqur and K.~Narasimhan, ``Multimodal graph networks for compositional
  generalization in visual question answering,'' in \emph{Proc. Advances Neural
  Inf. Process. Syst}, 2020, pp. 3070--3081.

\bibitem{andreas2016learning}
J.~Andreas, M.~Rohrbach, T.~Darrell, and D.~Klein, ``Learning to compose neural
  networks for question answering,'' in \emph{Proceedings of NAACL-HLT}, 2016,
  pp. 1545--1554.

\bibitem{hu2017learning}
R.~Hu, J.~Andreas, M.~Rohrbach, T.~Darrell, and K.~Saenko, ``Learning to
  reason: End-to-end module networks for visual question answering,'' in
  \emph{Proc. IEEE Int. Conf. Comput. Vis.}, 2017, pp. 804--813.

\bibitem{mao2018neuro}
J.~Mao, C.~Gan, P.~Kohli, J.~B. Tenenbaum, and J.~Wu, ``The neuro-symbolic
  concept learner: Interpreting scenes, words, and sentences from natural
  supervision,'' in \emph{Proc. Int. Conf. Learn. Representations}, 2018.

\bibitem{mascharka2018transparency}
D.~Mascharka, P.~Tran, R.~Soklaski, and A.~Majumdar, ``Transparency by design:
  Closing the gap between performance and interpretability in visual
  reasoning,'' in \emph{Proc. IEEE Conf. Comput. Vis. Pattern Recognit.}, 2018,
  pp. 4942--4950.

\bibitem{sun2021neuro}
J.~Sun, H.~Sun, T.~Han, and B.~Zhou, ``Neuro-symbolic program search for
  autonomous driving decision module design,'' in \emph{Conference on Robot
  Learning}, 2021.

\bibitem{silver2022learning}
T.~Silver, A.~Athalye, J.~B. Tenenbaum, T.~Lozano-Perez, and L.~P. Kaelbling,
  ``Learning neuro-symbolic skills for bilevel planning,'' in \emph{Conference
  on Robot Learning}, 2022.

\bibitem{zhu2021hierarchical}
Y.~Zhu, J.~Tremblay, S.~Birchfield, and Y.~Zhu, ``Hierarchical planning for
  long-horizon manipulation with geometric and symbolic scene graphs,'' in
  \emph{IEEE International Conference on Robotics and Automation}, 2021, pp.
  6541--6548.

\bibitem{xu2018neural}
D.~Xu, S.~Nair, Y.~Zhu, J.~Gao, A.~Garg, L.~Fei-Fei, and S.~Savarese, ``Neural
  task programming: Learning to generalize across hierarchical tasks,'' in
  \emph{IEEE International Conference on Robotics and Automation}, 2018, pp.
  3795--3802.

\bibitem{liu2019tree}
Q.~Liu, W.~Guan, S.~Li, and D.~Kawahara, ``Tree-structured decoding for solving
  math word problems,'' in \emph{Proceedings of the Conference on Empirical
  Methods in Natural Language Processing and the International Joint Conference
  on Natural Language Processing}, 2019, pp. 2370--2379.

\bibitem{xie2019goal}
Z.~Xie and S.~Sun, ``A goal-driven tree-structured neural model for math word
  problems.'' in \emph{International Joint Conferences on Artificial
  Intelligence}, 2019, pp. 5299--5305.

\bibitem{zhang2020graph}
J.~Zhang, L.~Wang, R.~K.-W. Lee, Y.~Bin, Y.~Wang, J.~Shao, and E.-P. Lim,
  ``Graph-to-tree learning for solving math word problems,'' in
  \emph{Proceedings of the Annual Meeting of the Association for Computational
  Linguistics}, 2020, pp. 3928--3937.

\bibitem{paliwal2020graph}
A.~Paliwal, S.~Loos, M.~Rabe, K.~Bansal, and C.~Szegedy, ``Graph
  representations for higher-order logic and theorem proving,'' in \emph{AAAI
  Conference on Artificial Intelligence}, 2020, pp. 2967--2974.

\bibitem{lin2021hms}
X.~Lin, Z.~Huang, H.~Zhao, E.~Chen, Q.~Liu, H.~Wang, and S.~Wang, ``Hms: A
  hierarchical solver with dependency-enhanced understanding for math word
  problem,'' in \emph{AAAI Conference on Artificial Intelligence}, 2021, pp.
  4232--4240.

\bibitem{qin2021neural}
J.~Qin, X.~Liang, Y.~Hong, J.~Tang, and L.~Lin, ``Neural-symbolic solver for
  math word problems with auxiliary tasks,'' in \emph{Proceedings of the Annual
  Meeting of the Association for Computational Linguistics and the
  International Joint Conference on Natural Language Processing (Volume 1: Long
  Papers)}, 2021, pp. 5870--5881.

\bibitem{li2022seeking}
Z.~Li, W.~Zhang, C.~Yan, Q.~Zhou, C.~Li, H.~Liu, and Y.~Cao, ``Seeking
  patterns, not just memorizing procedures: Contrastive learning for solving
  math word problems,'' in \emph{Proceedings of the Annual Meeting of the
  Association for Computational Linguistics}, 2022, pp. 2486--2496.

\bibitem{fawzi2022discovering}
A.~Fawzi, M.~Balog, A.~Huang, T.~Hubert, B.~Romera-Paredes, M.~Barekatain,
  A.~Novikov, F.~J. R~Ruiz, J.~Schrittwieser, G.~Swirszcz \emph{et~al.},
  ``Discovering faster matrix multiplication algorithms with reinforcement
  learning,'' \emph{Nature}, vol. 610, no. 7930, pp. 47--53, 2022.

\bibitem{li2023softened}
Z.~Li, Y.~Yao, T.~Chen, J.~Xu, C.~Cao, X.~Ma, L.~Jian \emph{et~al.}, ``Softened
  symbol grounding for neuro-symbolic systems,'' in \emph{The Eleventh
  International Conference on Learning Representations}, 2023.

\bibitem{atkinson2017towards}
K.~Atkinson, P.~Baroni, M.~Giacomin, A.~Hunter, H.~Prakken, C.~Reed, G.~Simari,
  M.~Thimm, and S.~Villata, ``Towards artificial argumentation,'' \emph{AI
  magazine}, vol.~38, no.~3, pp. 25--36, 2017.

\bibitem{proietti2023roadmap}
M.~Proietti and F.~Toni, ``A roadmap for neuro-argumentative learning,'' in
  \emph{Proceedings of the 17th International Workshop on Neural-Symbolic
  Learning and Reasoning}, 2023, pp. 1--8.

\bibitem{d2005value}
A.~S. D'Avila~Garcez, D.~M. Gabbay, and L.~C. Lamb, ``Value-based argumentation
  frameworks as neural-symbolic learning systems,'' \emph{Journal of Logic and
  Computation}, vol.~15, no.~6, pp. 1041--1058, 2005.

\bibitem{garcez2014neural}
A.~S.~d. Garcez, D.~M. Gabbay, and L.~C. Lamb, ``A neural cognitive model of
  argumentation with application to legal inference and decision making,''
  \emph{Journal of Applied Logic}, vol.~12, no.~2, pp. 109--127, 2014.

\bibitem{cocarascu2019extracting}
O.~Cocarascu, A.~Rago, and F.~Toni, ``Extracting dialogical explanations for
  review aggregations with argumentative dialogical agents.'' in \emph{AAMAS},
  2019, pp. 1261--1269.

\bibitem{graves2012long}
A.~Graves and A.~Graves, ``Long short-term memory,'' \emph{Supervised sequence
  labelling with recurrent neural networks}, pp. 37--45, 2012.

\bibitem{riveret2020neuro}
R.~Riveret, S.~Tran, and A.~d. Garcez, ``Neuro-symbolic probabilistic
  argumentation machines,'' in \emph{Proceedings of the International
  Conference on Principles of Knowledge Representation and Reasoning}, 2020,
  pp. 871--881.

\bibitem{capobianco2011argument}
M.~Capobianco and G.~R. Simari, ``An argument-based multi-agent system for
  information integration,'' in \emph{Argumentation in Multi-Agent Systems: 7th
  International Workshop}, 2011, pp. 171--189.

\bibitem{carrera2015systematic}
{\'A}.~Carrera and C.~A. Iglesias, ``A systematic review of argumentation
  techniques for multi-agent systems research,'' \emph{Artificial Intelligence
  Review}, vol.~44, pp. 509--535, 2015.

\bibitem{gao2016argumentation}
Y.~Gao, F.~Toni, H.~Wang, and F.~Xu, ``Argumentation-based multi-agent decision
  making with privacy preserved,'' in \emph{Proceedings of the 2016
  International Conference on Autonomous Agents \& Multiagent Systems}, 2016,
  pp. 1153--1161.

\bibitem{monte2024argumentation}
H.~Monte-Alto, M.~Morveli-Espinoza, and C.~Tacla, ``Argumentation-based
  multi-agent distributed reasoning in dynamic and open environments,''
  \emph{Knowledge and Information Systems}, pp. 1--36, 2024.

\bibitem{von2021informed}
L.~Von~Rueden, S.~Mayer, K.~Beckh, B.~Georgiev, S.~Giesselbach, R.~Heese,
  B.~Kirsch, J.~Pfrommer, A.~Pick, R.~Ramamurthy \emph{et~al.}, ``Informed
  machine learning--a taxonomy and survey of integrating prior knowledge into
  learning systems,'' \emph{IEEE Transactions on Knowledge and Data
  Engineering}, vol.~35, no.~1, pp. 614--633, 2021.

\bibitem{dash2022review}
T.~Dash, S.~Chitlangia, A.~Ahuja, and A.~Srinivasan, ``A review of some
  techniques for inclusion of domain-knowledge into deep neural networks,''
  \emph{Scientific Reports}, vol.~12, no.~1, p. 1040, 2022.

\bibitem{krenn2022scientific}
M.~Krenn, R.~Pollice, S.~Y. Guo, M.~Aldeghi, A.~Cervera-Lierta, P.~Friederich,
  G.~dos Passos~Gomes, F.~H{\"a}se, A.~Jinich, A.~Nigam \emph{et~al.}, ``On
  scientific understanding with artificial intelligence,'' \emph{Nature Reviews
  Physics}, vol.~4, no.~12, pp. 761--769, 2022.

\bibitem{hsu2023ns3d}
J.~Hsu, J.~Mao, and J.~Wu, ``Ns3d: Neuro-symbolic grounding of 3d objects and
  relations,'' in \emph{Proc. IEEE Conf. Comput. Vis. Pattern Recognit.}, 2023,
  pp. 2614--2623.

\bibitem{ryan2002using}
M.~R. Ryan, ``Using abstract models of behaviours to automatically generate
  reinforcement learning hierarchies,'' in \emph{Proc. ACM Int. Conf. Mach.
  Learn.}, 2002, pp. 522--529.

\bibitem{sutton2018reinforcement}
R.~S. Sutton and A.~G. Barto, \emph{Reinforcement learning: An
  introduction}.\hskip 1em plus 0.5em minus 0.4em\relax MIT press, 2018.

\bibitem{schneider2016s}
N.~Schneider, N.~Stiefl, and G.~A. Landrum, ``What's what: The (nearly)
  definitive guide to reaction role assignment,'' \emph{Journal of Chemical
  Information and Modeling}, vol.~56, no.~12, pp. 2336--2346, 2016.

\bibitem{liu2017retrosynthetic}
B.~Liu, B.~Ramsundar, P.~Kawthekar, J.~Shi, J.~Gomes, Q.~Luu~Nguyen, S.~Ho,
  J.~Sloane, P.~Wender, and V.~Pande, ``Retrosynthetic reaction prediction
  using neural sequence-to-sequence models,'' \emph{ACS central science},
  vol.~3, no.~10, pp. 1103--1113, 2017.

\bibitem{seo2021gta}
S.-W. Seo, Y.~Y. Song, J.~Y. Yang, S.~Bae, H.~Lee, J.~Shin, S.~J. Hwang, and
  E.~Yang, ``Gta: Graph truncated attention for retrosynthesis,'' in
  \emph{Proceedings of the AAAI Conference on Artificial Intelligence}, 2021,
  pp. 531--539.

\bibitem{wang2021retroprime}
X.~Wang, Y.~Li, J.~Qiu, G.~Chen, H.~Liu, B.~Liao, C.-Y. Hsieh, and X.~Yao,
  ``Retroprime: A diverse, plausible and transformer-based method for
  single-step retrosynthesis predictions,'' \emph{Chemical Engineering
  Journal}, vol. 420, p. 129845, 2021.

\bibitem{sun2021towards}
R.~Sun, H.~Dai, L.~Li, S.~Kearnes, and B.~Dai, ``Towards understanding
  retrosynthesis by energy-based models,'' in \emph{Proc. Advances Neural Inf.
  Process. Syst}, 2021, pp. 10\,186--10\,194.

\bibitem{somnath2021learning}
V.~R. Somnath, C.~Bunne, C.~Coley, A.~Krause, and R.~Barzilay, ``Learning graph
  models for retrosynthesis prediction,'' in \emph{Proc. Advances Neural Inf.
  Process. Syst}, 2021, pp. 9405--9415.

\bibitem{faldu2021towards}
K.~Faldu, A.~Sheth, P.~Kikani, M.~Gaur, and A.~Avasthi, ``Towards tractable
  mathematical reasoning: Challenges, strategies, and opportunities for solving
  math word problems,'' \emph{arXiv preprint arXiv:2111.05364}, 2021.

\bibitem{saxton2018analysing}
D.~Saxton, E.~Grefenstette, F.~Hill, and P.~Kohli, ``Analysing mathematical
  reasoning abilities of neural models,'' in \emph{Proc. Int. Conf. Learn.
  Representations}, 2018.

\bibitem{dung1995acceptability}
P.~M. Dung, ``On the acceptability of arguments and its fundamental role in
  nonmonotonic reasoning, logic programming and n-person games,''
  \emph{Artificial intelligence}, vol.~77, no.~2, pp. 321--357, 1995.

\bibitem{bondarenko1997abstract}
A.~Bondarenko, P.~M. Dung, R.~A. Kowalski, and F.~Toni, ``An abstract,
  argumentation-theoretic approach to default reasoning,'' \emph{Artificial
  intelligence}, vol.~93, no. 1-2, pp. 63--101, 1997.

\bibitem{bench2003persuasion}
T.~J. Bench-Capon, ``Persuasion in practical argument using value-based
  argumentation frameworks,'' \emph{Journal of Logic and Computation}, vol.~13,
  no.~3, pp. 429--448, 2003.

\bibitem{stepin2021survey}
I.~Stepin, J.~M. Alonso, A.~Catala, and M.~Pereira-Fari{\~n}a, ``A survey of
  contrastive and counterfactual explanation generation methods for explainable
  artificial intelligence,'' \emph{IEEE Access}, vol.~9, pp. 11\,974--12\,001,
  2021.

\bibitem{jiang2023formalising}
J.~Jiang, F.~Leofante, A.~Rago, and F.~Toni, ``Formalising the robustness of
  counterfactual explanations for neural networks,'' in \emph{AAAI Conference
  on Artificial Intelligence}, 2023, pp. 14\,901--14\,909.

\bibitem{liang2023logic}
C.~Liang, W.~Wang, J.~Miao, and Y.~Yang, ``Logic-induced diagnostic reasoning
  for semi-supervised semantic segmentation,'' in \emph{Proc. IEEE Int. Conf.
  Comput. Vis.}, 2023, pp. 16\,197--16\,208.

\bibitem{zhou2018brief}
Z.-H. Zhou, ``A brief introduction to weakly supervised learning,''
  \emph{National science review}, vol.~5, no.~1, pp. 44--53, 2018.

\bibitem{xia2017joint}
F.~Xia, P.~Wang, X.~Chen, and A.~L. Yuille, ``Joint multi-person pose
  estimation and semantic part segmentation,'' in \emph{Proc. IEEE Conf.
  Comput. Vis. Pattern Recognit.}, 2017, pp. 6769--6778.

\bibitem{chen2018encoder}
L.-C. Chen, Y.~Zhu, G.~Papandreou, F.~Schroff, and H.~Adam, ``Encoder-decoder
  with atrous separable convolution for semantic image segmentation,'' in
  \emph{Proc. Eur. Conf. Comput. Vis.}, 2018, pp. 801--818.

\bibitem{zhang2020part}
X.~Zhang, Y.~Chen, B.~Zhu, J.~Wang, and M.~Tang, ``Part-aware context network
  for human parsing,'' in \emph{Proc. IEEE Conf. Comput. Vis. Pattern
  Recognit.}, 2020, pp. 8971--8980.

\bibitem{wang2021exploring}
W.~Wang, T.~Zhou, F.~Yu, J.~Dai, E.~Konukoglu, and L.~Van~Gool, ``Exploring
  cross-image pixel contrast for semantic segmentation,'' in \emph{Proc. IEEE
  Int. Conf. Comput. Vis.}, 2021, pp. 7303--7313.

\bibitem{zhou2022rethinking}
T.~Zhou, W.~Wang, E.~Konukoglu, and L.~Van~Gool, ``Rethinking semantic
  segmentation: A prototype view,'' in \emph{Proc. IEEE Conf. Comput. Vis.
  Pattern Recognit.}, 2022, pp. 2582--2593.

\bibitem{cheng2022masked}
B.~Cheng, I.~Misra, A.~G. Schwing, A.~Kirillov, and R.~Girdhar,
  ``Masked-attention mask transformer for universal image segmentation,'' in
  \emph{Proc. IEEE Conf. Comput. Vis. Pattern Recognit.}, 2022, pp. 1290--1299.

\bibitem{lianggmmseg}
C.~Liang, W.~Wang, J.~Miao, and Y.~Yang, ``Gmmseg: Gaussian mixture based
  generative semantic segmentation models,'' in \emph{Proc. Advances Neural
  Inf. Process. Syst}, 2022, pp. 31\,360--31\,375.

\bibitem{liang2023clustseg}
J.~Liang, T.~Zhou, D.~Liu, and W.~Wang, ``Clustseg: Clustering for universal
  segmentation,'' in \emph{Proc. Int. Conf. Learn. Representations}, 2023.

\bibitem{wang2017deep}
Y.~Wang, X.~Liu, and S.~Shi, ``Deep neural solver for math word problems,'' in
  \emph{Proceedings of the Conference on Empirical Methods in Natural Language
  Processing}, 2017, pp. 845--854.

\bibitem{wang2018translating}
L.~Wang, Y.~Wang, D.~Cai, D.~Zhang, and X.~Liu, ``Translating a math word
  problem to a expression tree,'' in \emph{Proceedings of the Conference on
  Empirical Methods in Natural Language Processing}, 2018, pp. 1064--1069.

\bibitem{wang2019template}
L.~Wang, D.~Zhang, J.~Zhang, X.~Xu, L.~Gao, B.~T. Dai, and H.~T. Shen,
  ``Template-based math word problem solvers with recursive neural networks,''
  in \emph{AAAI Conference on Artificial Intelligence}, 2019, pp. 7144--7151.

\bibitem{li2019modeling}
J.~Li, L.~Wang, J.~Zhang, Y.~Wang, B.~T. Dai, and D.~Zhang, ``Modeling
  intra-relation in math word problems with different functional multi-head
  attentions,'' in \emph{Proceedings of the Annual Meeting of the Association
  for Computational Linguistics}, 2019, pp. 6162--6167.

\bibitem{bianchi2019capabilities}
F.~Bianchi and P.~Hitzler, ``On the capabilities of logic tensor networks for
  deductive reasoning.'' \emph{AAAI Spring Symposium: Combining Machine
  Learning with Knowledge Engineering}, 2019.

\bibitem{bianchi2019complementing}
F.~Bianchi, M.~Palmonari, P.~Hitzler, and L.~Serafini, ``Complementing logical
  reasoning with sub-symbolic commonsense,'' in \emph{International Joint
  Conference on Rules and Reasoning}, 2019, pp. 161--170.

\bibitem{eberhart2020completion}
A.~Eberhart, M.~Ebrahimi, L.~Zhou, C.~Shimizu, and P.~Hitzler, ``Completion
  reasoning emulation for the description logic el+,'' \emph{AAAI Spring
  Symposium: Combining Machine Learning with Knowledge Engineering}, 2020.

\bibitem{chaturvedi2019fuzzy}
I.~Chaturvedi, R.~Satapathy, S.~Cavallari, and E.~Cambria, ``Fuzzy commonsense
  reasoning for multimodal sentiment analysis,'' \emph{Pattern Recognition
  Letters}, vol. 125, pp. 264--270, 2019.

\bibitem{van2023nesi}
E.~van Krieken, T.~Thanapalasingam, J.~Tomczak, F.~Van~Harmelen, and
  A.~Ten~Teije, ``A-nesi: A scalable approximate method for probabilistic
  neurosymbolic inference,'' in \emph{Proc. Advances Neural Inf. Process.
  Syst}, 2023, pp. 24\,586--24\,609.

\bibitem{maene2024soft}
J.~Maene and L.~De~Raedt, ``Soft-unification in deep probabilistic logic,'' in
  \emph{Proc. Advances Neural Inf. Process. Syst}, 2023, pp. 60\,804--60\,820.

\bibitem{li2024neuro}
Z.~Li, Y.~Huang, Z.~Li, Y.~Yao, J.~Xu, T.~Chen, X.~Ma, and J.~Lu,
  ``Neuro-symbolic learning yielding logical constraints,'' in \emph{Proc.
  Advances Neural Inf. Process. Syst}, 2024, pp. 21\,635--21\,657.

\bibitem{kim2021improving}
J.~Kim, P.~Ravikumar, J.~Ainslie, and S.~Ontanon, ``Improving compositional
  generalization in classification tasks via structure annotations,'' in
  \emph{Proceedings of the Annual Meeting of the Association for Computational
  Linguistics and the International Joint Conference on Natural Language
  Processing}, 2021, pp. 637--645.

\bibitem{garcez2006connectionist}
A.~S.~d. Garcez, L.~C. Lamb, and D.~M. Gabbay, ``Connectionist computations of
  intuitionistic reasoning,'' \emph{Theor. Comput. Sci.}, vol. 358, no.~1, pp.
  34--55, 2006.

\bibitem{garcez2007connectionist}
------, ``Connectionist modal logic: Representing modalities in neural
  networks,'' \emph{Theor. Comput. Sci.}, vol. 371, no. 1-2, pp. 34--53, 2007.

\bibitem{khardon1997learning}
R.~Khardon and D.~Roth, ``Learning to reason,'' \emph{Journal of the ACM},
  vol.~44, no.~5, pp. 697--725, 1997.

\bibitem{felgenbaum1977art}
E.~A. Felgenbaum, ``The art of artificial intelligence: themes and case studies
  of knowledge engineering,'' in \emph{International Joint Conferences on
  Artificial Intelligence}, 1977, pp. 1014--1029.

\bibitem{ji2021survey}
S.~Ji, S.~Pan, E.~Cambria, P.~Marttinen, and S.~Y. Philip, ``A survey on
  knowledge graphs: Representation, acquisition, and applications,'' \emph{IEEE
  Trans. Neural Netw. Learning Sys.}, vol.~33, no.~2, pp. 494--514, 2021.

\bibitem{roese1997counterfactual}
N.~J. Roese, ``Counterfactual thinking.'' \emph{Psychological Bulletin}, vol.
  121, no.~1, p. 133, 1997.

\bibitem{liang2022visual}
C.~Liang, W.~Wang, T.~Zhou, and Y.~Yang, ``Visual abductive reasoning,'' in
  \emph{Proc. IEEE Conf. Comput. Vis. Pattern Recognit.}, 2022, pp.
  15\,565--15\,575.

\bibitem{yang2024doraemongpt}
Z.~Yang, G.~Chen, X.~Li, W.~Wang, and Y.~Yang, ``Doraemongpt: Toward
  understanding dynamic scenes with large language models (exemplified as a
  video agent),'' in \emph{Proc. ACM Int. Conf. Mach. Learn.}, 2024.

\end{thebibliography}
}

%


\vfill


\end{document}